\documentclass{article}

\PassOptionsToPackage{numbers, compress}{natbib}

\usepackage[final]{neurips_2022}

\usepackage[utf8]{inputenc} 
\usepackage[T1]{fontenc}    
\usepackage{hyperref}       
\usepackage{url}            
\usepackage{booktabs}       
\usepackage{amsfonts}       
\usepackage{nicefrac}       
\usepackage{microtype}      
\usepackage{xcolor}         

\usepackage{amsmath}
\usepackage[pdftex]{graphicx} 
\usepackage{subcaption}
\usepackage{wrapfig}
\usepackage{multirow}
\usepackage{tikz}
\usepackage{caption}
\usepackage{placeins}
\usepackage{xr} 

\makeatletter
\newcommand*{\addFileDependency}[1]{
	\typeout{(#1)}
	\@addtofilelist{#1}
	\IfFileExists{#1}{}{\typeout{No file #1.}}
}
\makeatother
\newcommand*{\myexternaldocument}[1]{%
	\externaldocument{#1}%
	\addFileDependency{#1.tex}%
	\addFileDependency{#1.aux}%
}
\myexternaldocument{neurips_2022_supp}

\DeclareMathOperator*{\argmax}{argmax}

\DeclareMathOperator*{\softmax}{softmax}
\DeclareMathOperator{\softplus}{softplus}
\DeclareMathOperator{\leakyrelu}{leakyReLU}

\title{What Makes Graph Neural Networks Miscalibrated?}

\author{%
	Hans Hao-Hsun Hsu\thanks{Equal contribution} $\ ^1$
	\quad Yuesong Shen\footnotemark[1]  $\ ^{1,2}$
	\quad Christian Tomani $^{1,2}$
	\quad Daniel Cremers  $^{1,2}$\\
	$^1\ $Technical University of Munich, Germany\\
	$^2\ $Munich Center for Machine Learning, Germany\\
	\texttt{\{hans.hsu, yuesong.shen, christian.tomani, cremers\}@tum.de}\\
}

\begin{document}

	\maketitle

	\begin{abstract}
		Given the importance of getting calibrated predictions and reliable uncertainty estimations, various post-hoc calibration methods have been developed for neural networks on standard multi-class classification tasks. 
		However, these methods are not well suited for calibrating graph neural networks (GNNs), which presents unique challenges such as accounting for the graph structure and the graph-induced correlations between the nodes.
		In this work, we conduct a systematic study on the calibration qualities of GNN node predictions. 
		In particular, we identify five factors which influence the calibration of GNNs: general under-confident tendency, diversity of nodewise predictive distributions, distance to training nodes, relative confidence level, and neighborhood similarity. 
		Furthermore, based on the insights from this study, we design a novel calibration method named Graph Attention Temperature Scaling (GATS), which is tailored for calibrating graph neural networks. 
		GATS incorporates designs that address all the identified influential factors and produces nodewise temperature scaling using an attention-based architecture. 
		GATS is accuracy-preserving, data-efficient, and expressive at the same time. 
		Our experiments empirically verify the effectiveness of GATS, demonstrating that it can consistently achieve state-of-the-art calibration results on various graph datasets for different GNN backbones.\footnote{Source code available at \url{https://github.com/hans66hsu/GATS}} 
	\end{abstract}

	\section{Introduction}

	
	Graph-structured data, such as social networks, knowledge graphs and internet of things, have wide-spread presence and learning on graphs using neural networks has been an active area of research. For node classification on graphs, a wide range of graph neural network (GNN) models, including GCN \citep{kipf2017gcn}, GAT \citep{velivckovic2018gat} and GraphSAGE \citep{hamilton2017graphsage}, have been proposed to achieve high classification accuracy. 
	
	This said, high accuracy is not the only desideratum for a classifier. Especially, reliable uncertainty estimation is crucial for applications like safety critical tasks and active learning. Neural networks are known to produce poorly calibrated predictions that are either overconfident or under-confident \citep{guo2017calibration,wang2021cagcn}. To mitigate this issue a variety of post-hoc calibration methods \citep{guo2017calibration,kull2019dirichlet,zhang2020mixnmatch,tomani2021calibration,gupta2021spline} have been introduced over the last few years for calibrating neural networks on standard multi-class classification problems. However, calibration of GNNs, in the context of node classification on graphs, is currently still an underexplored topic. While it is possible to apply existing calibration methods designed for multi-class classification to GNNs in a nodewise manner, this does not address the specific challenges of node classification on graphs. Especially, node predictions in a graph are not i.i.d.\ but correlated, and we are tackling a structured prediction problem \citep{nowozin2011structured}. A uniform treatment when calibrating node predictions would fail to account for the structural information from graphs and the non i.i.d.\ behavior of node predictions.
	
	\paragraph{Our contribution.}
	In this work, we focus on calibrating GNNs for the node classification task \citep{kipf2017gcn, velivckovic2018gat}.
	First, we aim at understanding the specific challenges posed by GNNs by conducting a systematic study on the calibration qualities of GNN node predictions. Our study reveals five factors that influence the calibration performance of GNNs: general under-confident tendency, diversity of nodewise predictive distributions, distance to training nodes, relative confidence level, and neighborhood similarity.
	Second, we develop Graph Attention Temperature Scaling (GATS) approach, which is designed in a way that accounts for the aforementioned influential factors. 
	GATS generates nodewise temperatures that calibrate GNN predictions based on the graph topology. 
	Third, we conduct a series of GNN calibration experiments and empirically verify the effectiveness of GATS in terms of calibration, data-efficiency, and expressivity.

	\section{Related work}

	
	For standard multi-class classification tasks, a variety of post-hoc calibration methods have been proposed in order to make neural networks uncertainty aware: temperature scaling (TS) \citep{guo2017calibration}, ensemble temperature scaling (ETS) \cite{zhang2020mixnmatch}, multi-class isotonic regression (IRM) \cite{zhang2020mixnmatch}, Dirichlet calibration \citep{kull2019dirichlet}, spline calibration \citep{gupta2021spline}, etc. Additionally, calibration has been formulated for regression tasks \citep{kuleshov2018quantile}. More generally, instead of transforming logits after training a classifier, a plethora of methods exists that modify either the model architecture or the training process itself. This includes methods that are based on Bayesian paradigm \cite{lobato2015pbp,blundell15weightuncertainty,gal2016mcdropout,maddox2019swag,wen2018flipout}, evidential theory \cite{sensoy2018evidentialdl}, adversarial calibration \cite{tomani2021towards} and model ensembling \cite{lakshminarayanan2017deepensemble}. One common caveat of these methods is the trade-off between accuracy and calibration, which oftentimes do not go hand in hand. 
	Post-hoc methods like temperature scaling, on the other hand, are accuracy preserving. They ensure that the per node logit rankings are unaltered. 
	
	Calibration of GNNs is currently a substantially less explored topic. Nodewise post-hoc calibration on GNNs using methods developed for the multi-class setting has been empirically evaluated by \citet{teixeira2019gnncalib}. They show that these methods, which perform uniform calibration of nodewise predictions, are unable to produce calibrated predictions for some harder tasks. \citet{wang2021cagcn} observe that GNNs tend to be under-confident in contrast to the majority of multi-class classifiers, which are generally overconfident \citep{guo2017calibration}. Based on their findings, \citet{wang2021cagcn} propose the CaGCN approach, which attaches a GCN on top of the backbone GNN for calibration. Some approaches improve the uncertainty estimation of GNNs by adjusting model training. This includes Bayesian learning approaches \citep{zhang2019bayesgcn,hasanzadeh2020graphdropconnect} and methods based on the evidential theory \citep{zhao2020graphedl,graph-postnet}.

	\section{Problem setup for GNN calibration} \label{sec:calibration}
	We consider the problem of calibrating GNNs for node classification tasks:
	given a graph $\mathcal{G} = (\mathcal{V}, \mathcal{E})$, the training data consist of nodewise input features $\{x_i\}_{i\in\mathcal{V}} \in \mathcal{X}$ and ground-truth labels $\{y_i\}_{i\in\mathcal{L}}\in \mathcal{Y}=\{1,\dots, K\}$ for a subset $\mathcal{L} \subset \mathcal{V}$ of nodes, and the goal is to predict the labels $\{y_i\}_{i\in\mathcal{U}}\in \mathcal{Y}$ for the rest of the nodes $\mathcal{U} = \mathcal{V} \setminus \mathcal{L}$. 
	A graph neural network tackles the problem by producing nodewise probabilistic forecasts $\hat{p}_i$. 
	These forecasts yield the corresponding label predictions $\hat{y}_i:=\argmax_y \hat{p}_i(y)$ and confidences $\hat{c}_i:=\max_{y} \hat{p}_i(y)$. 
	The GNN is calibrated when its probabilistic forecasts are reliable, e.g.,\ for predictions with confidence $0.8$, they should be correct $80\%$ of the time. 
	Formally, a GNN is \emph{perfectly calibrated} \citep{wang2021cagcn} if
	\begin{equation}
	    \forall c \in [0,1], \quad \mathbb{P}(y_i = \hat{y}_i|\hat{c}_i=c) = c.
	\end{equation}
	In practice, we quantify the calibration quality with the expected calibration error (ECE) \citep{Naeini_2015,guo2017calibration}. We follow the commonly used definition from \citet{guo2017calibration} which uses a equal width binning scheme to estimate calibration error for any node subset $\mathcal{N} \subset \mathcal{V}$: the predictions are regrouped according to $M$ equally spaced confidence intervals, i.e.\ $(B_1,\dots,B_M)$ with $B_m = \{j\in \mathcal{N} \,|\, \frac{m-1}{M} < \hat{c}_j \leq \frac{m}{M} \}$,
	and the expected calibration error of the GNN forecasts is defined as
		\begin{align}
	    &\text{ECE} = \sum_{m=1}^{M}\frac{|B_m|}{|\mathcal{N}|} \Big| \text{acc}(B_m) - \text{conf}(B_m) \Big|, \text{ with } \\
	    &\text{acc}(B_m)=\frac{1}{|B_m|}\sum_{i\in B_m} \mathbf{1}(y_i=\hat{y}_i) \text{ and }
	    \text{conf}(B_m)=\frac{1}{|B_m|}\sum_{i\in B_m}\hat{c}_i.
    	\end{align}
	
	\section{Factors that influence GNN calibration} \label{sec:gnn_calib_factors}

	
	To design calibration methods adapted to GNNs, we need to figure out the particular factors that influence the calibration quality of GNN predictions. For this we train a series of GCN \cite{kipf2017gcn} and GAT \citep{velivckovic2018gat} models on seven graph datasets: Cora \citep{sen2008coraciteseer}, Citeseer \citep{sen2008coraciteseer}, Pubmed \citep{namata2012pubmed}, Amazon Computers \citep{shchur2018pitfalls}, Amazon Photo \citep{shchur2018pitfalls}, Coauthor CS \citep{shchur2018pitfalls}, and Coauthor Physics \citep{shchur2018pitfalls}. We summarize the dataset statistics in Appendix~\ref{subsec:dataset_stat}. Details about model training are provided in Appendix~\ref{subsec:details_train} for reproducibility. To compare with the standard multi-class classification case, we additionally train ResNet-20 \citep{he2016preresnet} models on the CIFAR-10 image classification task \citep{krizhevsky2009cifar10} as a reference. 
	
	Our experiments uncover five decisive factors that affect the calibration quality of GNNs.
	In the following we discuss them in detail.

	\subsection{General under-confident tendency}
	
	\begin{figure}[ht]
		\centering
		\setlength{\tabcolsep}{0.04\textwidth} 
		\begin{tabular}{cccc}
			\includegraphics[width=0.15\textwidth]{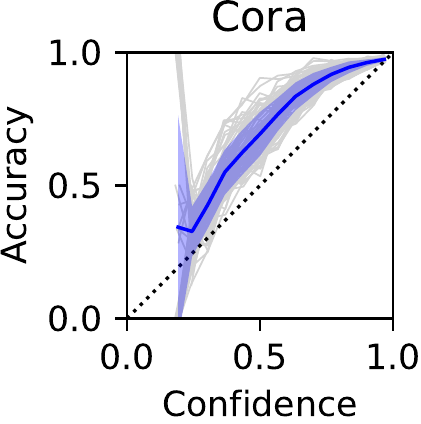} &
			\includegraphics[width=0.15\textwidth]{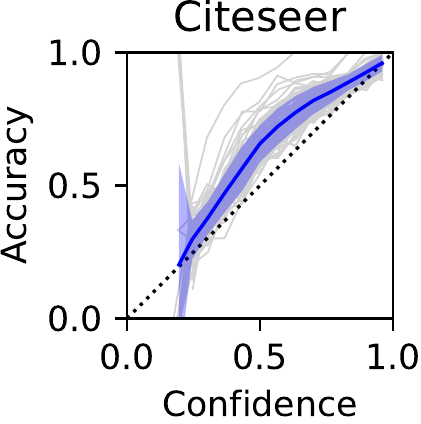} &
			\includegraphics[width=0.15\textwidth]{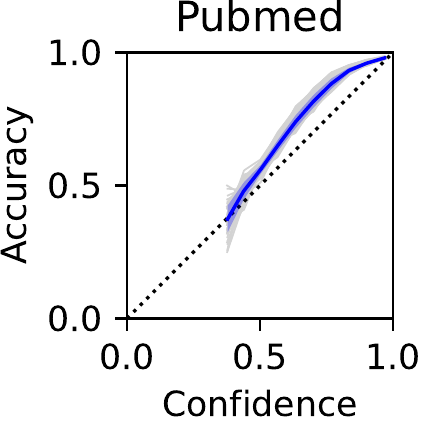} &
			\includegraphics[width=0.15\textwidth]{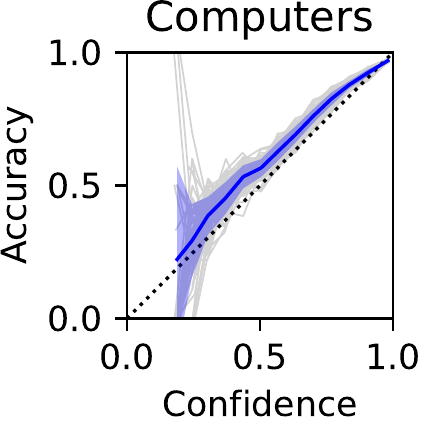} \\
			\includegraphics[width=0.15\textwidth]{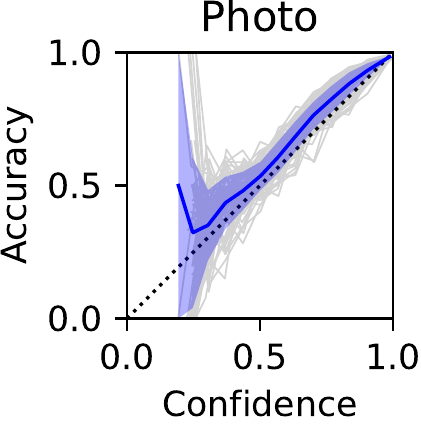} &
			\includegraphics[width=0.15\textwidth]{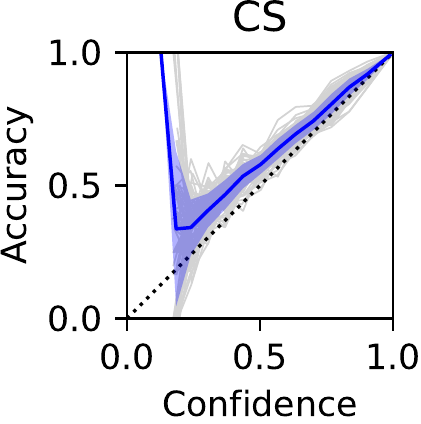} &
			\includegraphics[width=0.15\textwidth]{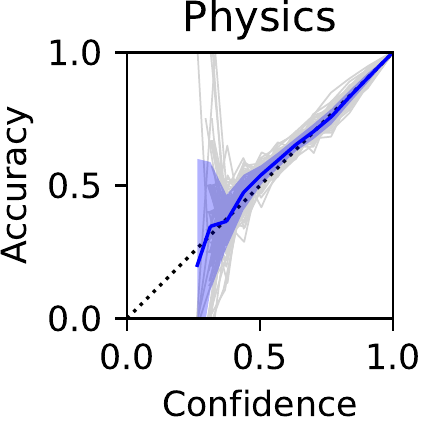} &
			\includegraphics[width=0.15\textwidth]{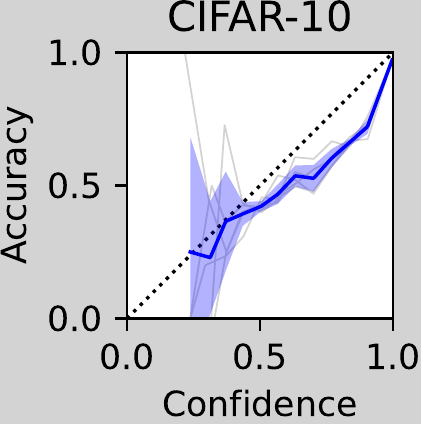}
		\end{tabular}
		\caption{Reliability diagrams of GCN models trained on various graph datasets. We see a general tendency of under-confident predictions (plots above the diagonal) except the Physics dataset. This is in contrast to the overconfident behavior of multi-class image classification using CNNs (in gray).} \label{fig:miscalib_tendency_gcn}
	\end{figure}
	
	Starting with a global perspective, we notice that GNNs tend to produce under-confident predictions. In Figure~\ref{fig:miscalib_tendency_gcn} we plot the reliability diagrams \citep{mizil2015} for results on different graph datasets using GCN. Similar to \citet{wang2021cagcn}, we see a general trend of under-confident predictions for GNNs. This is in contrast to the standard multi-class image classification case which has overconfident behavior. Also, it is interesting to see that this under-confident trend can be more or less pronounced depending on the dataset. For Coauthor Physics, the predictions are well calibrated and have no significant bias. 
	
	Results using GAT models lead to similar conclusions and are provided in Appendix~\ref{subsec:miscalib_tendency}.

	\subsection{Diversity of nodewise predictive distributions}
	
	\begin{figure}[ht]
		\centering
		\begin{subfigure}[c]{0.15\textwidth}
			\includegraphics[width=\textwidth]{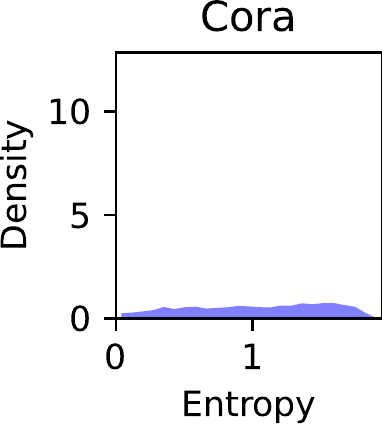}
		\end{subfigure}
		\hspace{0.06\textwidth}
		\begin{subfigure}[c]{0.15\textwidth}
			\includegraphics[width=\textwidth]{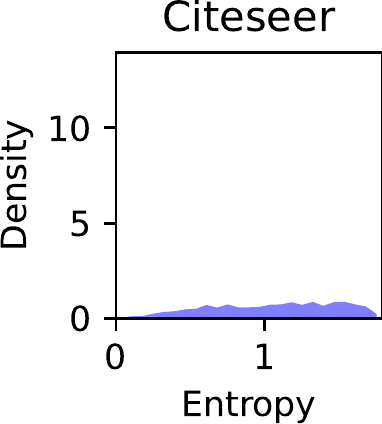}
		\end{subfigure}
		\hspace{0.06\textwidth}
		\begin{subfigure}[c]{0.15\textwidth}
			\includegraphics[width=\textwidth]{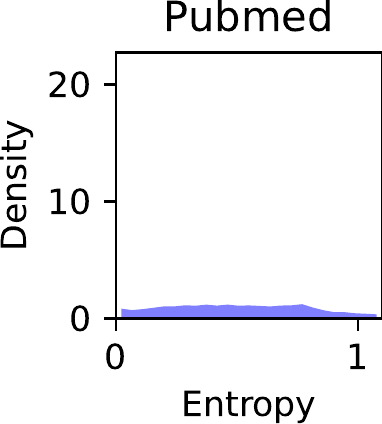}
		\end{subfigure}
		\hspace{0.06\textwidth}
		\begin{subfigure}[c]{0.15\textwidth}
			\includegraphics[width=\textwidth]{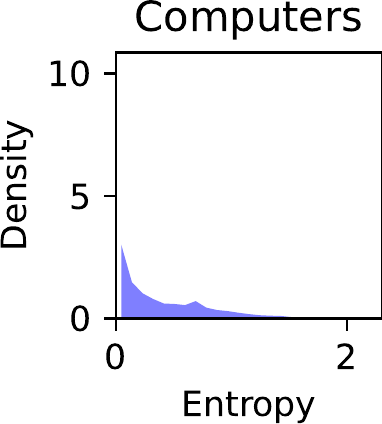}
		\end{subfigure} \\
		\begin{subfigure}[c]{0.15\textwidth}
			\includegraphics[width=\textwidth]{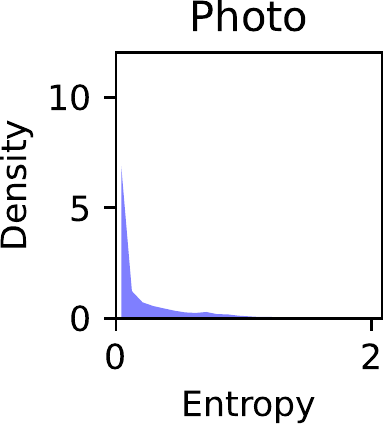}
		\end{subfigure}
		\hspace{0.06\textwidth}
		\begin{subfigure}[c]{0.15\textwidth}
			\includegraphics[width=\textwidth]{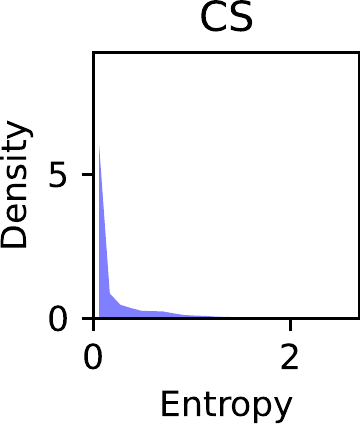}
		\end{subfigure}
		\hspace{0.06\textwidth}
		\begin{subfigure}[c]{0.15\textwidth}
			\includegraphics[width=\textwidth]{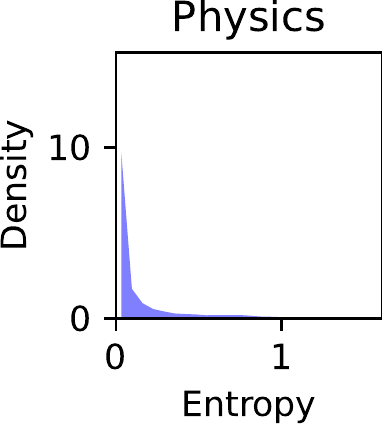}
		\end{subfigure}
		\hspace{0.06\textwidth}
		\begin{subfigure}[c]{0.15\textwidth}
			\includegraphics[width=\textwidth]{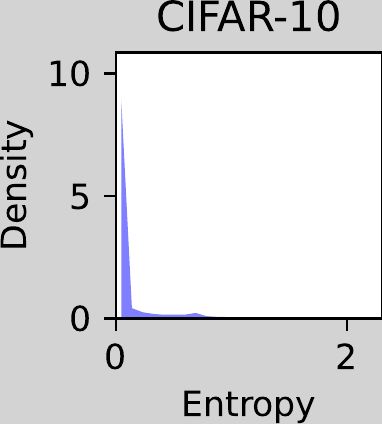}
		\end{subfigure}
		\caption{Entropy distributions of GCN predictions on graph datasets. Compared to the standard classification case, GNN predictions tend to be more dispersed, reflecting their disparate behaviors.} \label{fig:miscalib_nodestat_gcn}
	\end{figure}
	
	Contrary to the standard multi-class case, GNN outputs can have varying roles depending on their positions in the graph, which means that their output distributions could exhibit dissimilar behaviors. This is empirically evident in Figure~\ref{fig:miscalib_nodestat_gcn}, where we visualize the entropy distributions of GCN output predictions v.s.\ the standard multi-class results (GAT results are available in Appendix~\ref{subsec:miscalib_nodestat_gat}). We see that the entropies of GNN outputs have more spread-out distributions, which indicates that they have distinct roles and behaviors in graphs.
	
	In terms of GNN calibration, this observation implies that uniform node-agnostic adjustments like temperature scaling \citep{guo2017calibration} might be insufficient for GNNs, whereas nodewise adaptive approaches could be beneficial.

	\subsection{Distance to training nodes}
	
	\begin{figure}[ht]
		\centering
		\begin{subfigure}[c]{0.135\textwidth}
			\includegraphics[width=\textwidth]{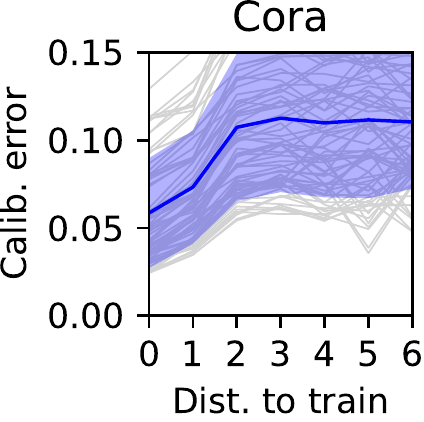}
		\end{subfigure}
		\begin{subfigure}[c]{0.135\textwidth}
			\includegraphics[width=\textwidth]{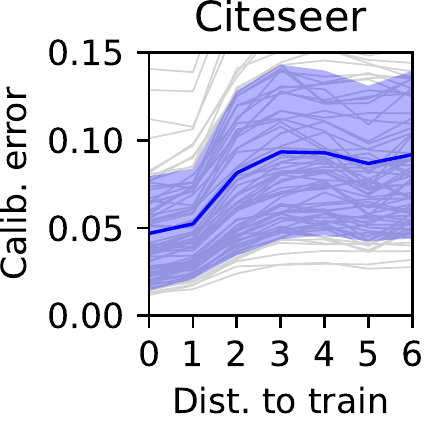}
		\end{subfigure}
		\begin{subfigure}[c]{0.135\textwidth}
			\includegraphics[width=\textwidth]{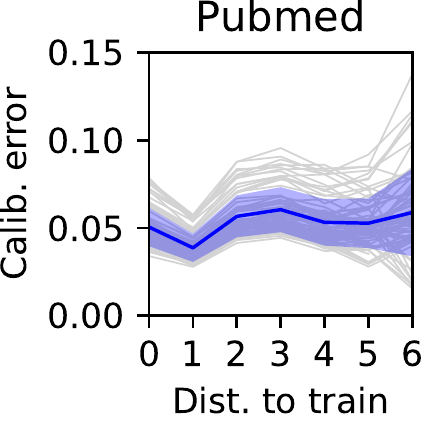}
		\end{subfigure}
		\begin{subfigure}[c]{0.135\textwidth}
			\includegraphics[width=\textwidth]{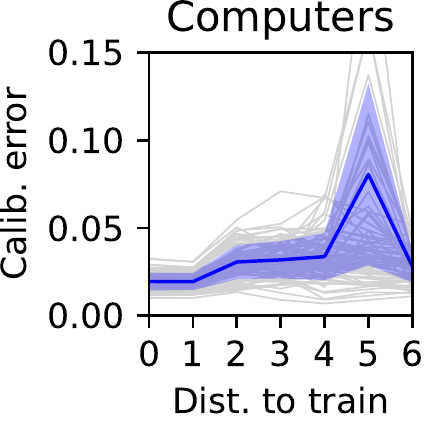}
		\end{subfigure}
		\begin{subfigure}[c]{0.135\textwidth}
			\includegraphics[width=\textwidth]{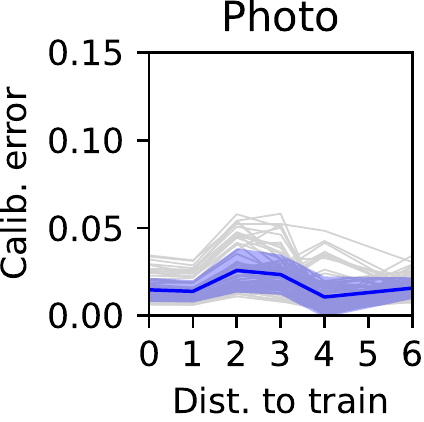}
		\end{subfigure}
		\begin{subfigure}[c]{0.135\textwidth}
			\includegraphics[width=\textwidth]{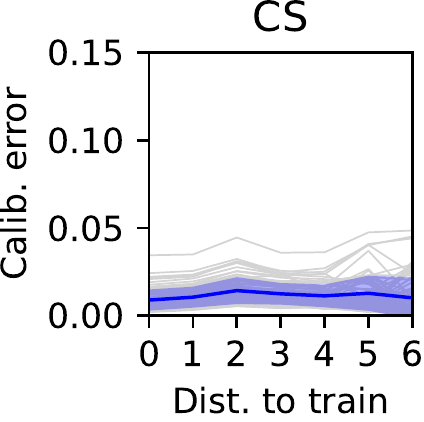}
		\end{subfigure}
		\begin{subfigure}[c]{0.135\textwidth}
			\includegraphics[width=\textwidth]{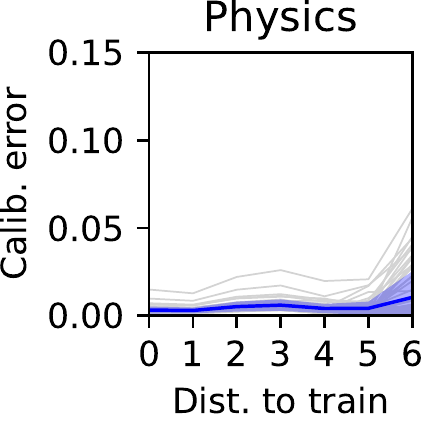}
		\end{subfigure}
		\caption{Nodewise calibration error of GCN results depending on the minimum distance to training nodes. We observe that training nodes and their neighbors tend to be better calibrated.} \label{fig:miscalib_traindist_gcn}
	\end{figure}
	
	A graph provides additional structural information for its nodes. One insightful feature is the minimum distance to training nodes. We discover that nodes with shorter distances, especially the training nodes themselves and their direct neighbors, tend to be better calibrated.  
	
	To evaluate the calibration quality nodewise, we propose the \textsl{nodewise calibration error}, which is based on the binning scheme used to compute the global expected calibration error (ECE) \citep{Naeini_2015, guo2017calibration}: for each node, we find its corresponding bin depending on its predicted confidence, and the calibration error of this bin is assigned to be its nodewise calibration error.
	
	Using this nodewise metric, in Figure~\ref{fig:miscalib_traindist_gcn} we visualize the influence of minimum  distance to training nodes on the nodewise calibration quality (c.f.\ Appendix~\ref{subsec:miscalib_traindist_gat} for GAT results). We see that nodes close to training ones typically have lower nodewise calibration error. This suggests that minimum distance to training nodes can be useful for GNN calibration.

	\subsection{Relative confidence level}
	
	\begin{figure}[ht]
		\centering
		\begin{subfigure}[c]{0.135\textwidth}
			\includegraphics[width=\textwidth]{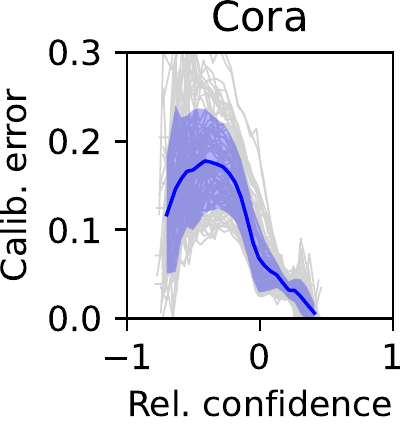}
		\end{subfigure}
		\begin{subfigure}[c]{0.135\textwidth}
			\includegraphics[width=\textwidth]{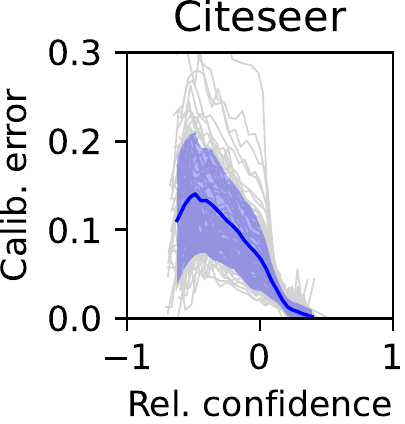}
		\end{subfigure}
		\begin{subfigure}[c]{0.135\textwidth}
			\includegraphics[width=\textwidth]{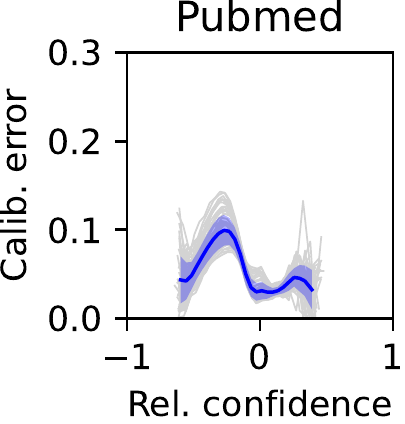}
		\end{subfigure}
		\begin{subfigure}[c]{0.135\textwidth}
			\includegraphics[width=\textwidth]{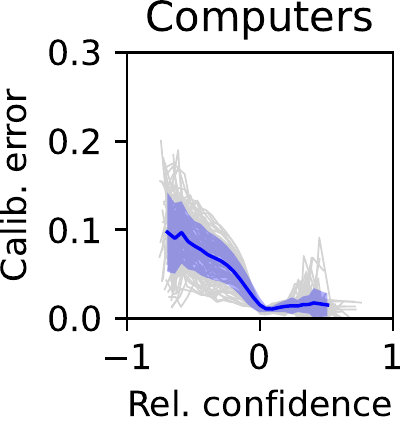}
		\end{subfigure}
		\begin{subfigure}[c]{0.135\textwidth}
			\includegraphics[width=\textwidth]{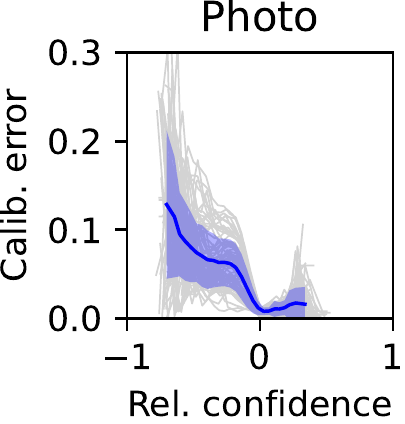}
		\end{subfigure}
		\begin{subfigure}[c]{0.135\textwidth}
			\includegraphics[width=\textwidth]{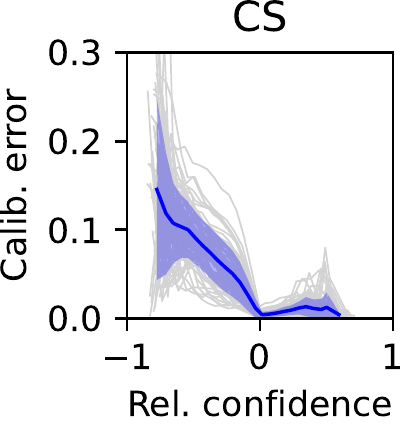}
		\end{subfigure}
		\begin{subfigure}[c]{0.135\textwidth}
			\includegraphics[width=\textwidth]{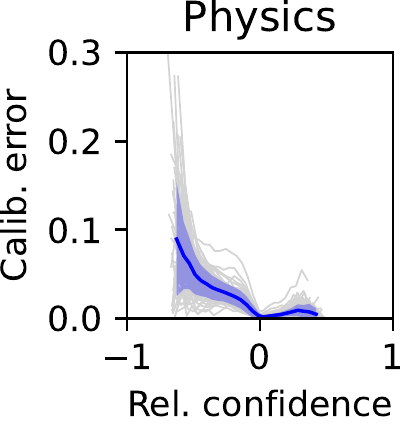}
		\end{subfigure}
		\hspace{0.2\textwidth}
		\caption{Nodewise calibration error of GCN results depending on the relative confidence level. We observe that nodes which are less confident than their neighbors tend to have worse calibration.} \label{fig:miscalib_diffconf_gcn}
	\end{figure}
	
	Another important structural information is the neighborhood relation. We find out that the relative confidence level $\delta \hat{c}_i$ of a node $i$, i.e.,\ the difference between the nodewise confidence $\hat{c}_i$ and the average confidence of its neighbors
	\begin{equation}
	\delta \hat{c}_i = \hat{c}_i - \frac{1}{|n(i)|}\sum_{j \in n(i)} \hat{c}_j
	\label{eq:delta_ci}
	\end{equation}
	has an interesting correlation to the nodewise calibration quality. In Figure~\ref{fig:miscalib_diffconf_gcn} we show the relation between the relative confidence level of a node and its nodewise calibration error (c.f.\ Appendix~\ref{subsec:miscalib_diffconf_gat} for GAT results). Especially, We observe that nodes which are less confident than their neighbors tend to have worse calibration, and it is in general desirable to have comparable confidence level w.r.t.\ the neighbors. 
	For GNN calibration, the relative confidence level $\delta\hat{c}_i$ can be a useful node feature to consider.
	
	\subsection{Neighborhood similarity}
	\label{subsec:similarity}
	
	\begin{figure}[ht]
		\centering
		\begin{subfigure}[c]{0.135\textwidth}
			\includegraphics[width=\textwidth]{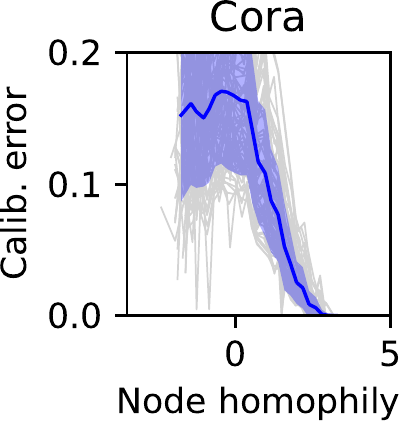}
		\end{subfigure}
		\begin{subfigure}[c]{0.135\textwidth}
			\includegraphics[width=\textwidth]{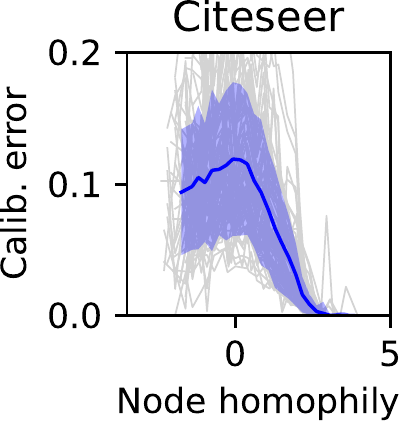}
		\end{subfigure}
		\begin{subfigure}[c]{0.135\textwidth}
			\includegraphics[width=\textwidth]{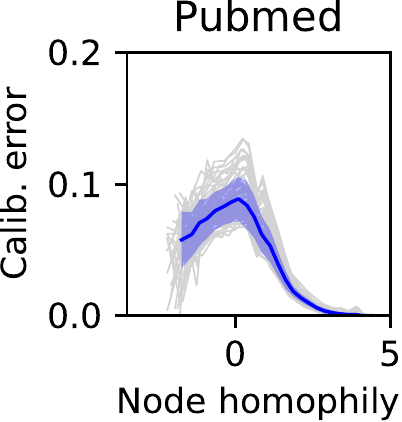}
		\end{subfigure}
		\begin{subfigure}[c]{0.135\textwidth}
			\includegraphics[width=\textwidth]{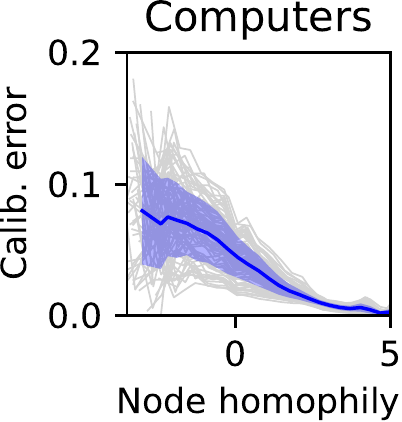}
		\end{subfigure}
		\begin{subfigure}[c]{0.135\textwidth}
			\includegraphics[width=\textwidth]{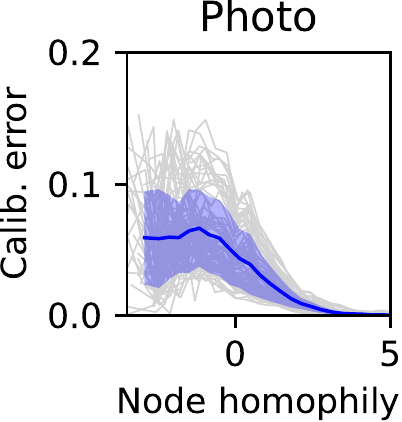}
		\end{subfigure}
		\begin{subfigure}[c]{0.135\textwidth}
			\includegraphics[width=\textwidth]{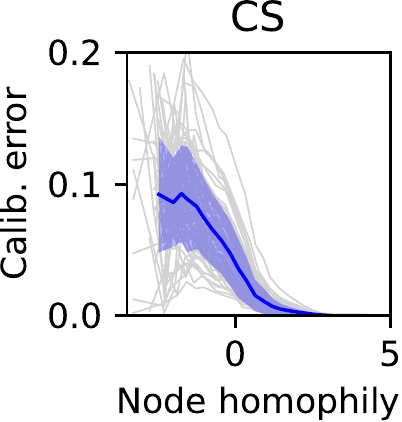}
		\end{subfigure}
		\begin{subfigure}[c]{0.135\textwidth}
			\includegraphics[width=\textwidth]{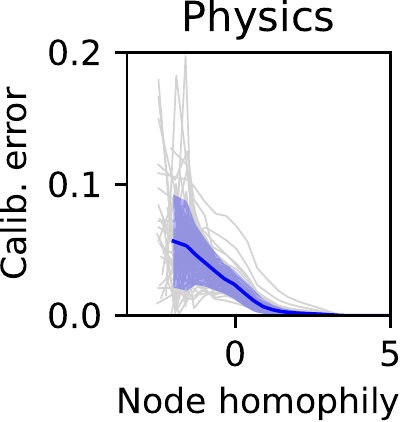}
		\end{subfigure}
		\hspace{0.2\textwidth}
		\caption{Nodewise calibration error of GCN results depending on the node homophily. Nodes with strongly agreeing neighbors tend to have significantly lower calibration errors.} \label{fig:miscalib_neighbor_gcn}
	\end{figure}
	
	Furthermore, we find that different neighbors tend to introduce distinct influences. For assortative graphs which are the focus of this work, we find out that calibration of nodes are affected by \textsl{node homophily}, i.e.,\ whether a node tends to have the same label prediction as its neighbors. For a node with $n_a$ agreeing neighbors and $n_d$ disagreeing ones, we measure the node homophily as
	\begin{equation}
	\text{Node homophily} = \log \Big( \frac{n_a + 1}{n_d + 1} \Big),
	\end{equation}
	where positive values indicate greater ratio of agree neighbors and vice versa.
	
	Figure~\ref{fig:miscalib_neighbor_gcn} summarizes the variation of nodewise calibration error w.r.t.\ the node homophily for different graph datasets (c.f.\ Appendix~\ref{subsec:miscalib_neighbor_gat} for GAT results). We find out that nodewise calibration errors tend to decrease significantly for nodes with strongly agreeing neighbors. This suggests that neighborhood predictive similarity should be considered when doing GNN calibration.

	\section{Graph attention temperature scaling (GATS)}

	
	Based on the findings in Section~\ref{sec:gnn_calib_factors}, we design a new post-hoc calibration method, named \textsl{Graph Attention Temperature Scaling (GATS)}, which is tailored for GNNs.

	\subsection{Formulation and design of GATS} 
	
	To obtain a calibration method that is adapted to the graph structure $\mathcal{G} = (\mathcal{V}, \mathcal{E})$ and reflects the observed influential factors in Section~\ref{sec:gnn_calib_factors}, the graph attention temperature scaling approach extends the temperature scaling \citep{guo2017calibration} method to produce a distinct temperature $T_i$ for each node $i \in \mathcal{V}$. $T_i$ is then used to scale the uncalibrated nodewise output logits $z_i$ and produce calibrated node predictions $\hat{p}_i$
	\begin{equation}
	\forall i \in \mathcal{V}, \quad \hat{p}_i = \softmax\Big( \frac{z_i}{T_i} \Big).
	\end{equation}
	
	\begin{figure}[t]
		\centering
		\includegraphics[width=0.98\textwidth]{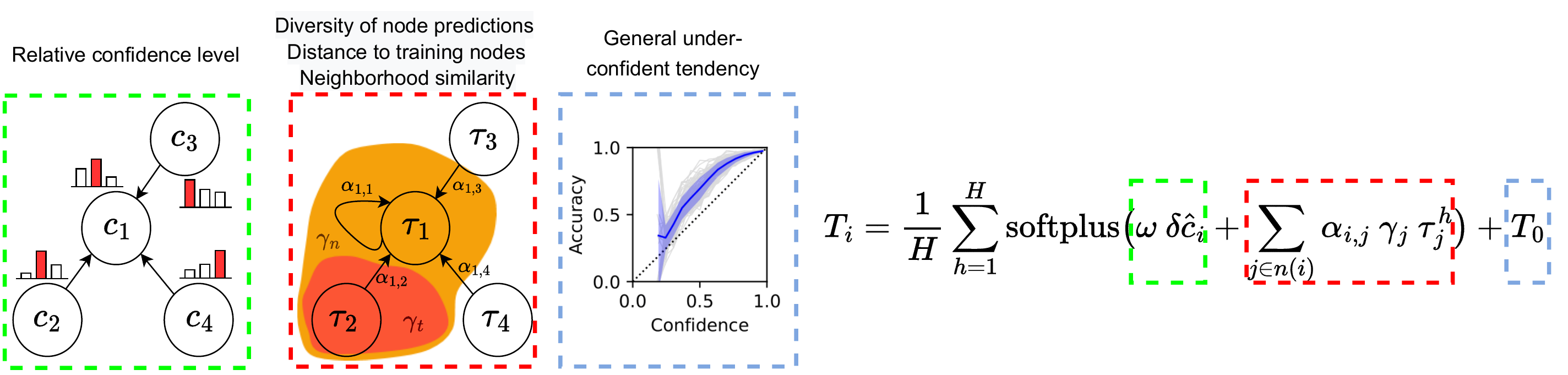}
		\caption{Illustration of graph attention temperature scaling (GATS) for a graph with four nodes where node 2 is a training node and node 1 is the target node which aggregates information from the other three nodes. GATS incorporates the five factors discussed in Section~\ref{sec:gnn_calib_factors} and produces nodewise temperatures $T_i$ using the graph structure and an attention mechanism.}
	\end{figure}
	\paragraph{Formulation of $T_i$.}
	The nodewise temperature $T_i$ should address the five factors discussed in Section~\ref{sec:gnn_calib_factors}. We achieve this via the following considerations:
	\begin{itemize}
		\item We introduce a global bias parameter $T_0$ to account for the general under-confident tendency;
		\item To tackle the diverse behavior of node predictions, we learn a nodewise temperature contribution $\tau_i$ based on the predicted nodewise logits $z_i$;
		\item To incorporate the relative confidence w.r.t.\ neighbors, we introduce $\delta \hat{c}_i$ from Eq. \ref{eq:delta_ci} as an additional contribution term scaled by a learnable coefficient $\omega$;
		\item To model the influence of neighborhood similarity, we use an attention mechanism \citep{vaswani2017attention} to aggregate neighboring contributions $\tau_j$ with attention coefficients $\alpha_{i,j}$ depending on the output similarities between the neighbors $i$ and $j$;
		\item Distance to training nodes is used to introduce a nodewise scaling factor $\gamma_i$ to adjust the node contribution and the aggregation process. It is learnable for training nodes and their direct neighbors and fixed to $1$ for the rest:
		\begin{equation}
		\gamma_i = \begin{cases}
		\gamma_t, &\text{ if } i \text{ is a training node} \\
		\gamma_n, &\text{ if } i \text{ is a neighbor of training node} \\
		1, &\text{ otherwise}
		\end{cases}, \quad \gamma_t, \gamma_n \text{ learnable parameters}.
		\end{equation} 
	\end{itemize}
	Putting together the above components, the nodewise temperature $T_i$ has the following expression:
	\begin{equation}
	\forall i \in \mathcal{V}, \quad T_i = \frac{1}{H} \sum_{h=1}^H \softplus \big( \omega\ \delta \hat{c}_i + \sum_{j \in \hat{n}(i)} \alpha_{i,j} \ \gamma_j\ \tau^h_j \big) + T_0.
	\label{eq:gats_t}
	\end{equation}
	Here we have a multi-head formulation where $h$ indicates the $h$-th attention head, and $\hat{n}(i)$ denotes the neighbors of node $i$ including self-loop. We uses 8 heads ($H=8$), which works well in practice.  (c.f.\ Section~\ref{subsec:exp_ablation}.)
	
	\paragraph{Defining $\tau^h_i$.} Nodewise contribution $\tau^h_i$ are computed as the outputs of parameterized linear layers $\phi^h(\cdot;\theta)$ which take transformed nodewise output logits $\tilde{z}_i$ as input
	\begin{equation}
	\forall i \in \mathcal{V}, \forall h \in [1, H], \quad \tau^h_i = \phi^h(\tilde{z}_i; \theta^h) = (\theta^h)^{\top} \ \tilde{z}_i.
	\end{equation}
	The transformed nodewise output logits $\tilde{z}_i$ are produced as follows: we first normalize the original logits $z_i$ to range $[0, 1]$, then sort the classwise logits for each node. This makes the linear layers $\phi^h$ focus on the general logit distributions rather than class predictions. A similar idea has been explored by \citet{rahimi2020intraorder}, where they show that intra order-preserving functions improve the model calibration. Here we find out that this sorting-based transformation helps GATS to learn useful representations of nodewise contributions $\tau^h_i$. (c.f.\ Section~\ref{subsec:exp_ablation}.)
	
	\paragraph{Defining $\alpha_{i,j}$.} The attention coefficients $\alpha_{i,j}$ are defined based on the neighbor similarity, which is determined by the inner product between rescaled nodewise logits $z_i / \gamma_i$ and $z_j / \gamma_j$. Inspired by \citet{velivckovic2018gat}, we compute the attention coefficients $\alpha_{i,j}$ as follows:
	\begin{equation}
	(\alpha_{i,j})_{j \in \hat{n}(i)} = \softmax_{j \in \hat{n}(i)} \Big( \leakyrelu \Big( \frac{1}{\gamma_i\ \gamma_j} z_i^{\top}\ z_j \Big) \Big).
	\label{eq:gats_alpha}
	\end{equation}
	
	\subsection{Calibration properties of GATS}
	
	\citet{zhang2020mixnmatch} propose three desiderata for calibration methods: accuracy-preserving, data-efficient, and expressive. GATS fulfills all of them:
	\begin{itemize}
		\item GATS is accuracy-preserving: since all node predictions are scaled by inverse temperatures $1 / T_i$ which are positive scalars, the order of output logits is preserved;
		\item GATS is data-efficient. It is a parametric calibration model with $C\cdot H+4$ learnable parameters ($T_0, \omega, \gamma_t, \gamma_n, (\theta^h)_{1 \leq h \leq H}$). According to \citet{zhang2020mixnmatch}, parametric methods are already data-efficient;
		\item GATS is expressive, as it produces nodewise temperatures $T_i$ adapted to the graph structure.
	\end{itemize}
	Experiments in Section~\ref{subsec:exp_gats_dataefficient} empirically confirm the data-efficiency and expressivity of GATS.

	\subsection{Comparison with CaGCN}
	
	It is interesting to compare our proposed GATS approach to the CaGCN method proposed by \citet{wang2021cagcn}, since both approaches aim at calibrating GNNs, and both make use of the graph structure to produce nodewise temperatures. While CaGCN uses a GCN to generate nodewise temperatures straightforwardly, GATS uses an attention mechanism which differentiates the influence from various neighbors. GATS also integrates a series of careful designs following the insights from the study in Section~\ref{sec:gnn_calib_factors}. Experiments in Section~\ref{subsec:expc_comparison} shows that GATS tends to produce better calibration results compared to CaGCN.

	\section{Experiments} \label{sec:experiment}
	
	To evaluate the performance of GATS on GNN calibration and understand the effects of its designs, we conduct a series of experiments for baseline comparison and ablation study.
	We use two representative GNNs: GCN and GAT, which are trained on the seven aforementioned graph datasets plus a larger graph, CoraFull \citep{bojchevski2018citationfull}, for post-hoc calibration.
	We display the ECE results with $M=15$ bins.
	We use the expected calibration error (ECE) \citep{Naeini_2015,guo2017calibration} with $M=15$ bins as an evaluation metric, and follow an experimental protocol similar to \citet{kull2019dirichlet, 10.1214/17-EJS1338SI}:
	For all the experiments, we randomly split the labeled/unlabeled ($15\%$/$85\%$) data five times, and use three-fold internal cross-validation of the labeled data to train the GNNs and the calibrators. We also utilize five random initializations, resulting in 75 total runs for each experiment. We provide detailed experimental settings in Appendix~\ref{sec:exp}.

	\subsection{Performance comparison} \label{subsec:expc_comparison}
	
	
	We benchmark GATS against existing baselines on a variety of GNN calibration tasks. 
	We compare GATS with the following baselines:
	\begin{itemize}
		\item \textbf{Temperature scaling (TS)} \citep{guo2017calibration} simply uses a global temperature to scale the logits.
		\item \textbf{Vector scaling (VS)} \citep{guo2017calibration} scales the logits separately over the class dimension and additionally introduces a classwise bias for the recalibrated output logits.
		\item \textbf{Ensemble temperature scaling (ETS)} \citep{zhang2020mixnmatch} learns a mixture of uncalibrated, TS-calibrated, and uniform probabilistic outputs.
		\item \textbf{GCN as a calibration function (CaGCN)} \citep{wang2021cagcn} is specifically designed for calibrating GNNs. It uses a GCN on top to generate nodewise temperatures.
	\end{itemize}
	Additionally, we also report the ECEs of uncalibrated predictions as a reference. Among the above baselines, TS, VS, and ETS are calibration methods designed for standard classification cases and operate on nodes uniformly. CaGCN on the other hand performs separate nodewise adjustments and uses the graph structure, similar to our proposed GATS approach. 
	
	For the post-hoc calibration experiments, we fix the weight of the trained GNN backbones and adjust the parameters of the calibration methods on the validation set. Negative log-likelihood is chosen as the objective for the calibration process. We provide details of method configurations and calibration settings in Appendix~\ref{subsec:details_calib}. Table~\ref{tab:calib_comparison} summarizes the calibration results.
	
	\begin{table}[ht]
		\caption{GNN calibration results in terms of ECE (in percentage, lower is better) of GATS and other baseline methods on various graph datasets. Overall, GATS achieves state-of-the-art performance, getting the best results in most scenarios. Also, all best results are achieved by methods that consider the graph structure. This shows the need for dedicated methods to tackle GNN calibration.}
		\label{tab:calib_comparison}
		\vspace{7pt} 
		\centering
		\small
		\begin{tabular}{lccccccc}
			\toprule
			Dataset 
			&Model 
			&Uncal
			&TS
			&VS
			&ETS
			&CaGCN
			&GATS\\
			\midrule
			\multirow{2}{*}{Cora} & GCN &13.04$\pm$5.22 &3.92$\pm$1.29 &4.36$\pm$1.34 &3.79$\pm$1.35 &5.29$\pm$1.47 &\textbf{3.64$\pm$1.34}\\
			& GAT &23.31$\pm$1.81 &3.69$\pm$0.90 &3.30$\pm$1.12 &3.54$\pm$1.01 &4.09$\pm$1.06 &\textbf{3.18$\pm$0.90}\\
			\midrule
			\multirow{2}{*}{Citeseer} & GCN &10.66$\pm$5.92 &5.15$\pm$1.50 &4.92$\pm$1.44 &4.65$\pm$1.69 &6.86$\pm$1.41 &\textbf{4.43$\pm$1.30}\\
			& GAT &22.88$\pm$3.53 &4.74$\pm$1.47 &4.25$\pm$1.48 &4.11$\pm$1.64 &5.75$\pm$1.31 &\textbf{3.86$\pm$1.56}\\
			\midrule
			\multirow{2}{*}{Pubmed} & GCN &7.18$\pm$1.51 &1.26$\pm$0.28 &1.46$\pm$0.29 &1.24$\pm$0.30 &1.09$\pm$0.52 &\textbf{0.98$\pm$0.30}\\
			& GAT &12.32$\pm$0.80 &1.19$\pm$0.36 &1.00$\pm$0.32 &1.20$\pm$0.32 &\textbf{0.98$\pm$0.31} &1.03$\pm$0.32\\
			\midrule
			\multirow{2}{*}{Computers} & GCN &3.00$\pm$0.80 &2.65$\pm$0.57 &2.70$\pm$0.63 &2.58$\pm$0.70 &\textbf{1.72$\pm$0.53} &2.23$\pm$0.49 \\
			& GAT &1.88$\pm$0.82 &1.63$\pm$0.46 &1.67$\pm$0.52 &1.54$\pm$0.67 &2.03$\pm$0.80 &\textbf{1.39$\pm$0.39}\\
			\midrule
			\multirow{2}{*}{Photo} & GCN &2.24$\pm$1.03 &1.68$\pm$0.63 &1.75$\pm$0.63 &1.68$\pm$0.89 &1.99$\pm$0.56 &\textbf{1.51$\pm$0.52}\\
			& GAT &2.02$\pm$1.11 &1.61$\pm$0.63 &1.63$\pm$0.69 &1.67$\pm$0.73 &2.10$\pm$0.78  &\textbf{1.48$\pm$0.61}\\
			\midrule
			\multirow{2}{*}{CS} & GCN &1.65$\pm$0.92 &0.98$\pm$0.27 &0.96$\pm$0.30 &0.94$\pm$0.24 &2.27$\pm$1.07 &\textbf{0.88$\pm$0.30}\\
			& GAT &1.40$\pm$1.25 &0.93$\pm$0.34 &0.87$\pm$0.35 &0.88$\pm$0.33 &2.52$\pm$1.04 &\textbf{0.81$\pm$0.30}\\
			\midrule
			\multirow{2}{*}{Physics} & GCN &0.52$\pm$0.29 &0.51$\pm$0.19 &0.48$\pm$0.16 &0.52$\pm$0.19 &0.94$\pm$0.51 &\textbf{0.46$\pm$0.16}\\
			& GAT &0.45$\pm$0.21 &0.50$\pm$0.21 &0.52$\pm$0.20 &0.50$\pm$0.21 &1.17$\pm$0.42 &\textbf{0.42$\pm$0.14}\\
			\midrule
			\multirow{2}{*}{CoraFull}
			&GCN &6.50$\pm$1.26 &5.54$\pm$0.43 &5.76$\pm$0.42 &5.38$\pm$0.49 &5.86$\pm$2.52 &\textbf{3.76$\pm$0.74}\\
			&GAT &4.73$\pm$1.39 &4.00$\pm$0.50 &4.17$\pm$0.43 &3.89$\pm$0.56 &6.55$\pm$3.69 &\textbf{3.54$\pm$0.63}\\
			\bottomrule
		\end{tabular}
	\end{table}
	
	Overall, we observe that GATS consistently produces well-calibrated predictions for all graph datasets and GNN backbones. Except for the GAT model trained on Pubmed (3rd best) and the GCN model trained on Amazon Computers (2nd best), GATS achieves the highest calibration quality in all cases. 
	
	Also, it is interesting to see that for all cases the best result is achieved by methods which use the graph structure and produce adapted adjustments for different nodes. This demonstrates the necessity of designing calibration methods that address the unique challenges posed by GNN calibration.
	
	Although CaGCN can get the best results for Pubmed using GAT and Amazon Computers using GCN, we see that its performance is rather unstable for different scenarios, and sometimes it can even produce worse calibration results than the uncalibrated baseline. Using their proposed margin-based loss did not help in our settings. We suspect that CaGCN might have an overly complex architecture for the task, and it cannot differentiate neighborhood influences with the common normalized adjacency matrix. Our proposed GATS model does not have this issue. It has consistent and good calibration performance in all cases. 
	
	We also observe that the results tend to have high variations, since GNN backbones tend to predict highly varying results when trained with different initial weights and random splits \citep{shchur2018pitfalls}. We ensure the reliability of the results by averaging over a total of 75 runs with various initial weights and random splits for each case.

	\subsection{Data-efficiency and expressivity of GATS} \label{subsec:exp_gats_dataefficient}
	
	\begin{wrapfigure}[18]{r}{0.48\linewidth}
		\centering
		\includegraphics[width=0.72\linewidth]{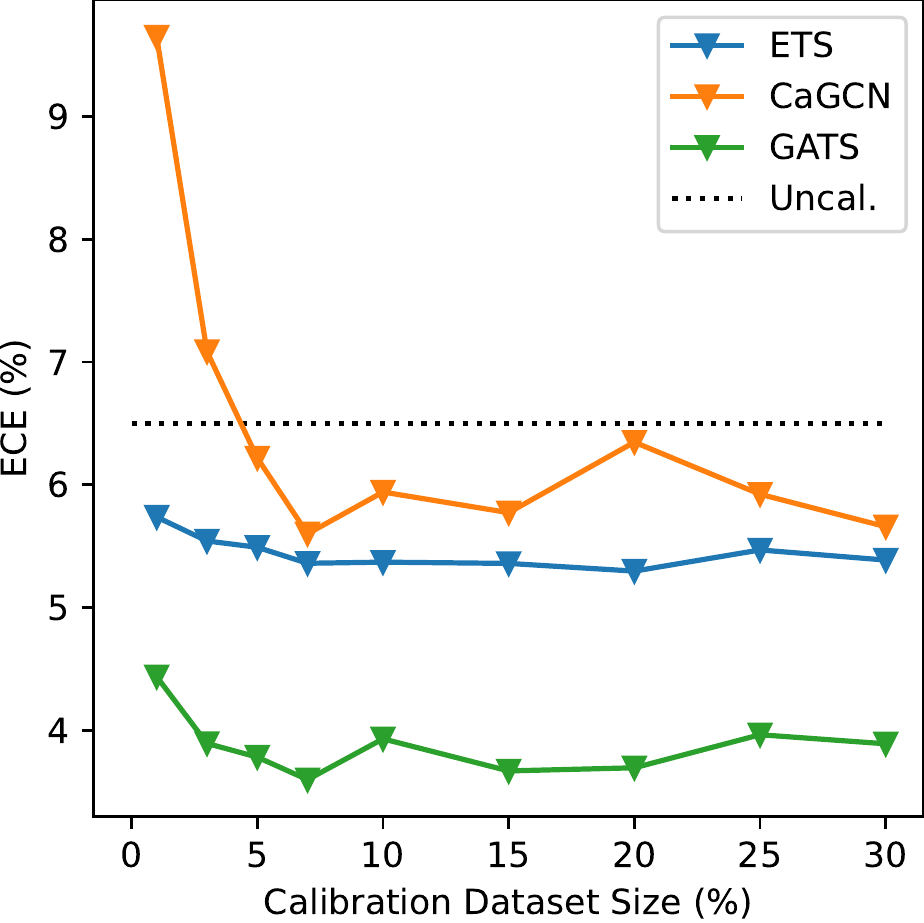}
		\caption{ECEs (in percentage) on CoraFull for ETS, CaGCN, and GATS using various amounts of calibration data. We see that GATS is data-efficient and expressive for GNN calibration.} \label{fig:data_efficiency_gcn}
	\end{wrapfigure}
	
	Furthermore, we analyze the data-efficiency and the expressivity of GATS for GNN calibration. For this we reuse the GNN models trained on the CoraFull dataset, and consider the influence of calibration sample size on the GNN calibration performance. For comparison we also report the corresponding results using ensemble temperature scaling and CaGCN. Figure~\ref{fig:data_efficiency_gcn} visualizes the results with GCN backbone. The results for GAT backbone are summarized in Appendix~\ref{sec:data_efficiency_gat}.
	
	Overall, we see that GATS is both data-efficient and expressive. It requires few calibration samples to get decent calibration performance. This is in contrast to CaGCN which needs more than $5\%$ of nodes for calibration to get acceptable results. Compared to ETS, GATS is more expressive and has a considerably lower calibration error for CoraFull, which is a large graph dataset.

	\subsection{Ablation study} \label{subsec:exp_ablation}
	
	To empirically analyze the effect of various GATS design choices, we conduct a series of ablation study experiments in this section. Overall, we notice that all designs are advantageous and removing any of them leads to a general decrease in performance.
	
	\begin{table}[ht]
		\caption{Ablation study results in terms of ECE (in percentage) for various GATS designs. Overall, all designs are beneficial and removing any of them leads to worse results in general.}
		\vspace{7pt} 
		\label{tab:ablation_designs}
		\centering
		\small
		\begin{tabular}{lccccccc}
			\toprule
			Dataset
			&Model 
			&w/o $T_0$
			&w/o $\gamma_i$
			&w/o $\delta \hat{c}_i$
			&w/o attention
			&w/o sorting
			&GATS\\
			\midrule
			\multirow{2}{*}{Cora} 
			&GCN &3.72$\pm$1.20 &3.80$\pm$1.51 &3.71$\pm$1.18 &3.63$\pm$1.48 &4.35$\pm$1.77 &3.64$\pm$1.34\\
			&GAT &3.25$\pm$1.00 &3.46$\pm$1.00 &3.24$\pm$0.89 &3.69$\pm$0.96 &4.18$\pm$1.70 &3.18$\pm$0.90\\\midrule
			\multirow{2}{*}{Citeseer}
			&GCN &5.50$\pm$1.76 &4.73$\pm$1.45 &4.49$\pm$1.30 &4.95$\pm$1.56 &5.87$\pm$1.99 &4.43$\pm$1.30\\
			&GAT &3.56$\pm$1.73 &4.39$\pm$1.46 &3.87$\pm$1.55 &4.81$\pm$1.54 &4.99$\pm$2.34 &3.86$\pm$1.56\\\midrule
			\multirow{2}{*}{Photo}
			&GCN &2.20$\pm$0.88 &1.60$\pm$0.64 &1.54$\pm$0.52 &1.59$\pm$0.67 &1.68$\pm$0.61 &1.51$\pm$0.52\\
			&GAT &2.37$\pm$1.01 &1.53$\pm$0.63 &1.47$\pm$0.62 &1.62$\pm$0.66 &1.77$\pm$0.70 &1.48$\pm$0.61\\
			\bottomrule
		\end{tabular}
	\end{table}
	
	\paragraph{Effect of global bias $T_0$.} 
	We consider the effect of global bias $T_0$ by comparing to a GATS variant without it (i.e.,\ setting $T_0=0$ in Eq.~\eqref{eq:gats_t}). Its results are collected in column ``w/o $T_0$'' of Table~\ref{tab:ablation_designs}. Overall we see that a learnable bias $T_0$ is beneficial in most cases.
	
	\paragraph{Effect of nodewise scaling factor $\gamma_i$.}
	The nodewise scaling factors $\gamma_i$ provide custom adjustment for training nodes and their neighbors. Removing its influence can be done by fixing all $\gamma_i$ to one in Eqs.~\eqref{eq:gats_t} and \eqref{eq:gats_alpha}. The results of the variant without $\gamma_i$ are recorded in column ``w/o $\gamma_i$'' of Table~\ref{tab:ablation_designs}. They are worse in general, suggesting that nodewise scaling factors $\gamma_i$ are indeed helpful.
	
	\paragraph{Effect of relative confidence level $\delta \hat{c}_i$.}
	To evaluate the impact of introducing nodewise relative confidence level $\delta \hat{c}_i$, we create a GATS variant where $\omega$ in Eq.~\eqref{eq:gats_t} is fixed to zero, which effectively removes the influence of the relative confidence level $\delta \hat{c}_i$. In column ``w/o $\delta \hat{c}_i$'' of Table~\ref{tab:ablation_designs} we have the results corresponding to this variant, which is in general slightly worse than the standard GATS which includes $\delta \hat{c}_i$.

	\paragraph{Effect of attention-based aggregation.}
	To understand the role played by the attention mechanism, we create a GATS variant which completely removes the attention related term ($\sum_{j \in \hat{n}(i)} \alpha_{i,j} \ \gamma_j\ \tau^h_j$) from  Eq.~\eqref{eq:gats_t}. Its performance is shown in column ``w/o attention'' of Table~\ref{tab:ablation_designs}. Again, we observe that removing the attention-based component tends to worsen the calibration performance.
	
	\paragraph{Effect of logit sorting.} GATS uses normalized and sorted logits $\tilde{z}_i$ as input to generate nodewise temperature contributions $\tau_i^h$. And we find that this sorting transform is essential for learning good representations of $\tau_i^h$. In column ``w/o sorting'' of Table~\ref{tab:ablation_designs} we have the results for GATS variants which uses the logits without sorting to compute $\tau_i^h$. And we observe that this deteriorates the calibration performance.
	
	\begin{table}[ht]
		\caption{Calibration results (ECE in percentage) of GATS models with different numbers of attention heads. 8 attention heads are sufficient for optimal GNN calibration results.}
		\label{tab:ablation_multihead}
		\vspace{7pt} 
		\centering
		\small
		\begin{tabular}{lcccccc}
			\toprule
			\multirow{2}{*}{Dataset}
			&\multirow{2}{*}{Model} 
			& \multicolumn{5}{c}{Number of Heads}\\
			&
			&1 
			&2
			&4
			&8
			&16\\
			\midrule
			\multirow{2}{*}{Cora} 
			&GCN &3.95$\pm$1.45 &3.75$\pm$1.42 &3.79$\pm$1.45 &3.66$\pm$1.33 &3.50$\pm$1.25\\
			&GAT &3.48$\pm$1.22 &3.49$\pm$1.18 &3.43$\pm$1.15 &3.20$\pm$0.90 &3.31$\pm$0.74\\\midrule
			\multirow{2}{*}{Citeseer}
			&GCN &4.74$\pm$1.43 &4.77$\pm$1.66 &4.70$\pm$1.52 &4.43$\pm$1.30 &4.42$\pm$1.04\\
			&GAT &4.53$\pm$1.66 &4.57$\pm$1.55 &4.20$\pm$1.50 &3.86$\pm$1.56 &4.20$\pm$1.54\\\midrule
			\multirow{2}{*}{Photo}
			&GCN &1.51$\pm$0.54 &1.54$\pm$0.59 &1.52$\pm$0.58 &1.51$\pm$0.52 &1.50$\pm$0.51\\
			&GAT &1.52$\pm$0.61 &1.52$\pm$0.65 &1.53$\pm$0.60 &1.48$\pm$0.61 &1.50$\pm$0.60\\
			\bottomrule
		\end{tabular}
	\end{table}%
	
	\paragraph{Effect of attention head count $H$.}
	Finally, we analyze the influence of multi-head count $H$ on the GNN calibration results. For this we run a series of experiments using GATS models with 1, 2, 4, 8, and 16 attention heads. The results are collected in Table~\ref{tab:ablation_multihead}. We see that for GCN backbones, GATS models with more attention heads tend to get better results. However, for GAT backbones, using 16 heads results in worse performance compared to 8 heads. Accounting also for the fact that doubling the attention head count effectively doubles the computational requirements, 8 attention heads is a decent general setting for GATS.

	\section{Conclusion} \label{sec:conclusion}
	
	In this work, we tackle the GNN calibration problem. We conduct a systematic study to analyze the calibration properties of GNNs predictions. 
	Our study reveals five influential factors and manifests the unique challenges raised by GNN calibration.
	Based on the insights from our studies, we propose a novel calibrator, GATS, which accounts for the identified factors and is tailored for calibrating GNNs.
	GATS is accuracy-preserving, data-efficient, and expressive at the same time.
	Our experiments demonstrate that GATS achieves state-of-the-art performance for GNN calibration on various graph datasets and for different GNN backbones.
	
	Our work focuses on the node classification tasks for assortative graphs, where neighbors tend to agree with each other. 
	It is thus important to realize that the validity of the conclusions from Section~\ref{sec:gnn_calib_factors} is limited to the assortative case, and might no longer hold for disassortative graphs \citep{zhu2020beyond, Pei2020Geom-GCN}. 
	It can be an interesting future work to conduct similar studies for GNN calibration in the heterophilous case, especially when more established GNN architectures are available. 
	More generally, devising calibration methodologies for other graph learning tasks such as link prediction \citep{zhang2018link} and graph classification \citep{Errica2020A} could also be an interesting direction for future research.
	
	\begin{ack}
		This work was supported by the Munich Center for Machine Learning (MCML) and by the ERC Advanced Grant SIMULACRON. The authors would like to thank Nikita Araslanov for proofreading and helpful discussions, as well as the anonymous reviewers for their constructive feedback.
	\end{ack}

	{\small
		\bibliographystyle{abbrvnat}
		\bibliography{ref}
	}

	\section*{Checklist}

	\begin{enumerate}

		\item For all authors...
		\begin{enumerate}
			\item Do the main claims made in the abstract and introduction accurately reflect the paper's contributions and scope?
			\answerYes{}
			\item Did you describe the limitations of your work?
			\answerYes{} c.f.\ Section~\ref{sec:conclusion}.
			\item Did you discuss any potential negative societal impacts of your work?
			\answerNA{}
			\item Have you read the ethics review guidelines and ensured that your paper conforms to them?
			\answerYes{}
		\end{enumerate}

		\item If you are including theoretical results...
		\begin{enumerate}
			\item Did you state the full set of assumptions of all theoretical results?
			\answerNA{}
			\item Did you include complete proofs of all theoretical results?
			\answerNA{}
		\end{enumerate}

		\item If you ran experiments...
		\begin{enumerate}
			\item Did you include the code, data, and instructions needed to reproduce the main experimental results (either in the supplemental material or as a URL)?
			\answerYes{}
			\item Did you specify all the training details (e.g., data splits, hyperparameters, how they were chosen)?
			\answerYes{}
			\item Did you report error bars (e.g., with respect to the random seed after running experiments multiple times)?
			\answerYes{}
			\item Did you include the total amount of compute and the type of resources used (e.g., type of GPUs, internal cluster, or cloud provider)?
			\answerYes{}
		\end{enumerate}

		\item If you are using existing assets (e.g., code, data, models) or curating/releasing new assets...
		\begin{enumerate}
			\item If your work uses existing assets, did you cite the creators?
			\answerYes{}
			\item Did you mention the license of the assets?
			\answerNA{}
			\item Did you include any new assets either in the supplemental material or as a URL?
			\answerYes{}
			\item Did you discuss whether and how consent was obtained from people whose data you're using/curating?
			\answerNA{}
			\item Did you discuss whether the data you are using/curating contains personally identifiable information or offensive content?
			\answerNA{}
		\end{enumerate}

		\item If you used crowdsourcing or conducted research with human subjects...
		\begin{enumerate}
			\item Did you include the full text of instructions given to participants and screenshots, if applicable?
			\answerNA{}
			\item Did you describe any potential participant risks, with links to Institutional Review Board (IRB) approvals, if applicable?
			\answerNA{}
			\item Did you include the estimated hourly wage paid to participants and the total amount spent on participant compensation?
			\answerNA{}
		\end{enumerate}

	\end{enumerate}

	\newpage 
	\appendix

	\section{Experimental settings}\label{sec:exp}
	
	\subsection{Dataset statistics} \label{subsec:dataset_stat}
	
	We consider eight real-world graph datasets including citation networks Cora \citep{sen2008coraciteseer}, Citeseer \citep{sen2008coraciteseer}, Pubmed \citep{namata2012pubmed}, Coauthor CS \citep{shchur2018pitfalls}, Coauthor Physics \citep{shchur2018pitfalls}, and CoraFull \citep{bojchevski2018citationfull} together with Amazon co-purchase networks Computers \citep{shchur2018pitfalls} and Photo \citep{shchur2018pitfalls}.
	Table \ref{tab:data_stat} summarizes their statistics.
	
	\begin{table}[ht]
		\caption{Benchmark dataset statistics.}
		\label{tab:data_stat}
		\vspace{7pt} 
		\centering
		\small
		\begin{tabular}{lcccccccc}
			\toprule
			&Cora
			&Citeseer
			&Pubmed
			&Computers
			&Photo
			&CS
			&Physics
			&CoraFull\\
			\midrule
			Nodes &2,708 &3,327 &19,717 &13,752 &7,650 &18,333 &34,493 &19,793\\
			Edges &10,556 &9,104 &88,648 &491,722 &238,162 &163,788 &495,924 &126,842\\
			Features &1,433 &3,703 &500 &767 &745 &6,805 &8,415 &8,710\\
			Classes &7 &6 &3 &10 &8 &15 &5 &70\\
			Homophily &82.52\% &70.62\% &79.24\% &78.53\% &83.65\% &83.20\% &91.53\% &58.61\%\\
			\bottomrule
		\end{tabular}
	\end{table}
	
	We report the homophily index proposed by \citet{Pei2020Geom-GCN}, which provides a global view of the neighborhood similarity for a graph. Given a graph $\mathcal{G} = (\mathcal{V} , \mathcal{E})$, the homophily is defined as
	\begin{equation}
	    \mathcal{H}(\mathcal{G})=\frac{1}{|\mathcal{V}|}\sum_{i\in \mathcal{V}}\frac{\text{Number of node }i\text{'s neighbors who have the same label as } i}{\text{Number of }i\text{'s neighbors}}.
	 \label{eq:homo}
	\end{equation}
	
	\subsection{Details of model training setup} \label{subsec:details_train}
	We follow the setting of \citet{shchur2018pitfalls} to define GCN \citep{kipf2017gcn} and GAT \citep{velivckovic2018gat} models.
	Both models consist of 2 layers and the hidden dimension is fixed to 64.
	For the multi-head layer in GAT, the number of attention heads is fixed to 8 with 8 hidden units per head.
	We implement the models and training pipelines in PyTorch \citep{pytorch} and PyTorch Geometric \citep{pyg}.
	All models are trained for a maximum of 2000 epochs, using early stopping with a patience of 100 epochs.
	We choose Adam \citep{kingma2014method} as the optimizer with initial learning rate 0.01. We add a weight decay of 5e-4 for Cora, Citeseer, and Pubmed, and 0 for the rest.
	
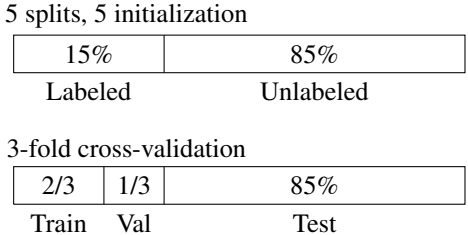
\begin{figure}[ht!]
    \centering
    	\begin{tikzpicture}
          \node at (1.5,0.75){5 splits, 5 initialization};   
           \node at (1,0.25){15\%};
           \node at (4,0.25){85\%};
           \node at (1,-0.25){Labeled};
           \node at (4,-0.25){Unlabeled};
           
           \draw (0,0) -- (6,0) -- (6,0.5) -- (0,0.5) -- (0,0);
           \draw (2,0) -- (2,0.5);
          \node at (1.5,-1){3-fold cross-validation};   
           \draw (0,-1.75) -- (6,-1.75) -- (6,-1.25) -- (0,-1.25) -- (0,-1.75);
           \draw (2,-1.75) -- (2,-1.25); 
           \draw (1.2,-1.75) -- (1.2,-1.25); 
           \node at (0.6,-1.5){2/3};
           \node at (1.6,-1.5){1/3};
           \node at (4,-1.5){85\%};
           \node at (0.6,-2){Train};
           \node at (1.6,-2){Val};
           \node at (4,-2){Test};
        \end{tikzpicture}
      \caption{Data partition schema for graph datasets.}\label{fig:split}
    \end{figure}

	We use stratified sampling to randomly select 15\% of the nodes as observed set, mask out the output labels of the rest 85\% of the nodes for test prediction, and ensure that the nodes with the same label are split proportionally.
	Following \citet{kull2019dirichlet, 10.1214/17-EJS1338SI}, we further divide the labeled set with three-fold cross-validation. 
	The bigger portions (10\%) are used as training sets and the rest (5\%) are used as validation sets. 
	The GNN models (GCN and GAT) are trained on the training set, then used to predict the masked-out test set.
	Figure~\ref{fig:split} illustrates the aforementioned data partition in our experiments.
	In total, we use 5 random data splits, three-fold cross-validation for each split, and 5 random model initializations per data partition, resulting in 75 total runs for each experiment.
	
	\subsection{Details of model calibration setup} \label{subsec:details_calib}
	
	We compare GATS with four scaling-based calibrators: temperature scaling (TS) \citep{guo2017calibration}, vector scaling (VS) \citep{guo2017calibration}, ensemble temperature scaling (ETS) \citep{zhang2020mixnmatch} , CaGCN \citep{wang2021cagcn}.
	The calibrators are trained on the validation set using the negative log-likelihood (NLL) loss and validated on the training set for early stopping and hyperparameter search.
	The optimizer configuration and the training schedule are the same as Section~\ref{subsec:details_train}.
	We observe TS and VS using Adam with weight decay 0 achieves better performance than using L-BFGS \citep{lbfgs} in the original implementation\footnote{\url{https://github.com/gpleiss/temperature_scaling}}.
    For ETS, we follow the official implementation that uses Sequential Least SQuares Programming (SLSQP) \citep{sqlsp}.
    For CaGCN, we use a two-layer GCN with 16 hidden units and choose the hyperparameters following the original paper.
    For GATS, we utilize one message passing layer and
    initialize $T_0$, $\gamma_t$, and $\gamma_n$ to 1 and $\omega$ to 0. 
    We find the best hyperparameter using cross-validation.
    Table~\ref{tab:hyper_search} shows the search space for GATS hyperparameters.
	
	\begin{table}[ht]
	    \caption{Hyperparameter search space for GATS.}
	    \label{tab:hyper_search}
	    \vspace{7pt}
	    \centering
		\begin{tabular}{lc}
			\toprule
			Hyperparameter
			&Search space\\
			\midrule
			Weight decay &0, 1e-3, 5e-3, 1e-2, 5e-2, 1e-1, 2e-1, 3e-1\\
			Initial $T_0$ &1, 1.5 \\
			\bottomrule
		\end{tabular}
	\end{table}
		
	\section{Additional plots}
	
	Here we include additional plots which shows the corresponding factors influencing the calibration of GAT models (c.f.\ Section~\ref{sec:gnn_calib_factors}). Overall, we reach the same conclusions as the GCN case.
	
	\FloatBarrier
	\subsection{General under-confident tendency for GAT} \label{subsec:miscalib_tendency}
	
\begin{figure}[ht!]
		\centering
		\setlength{\tabcolsep}{0.04\textwidth} 
		\begin{tabular}{cccc}
			\includegraphics[width=0.16\textwidth]{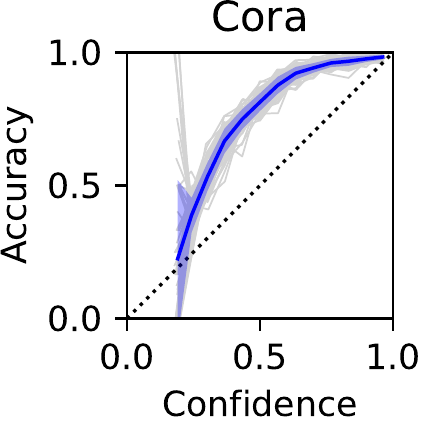} &
			\includegraphics[width=0.16\textwidth]{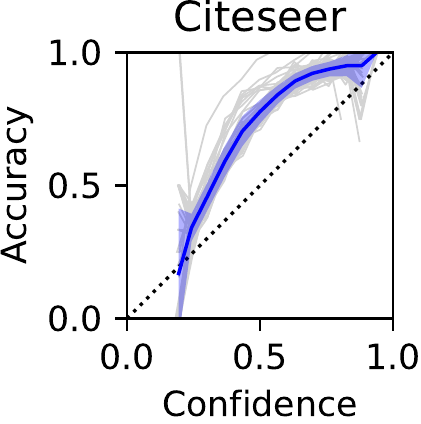} &
			\includegraphics[width=0.16\textwidth]{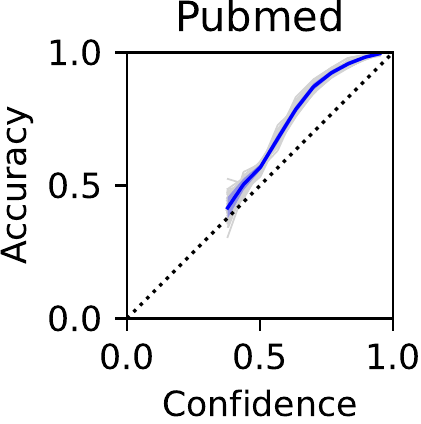} &
			\includegraphics[width=0.16\textwidth]{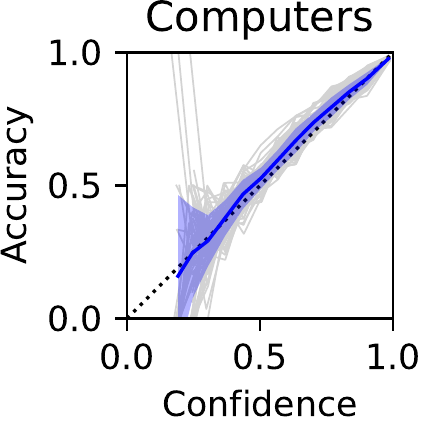} \\
			\includegraphics[width=0.16\textwidth]{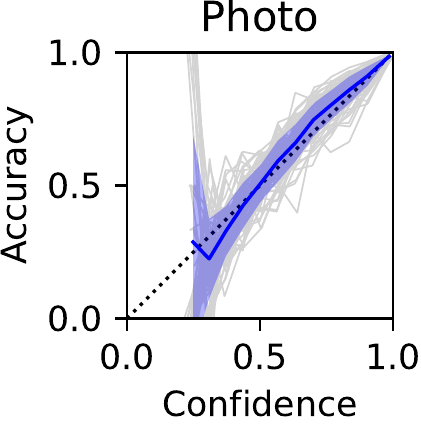} &
			\includegraphics[width=0.16\textwidth]{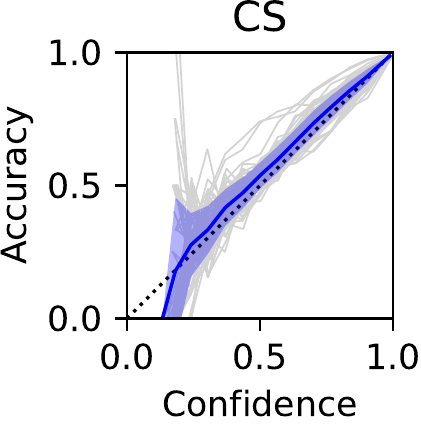} &
			\includegraphics[width=0.16\textwidth]{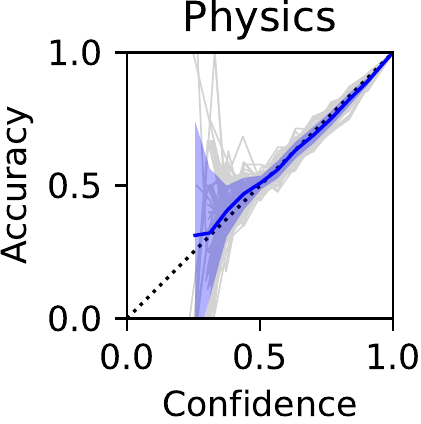} &
			\includegraphics[width=0.16\textwidth]{figure/reliability_uncalib_cifar10_resnet20.pdf}
		\end{tabular}		
		\caption{Reliability diagrams of GAT models trained on various graph datasets.} \label{fig:miscalib_tendency_gat}
	\end{figure}

	Figure~\ref{fig:miscalib_tendency_gat} summarizes the GAT results. 
	We see a general tendency of under-confident predictions (plots above the diagonal) except for the Physics dataset, which differs from the overconfident behavior of multiclass image classification using CNNs.
	
	\FloatBarrier
	\subsection{Diversity of node distributions for GAT} \label{subsec:miscalib_nodestat_gat}
	
\begin{figure}[ht!]
		\centering
		\begin{subfigure}[c]{0.16\textwidth}
			\includegraphics[width=\textwidth]{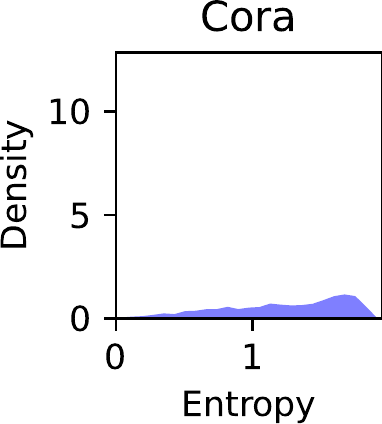}
		\end{subfigure}
		\hspace{0.06\textwidth}
		\begin{subfigure}[c]{0.16\textwidth}
			\includegraphics[width=\textwidth]{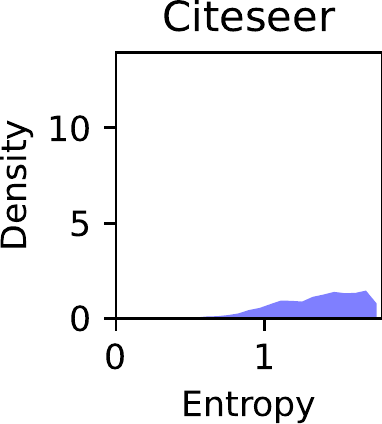}
		\end{subfigure}
		\hspace{0.06\textwidth}
		\begin{subfigure}[c]{0.16\textwidth}
			\includegraphics[width=\textwidth]{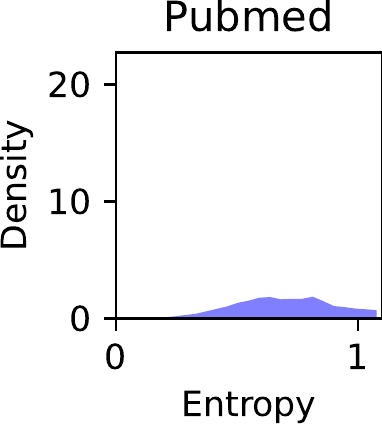}
		\end{subfigure}
		\hspace{0.06\textwidth}
		\begin{subfigure}[c]{0.16\textwidth}
			\includegraphics[width=\textwidth]{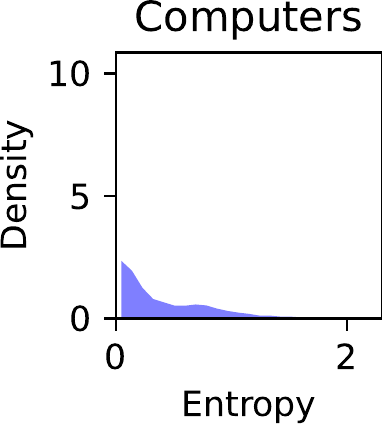}
		\end{subfigure}\\
		\begin{subfigure}[c]{0.16\textwidth}
			\includegraphics[width=\textwidth]{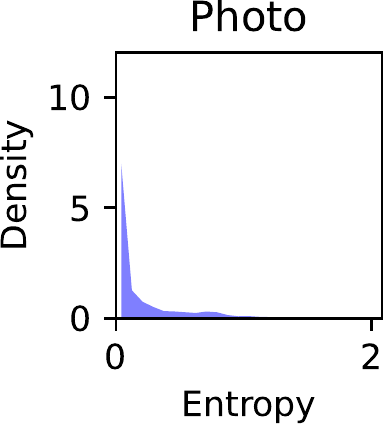}
		\end{subfigure}
		\hspace{0.06\textwidth}
		\begin{subfigure}[c]{0.16\textwidth}
			\includegraphics[width=\textwidth]{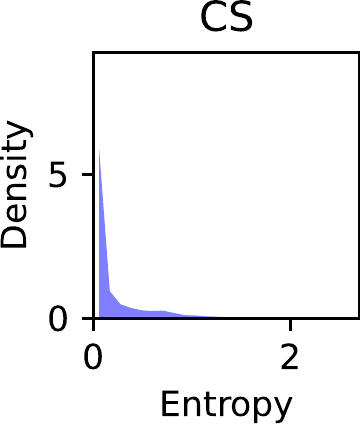}
		\end{subfigure}
		\hspace{0.06\textwidth}
		\begin{subfigure}[c]{0.16\textwidth}
			\includegraphics[width=\textwidth]{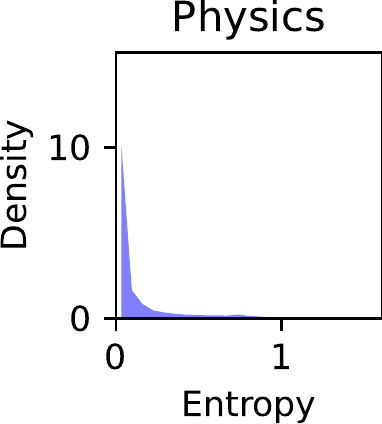}
		\end{subfigure}
		\hspace{0.06\textwidth}
		\begin{subfigure}[c]{0.16\textwidth}
			\includegraphics[width=\textwidth]{figure/entropy_hist_cifar10_resnet20.pdf}
		\end{subfigure}
		\caption{Entropy distributions of GAT predictions on graph datasets.} \label{fig:miscalib_nodestat_gat}
	\end{figure}

	Figure~\ref{fig:miscalib_nodestat_gat} shows the GAT results. Compared to the standard classification case, predictions of GAT also tend to be more spread out.
	
	\FloatBarrier
	\subsection{Effect of distance to training nodes for GAT} \label{subsec:miscalib_traindist_gat}
	
\begin{figure}[ht!]
		\centering
		\begin{subfigure}[c]{0.135\textwidth}
			\includegraphics[width=\textwidth]{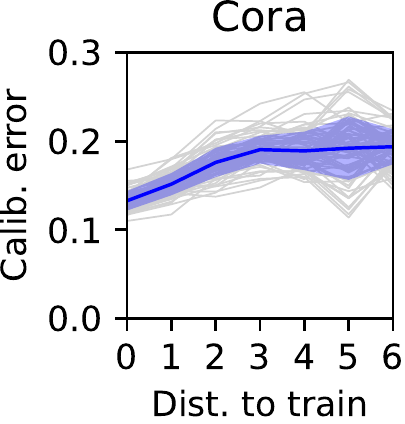}
		\end{subfigure}
		\begin{subfigure}[c]{0.135\textwidth}
			\includegraphics[width=\textwidth]{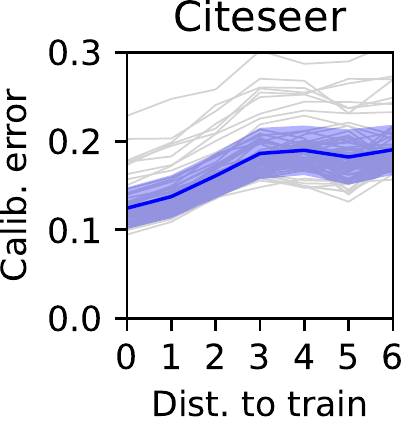}
		\end{subfigure}
		\begin{subfigure}[c]{0.135\textwidth}
			\includegraphics[width=\textwidth]{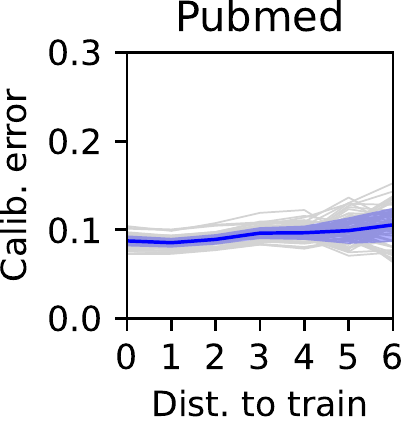}
		\end{subfigure}
		\begin{subfigure}[c]{0.135\textwidth}
			\includegraphics[width=\textwidth]{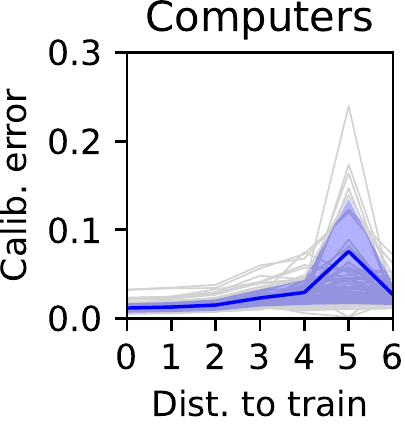}
		\end{subfigure}
		\begin{subfigure}[c]{0.135\textwidth}
			\includegraphics[width=\textwidth]{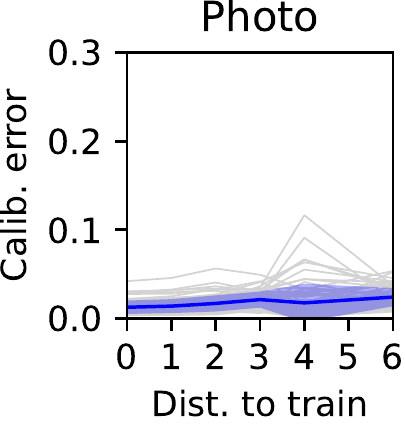}
		\end{subfigure}
		\begin{subfigure}[c]{0.135\textwidth}
			\includegraphics[width=\textwidth]{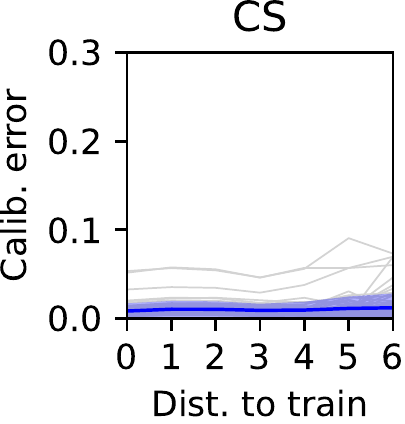}
		\end{subfigure}
		\begin{subfigure}[c]{0.135\textwidth}
			\includegraphics[width=\textwidth]{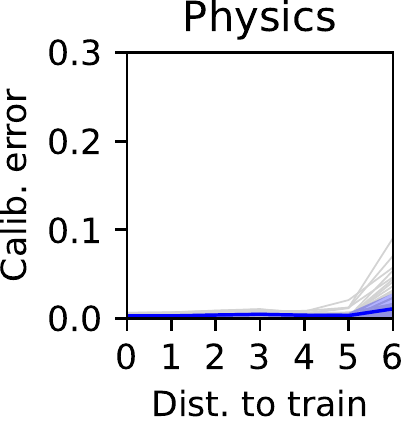}
		\end{subfigure}
		\hspace{0.2\textwidth}
		\caption{Nodewise calibration error of GAT results w.r.t\ the minimum distance to training nodes.} \label{fig:miscalib_traindist_gat}
	\end{figure}

	GAT result are shown in Figure~\ref{fig:miscalib_traindist_gat}. We also see that training nodes and their neighbors tend to be better calibrated.
	
\FloatBarrier
\subsection{Relative confidence level for GAT} \label{subsec:miscalib_diffconf_gat}

\begin{figure}[ht!]
\centering
\begin{subfigure}[c]{0.135\textwidth}
	\includegraphics[width=\textwidth]{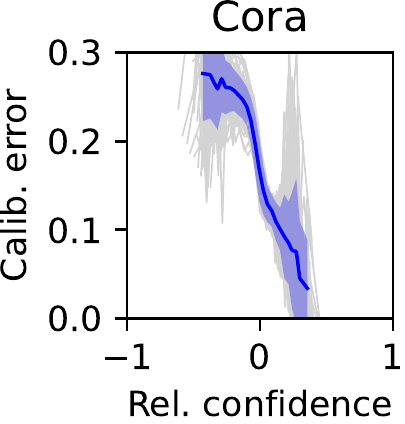}
\end{subfigure}
\begin{subfigure}[c]{0.135\textwidth}
	\includegraphics[width=\textwidth]{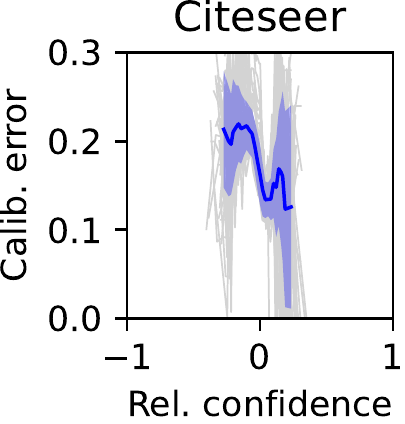}
\end{subfigure}
\begin{subfigure}[c]{0.135\textwidth}
	\includegraphics[width=\textwidth]{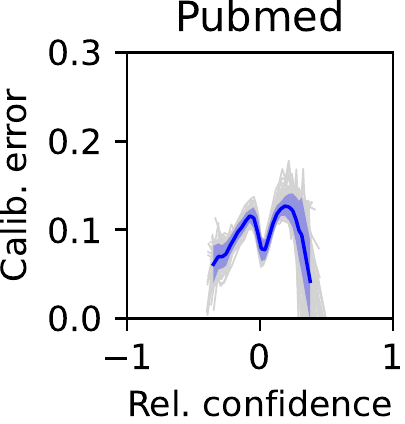}
\end{subfigure}
\begin{subfigure}[c]{0.135\textwidth}
	\includegraphics[width=\textwidth]{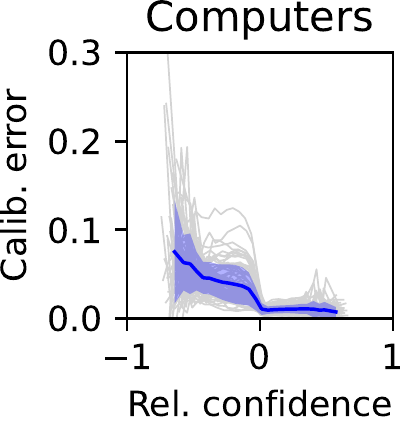}
\end{subfigure}
\begin{subfigure}[c]{0.135\textwidth}
	\includegraphics[width=\textwidth]{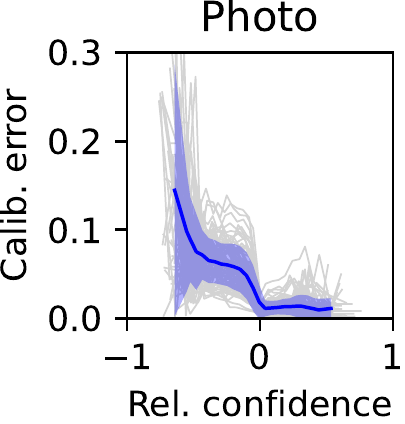}
\end{subfigure}
\begin{subfigure}[c]{0.135\textwidth}
	\includegraphics[width=\textwidth]{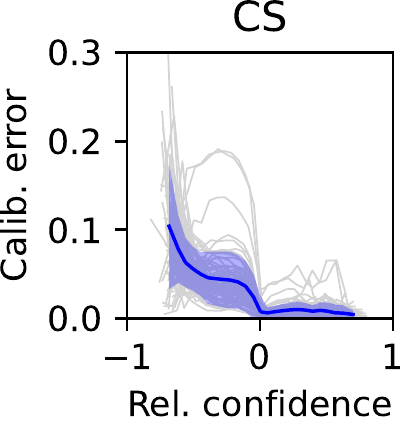}
\end{subfigure}
\begin{subfigure}[c]{0.135\textwidth}
	\includegraphics[width=\textwidth]{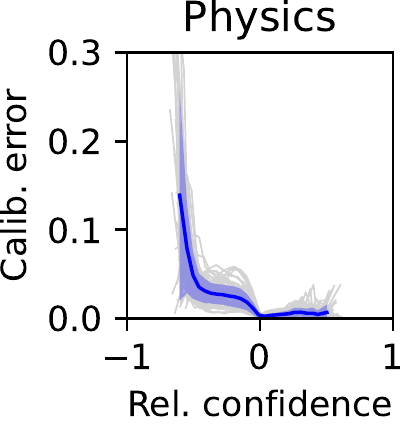}
\end{subfigure}
\hspace{0.2\textwidth}
\caption{Nodewise calibration error of GAT results depending on the relative confidence level.} \label{fig:miscalib_diffconf_gat}
\end{figure}

The plots for the GAT case are shown in Figure~\ref{fig:miscalib_diffconf_gat}. Similar to the GCN case, we observe that nodes which are less confident than their neighbors tend to be poorly calibrated and it is in general desirable to have a comparable confidence level w.r.t.\ the neighbors.

	\FloatBarrier
	\subsection{Neighborhood similarity for GAT} \label{subsec:miscalib_neighbor_gat}
	
\begin{figure}[ht!]
		\centering
		\begin{subfigure}[c]{0.135\textwidth}
			\includegraphics[width=\textwidth]{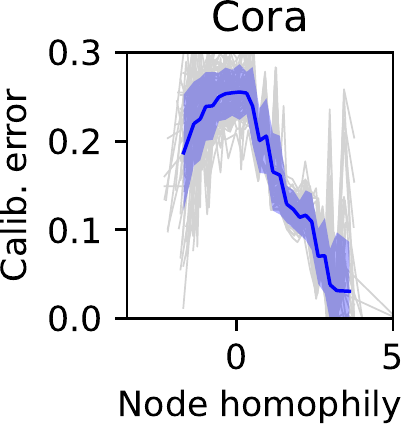}
		\end{subfigure}
		\begin{subfigure}[c]{0.135\textwidth}
			\includegraphics[width=\textwidth]{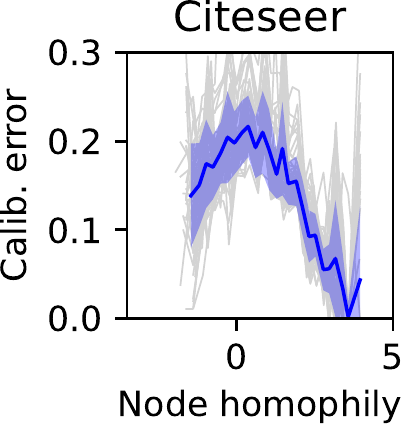}
		\end{subfigure}
		\begin{subfigure}[c]{0.135\textwidth}
			\includegraphics[width=\textwidth]{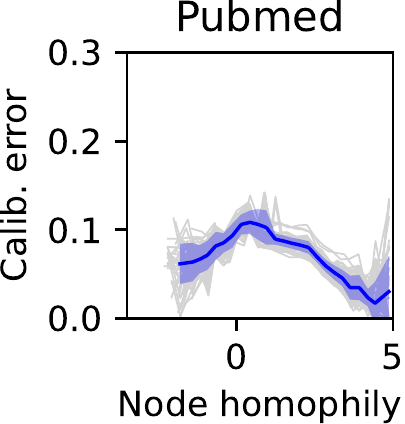}
		\end{subfigure}
		\begin{subfigure}[c]{0.135\textwidth}
			\includegraphics[width=\textwidth]{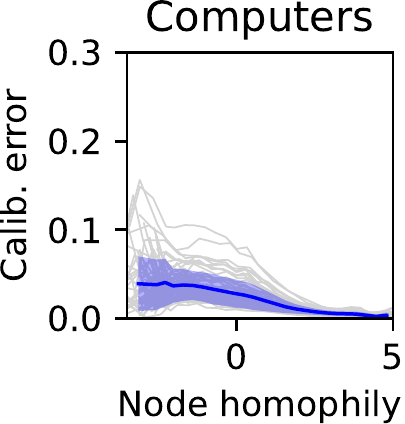}
		\end{subfigure}
		\begin{subfigure}[c]{0.135\textwidth}
			\includegraphics[width=\textwidth]{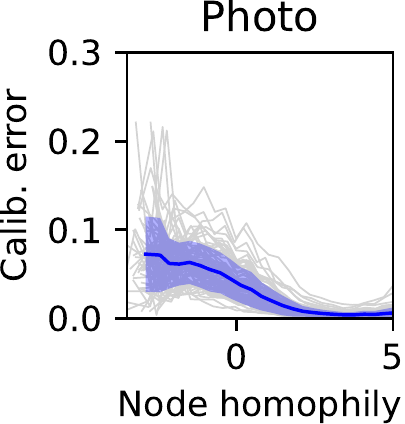}
		\end{subfigure}
		\begin{subfigure}[c]{0.135\textwidth}
			\includegraphics[width=\textwidth]{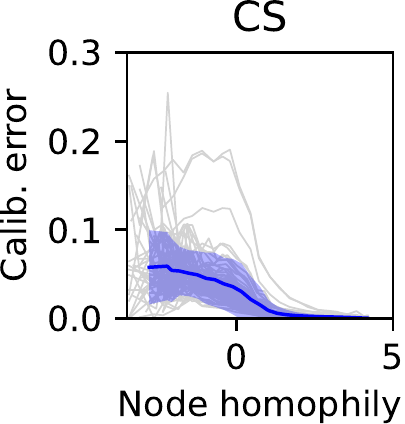}
		\end{subfigure}
		\begin{subfigure}[c]{0.135\textwidth}
			\includegraphics[width=\textwidth]{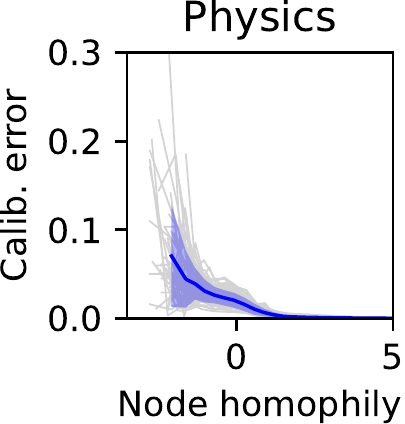}
		\end{subfigure}
		\hspace{0.2\textwidth}
		\caption{Nodewise calibration error of GCN results depending on the node homophily.} \label{fig:miscalib_neighbor_gat}
	\end{figure}

	Figure~\ref{fig:miscalib_neighbor_gat} shows tha GAT results. Analogue to the GCN case, nodes with strongly agreeing neighbors tend to have lower calibration errors.
	
	\section{Additional calibration results} \label{sec:exp_calib_supp}
	
	This section includes supplementary results with additional metrics and calibration methods.
	\FloatBarrier
	\subsection{Results using additional metrics} \label{sec:exp_calib_nll_brier}
	
	\paragraph{Variants of ECE}
	Even though ECE \citep{Naeini_2015,guo2017calibration} is the most commonly used metric for measuring calibration, it has some limitations: 
	(1) ECE only considers top-1 probabilistic output but can not reflect classwise calibration.
	(2) The binning-based estimator \citep{Naeini_2015} is dependent on the choice of binning scheme.
	To alleviate these disadvantages, we evaluate the trained calibrators using the classwise-ECE \citep{kull2019dirichlet} and the kernel density estimation (KDE) based KDE-ECE \citep{zhang2020mixnmatch} which is a binning-free metric:
	\begin{itemize}
	    \item \textbf{Classwise-ECE} measures the gap between the classwise average prediction $\text{conf}(B_{m,k})$ and the actual frequency of that class $\text{freq}(B_{m,k})$ in $M$ equally spaced bins across all classes $k \in K$:
	    \begin{align}
	    \text{Classwise-ECE} = \frac{1}{K}\sum_{k}^{K}\sum_{m}^{M}&\frac{|B_{m,k}|}{|\mathcal{N}|}|\text{freq}(B_{m,k}) - \text{conf}(B_{m,k})| \\
	    \text{freq}(B_{m,k})&=\frac{1}{|B_{m,k}|}\sum_{i\in B_{m,k}} \mathbf{1}(y_i=k) \\
	    \text{conf}(B_{m,k})&=\frac{1}{|B_{m,k}|}\sum_{i\in B_{m,k}}\hat{p}_{i,k}
    	\end{align}
	where $\hat{p}_{i,k}$ denotes the probability of predicting class $k$ for sample $i$ in bin $B_{m,k}$.
	We compute classwise-ECE using $M=15$ bins in our implementation.
	
	\item Instead of using a binning-based estimator, \textbf{KDE-ECE} uses a kernel function $K_h$ to estimate the accuracy $\tilde{\pi}(c)$ given confidence prediction $c$ and the marginal density function $f(c)$ of the predictive confidence:
	\begin{align}
	    \text{KDE-ECE} = \int&\left| \tilde{\pi}(c) - c \right|\tilde{f}(c) dc\\
	    \tilde{\pi}(c) =& \frac{\sum\limits_{i \in \mathcal{N}}  \mathbf{1}(y_i=\hat{y}_i) K_{h}(c - \hat{c}_{i})}{\sum\limits_{i \in \mathcal{N}} K_{h}(c - \hat{c}_i)}\\
	    \tilde{f}(c) =& \frac{h^{-1}}{|\mathcal{N}|}\sum_{i \in \mathcal{N}}K_{h}(c - \hat{c}_{i})
	\end{align}
	where $i\in \mathcal{N} \subset \mathcal{V}$ denotes the evaluated node, and $h$ is the bandwidth of the kernel function. 
	We follow the official implementation\footnote{\url{https://github.com/zhang64-llnl/Mix-n-Match-Calibration}} of KDE-ECE, where the the Triweight Kernel $K_h(u)=(1/h)\frac{35}{32}(1-(u/h)^2)^3$ \citep{HAAN19992007} on $[-1,1]$  is chosen as the kernel function and bandwidth is calculated as $h=1.06\sigma |\mathcal{N}|^{-1/5}$  \citep{Scott} with $\sigma$ being the standard deviation of the confidence.
	\end{itemize}

	The classwise-ECEs are summarized in Table~\ref{tab:calib_class_ece}, and the KDE-ECEs are collected in Table~\ref{tab:calib_kde}. In general, we observe similar conclusions as in the confidence ECE case (c.f.\ Section~\ref{subsec:expc_comparison}): Overall GATS achieves the state-of-the-art calibration results.
	
		\begin{table}[ht]
		\caption{GNN calibration results in terms of classwise-ECE (in percentage, lower is better).}
		\label{tab:calib_class_ece}
		\vspace{7pt} 
		\centering
		\small
		\begin{tabular}{lccccccc}
			\toprule
			Dataset 
			&Model 
			&Uncal
			&TS
			&VS
			&ETS
			&CaGCN
			&GATS\\
			\midrule
			\multirow{2}{*}{Cora} 
			&GCN &4.34$\pm$1.41 &2.06$\pm$0.27 &2.11$\pm$0.30
			&2.07$\pm$0.26
			&2.23$\pm$0.29
			&\textbf{2.03$\pm$0.24}\\
			&GAT &7.24$\pm$0.46
			&2.35$\pm$0.23
			&\textbf{2.03$\pm$0.23}
			&2.34$\pm$0.24
			&2.24$\pm$0.26 
			&2.34$\pm$0.28\\
			\midrule
			\multirow{2}{*}{Citeseer} 
			&GCN &4.51$\pm$1.86 &2.85$\pm$0.40
			&2.77$\pm$0.39
			&2.82$\pm$0.42
			&3.16$\pm$0.47
			&\textbf{2.74$\pm$0.39}\\
			&GAT &8.34$\pm$1.02 &3.12$\pm$0.52
			&\textbf{2.86$\pm$0.48}
			&3.09$\pm$0.51
			&3.22$\pm$0.53
			&3.10$\pm$0.58\\
			\midrule
			\multirow{2}{*}{Pubmed} 
			&GCN &4.96$\pm$0.99
			&1.38$\pm$0.26
			&1.53$\pm$0.30
			&1.39$\pm$0.26
			&1.35$\pm$0.33
			&\textbf{1.26$\pm$0.26} \\
			&GAT &8.54$\pm$0.49 &1.94$\pm$0.31
			&1.96$\pm$0.27
			&1.94$\pm$0.31
			&\textbf{1.89$\pm$0.37}
			&2.00$\pm$0.32 \\
			\midrule
			\multirow{2}{*}{Computers} 
			&GCN &0.97$\pm$0.16 &0.93$\pm$0.11 &0.91$\pm$0.12 
			&0.95$\pm$0.11
			&\textbf{0.84$\pm$0.10}
			&0.89$\pm$0.08\\
			&GAT &0.83$\pm$0.13
			&0.81$\pm$0.09
			&\textbf{0.80$\pm$0.10}
			&0.82$\pm$0.10
			&0.84$\pm$0.11
			&\textbf{0.80$\pm$0.08}\\
			\midrule
			\multirow{2}{*}{Photo} 	
			&GCN &0.89$\pm$0.22
			&0.78$\pm$0.12
			&0.81$\pm$0.15
			&0.78$\pm$0.15
			&0.80$\pm$0.08
			&\textbf{0.76$\pm$0.10}\\
			&GAT &0.92$\pm$0.26
			&0.84$\pm$0.15
			&\textbf{0.82$\pm$0.15}
			&0.84$\pm$0.17
			&0.89$\pm$0.12
			&0.83$\pm$0.15\\
			\midrule
			\multirow{2}{*}{CS} 
			&GCN &0.39$\pm$0.11
			&\textbf{0.29$\pm$0.02}
			&0.32$\pm$0.03
			&\textbf{0.29$\pm$0.02}
			&0.42$\pm$0.10
			&\textbf{0.29$\pm$0.03}\\
			&GAT &0.39$\pm$0.15
			&0.34$\pm$0.04
			&0.34$\pm$0.03
			&0.34$\pm$0.04
			&0.47$\pm$0.09
			&\textbf{0.33$\pm$0.04}\\
			\midrule
			\multirow{2}{*}{Physics} 
			&GCN &0.39$\pm$0.11 
			&0.36$\pm$0.06 
			&\textbf{0.35$\pm$0.05}
			&0.36$\pm$0.06
			&0.47$\pm$0.15
			&0.36$\pm$0.05\\
			&GAT &0.39$\pm$0.06
			&0.39$\pm$0.05
			&\textbf{0.37$\pm$0.04}
			&0.39$\pm$0.05
			&0.55$\pm$0.14
			&0.39$\pm$0.06\\
			\midrule
			\multirow{2}{*}{CoraFull}
			&GCN &0.35$\pm$0.03 
			&0.34$\pm$0.01
			&0.34$\pm$0.01
			&0.34$\pm$0.02
			&0.34$\pm$0.04
			&\textbf{0.33$\pm$0.02}\\
			&GAT &0.33$\pm$0.03
			&0.32$\pm$0.01
			&0.32$\pm$0.02
			&0.32$\pm$0.01
			&0.34$\pm$0.06
			&\textbf{0.31$\pm$0.02}\\
			\bottomrule
		\end{tabular}
	\end{table}	

		\begin{table}[ht]
		\caption{GNN calibration results in terms of KDE-ECE (in percentage, lower is better).}
		\label{tab:calib_kde}
		\vspace{7pt} 
		\centering
		\small
		\begin{tabular}{lccccccc}
			\toprule
			Dataset 
			&Model 
			&Uncal
			&TS
			&VS
			&ETS
			&CaGCN
			&GATS\\
			\midrule
			\multirow{2}{*}{Cora} 
			&GCN &13.35$\pm$5.07 &3.21$\pm$1.18 &3.56$\pm$1.25
			&3.26$\pm$1.23
			&4.17$\pm$1.48
		&\textbf{3.11$\pm$1.19}\\
			&GAT &23.33$\pm$1.79 &3.00$\pm$0.80
			&\textbf{2.79$\pm$0.81}
			&2.97$\pm$0.85
			&3.17$\pm$0.87
			&2.93$\pm$0.95
			\\
			\midrule
	\multirow{2}{*}{Citeseer} 
			&GCN &10.72$\pm$5.86 &4.80$\pm$1.40
			&4.56$\pm$1.45
			&4.66$\pm$1.46
			&6.09$\pm$1.30
		&\textbf{4.17$\pm$1.31}\\
			&GAT &22.86$\pm$3.54
			&4.42$\pm$1.46
			&3.79$\pm$1.53
			&4.32$\pm$1.42
			&5.17$\pm$1.23
			&\textbf{3.66$\pm$1.58}\\
			\midrule
			\multirow{2}{*}{Pubmed} 
			&GCN &7.33$\pm$1.48 &1.32$\pm$0.27 &1.57$\pm$0.38
			&1.35$\pm$0.29
			&1.29$\pm$0.48
	   	    &\textbf{1.07$\pm$0.26} \\
			&GAT &12.32$\pm$0.80 &1.20$\pm$0.29
			&1.12$\pm$0.29
			&1.20$\pm$0.29
			&\textbf{1.08$\pm$0.30}
	        &1.16$\pm$0.30 \\
			\midrule
			\multirow{2}{*}{Computers} 
			&GCN &3.05$\pm$0.97  &2.60$\pm$0.71 &2.68$\pm$0.75
			&2.73$\pm$0.76
			&\textbf{1.65$\pm$0.62}
			&2.16$\pm$0.61 	
			\\
			&GAT &1.89$\pm$1.00
			&1.62$\pm$0.63
			&1.70$\pm$0.69
			&1.67$\pm$0.72
			&1.75$\pm$0.62
			&\textbf{1.47$\pm$0.52}\\
			\midrule
			\multirow{2}{*}{Photo} 	
			&GCN &2.59$\pm$1.29 
			 &1.82$\pm$0.87
			 &1.94$\pm$0.92
			 &1.90$\pm$0.95
			 &\textbf{1.65$\pm$0.45}
			 &1.67$\pm$0.70\\
			&GAT &2.25$\pm$1.16
			&1.74$\pm$0.67
			&1.81$\pm$0.75
			&1.81$\pm$0.74
			&1.75$\pm$0.60
			&\textbf{1.73$\pm$0.65}\\
			\midrule
			\multirow{2}{*}{CS} 
			&GCN &2.14$\pm$0.98
			&1.10$\pm$0.11
			&1.11$\pm$0.17
			&1.09$\pm$0.11
			&1.95$\pm$0.90
			&\textbf{1.06$\pm$0.12}\\
			&GAT &1.74$\pm$1.30
			&1.12$\pm$0.25
			&1.10$\pm$0.26
			&1.14$\pm$0.26
			&2.10$\pm$0.88
			&\textbf{1.07$\pm$0.21}\\
			\midrule
			\multirow{2}{*}{Physics} 
			&GCN &0.94$\pm$0.27
			&0.83$\pm$0.09
			&\textbf{0.82$\pm$0.07}
			&0.83$\pm$0.09
			&0.96$\pm$0.24
			&0.85$\pm$0.09\\
			&GAT &0.84$\pm$0.10
			&0.84$\pm$0.08
			&0.85$\pm$0.09
			&0.84$\pm$0.08
			&1.07$\pm$0.24
			&\textbf{0.83$\pm$0.08}\\
			\midrule
			\multirow{2}{*}{CoraFull}
			&GCN &6.46$\pm$1.30
			&5.45$\pm$0.43
			&5.66$\pm$0.41
			&5.43$\pm$0.45
			&5.73$\pm$2.54
			&\textbf{3.78$\pm$0.90}\\
			&GAT &4.76$\pm$1.44
			&3.98$\pm$0.51
			&4.14$\pm$0.45
			&3.96$\pm$0.53
			&5.90$\pm$3.11
			&\textbf{3.56$\pm$0.66}\\
			\bottomrule
		\end{tabular}
	\end{table}
	
	\newpage
	\paragraph{Non calibration metrics}
	Although not calibration metrics, we also report the results in terms of negative log-likelihood (Table~\ref{tab:calib_comparison_nll}) and Brier score (Table~\ref{tab:calib_comparison_brier}) for reference.


	\begin{table}[ht]
		\caption{GNN calibration results in terms of negative log-likelihood $(\times 10^{-2})$.}
		\label{tab:calib_comparison_nll}
		\vspace{7pt} 
		\centering
		\small
		\setlength{\tabcolsep}{2.5pt} 
		\begin{tabular}{lccccccc}
			\toprule
			Dataset 
			&Model 
			&Uncal
			&TS
			&VS
			&ETS
			&CaGCN
			&GATS\\
			\midrule
			\multirow{2}{*}{Cora} 
			&GCN &62.90$\pm$5.68  &56.37$\pm$3.12 
			&57.66$\pm$4.37 
			&56.01$\pm$3.00 
			&66.88$\pm$7.78
			&55.91$\pm$3.17\\
			&GAT &75.67$\pm$2.37
			&57.51$\pm$2.87
			&55.91$\pm$4.02
			&57.15$\pm$2.70
			&60.79$\pm$4.66
			&57.02$\pm$2.33\\
			\midrule
			\multirow{2}{*}{Citeseer} 
			& GCN &90.13$\pm$6.00
			&86.99$\pm$2.74
			&87.01$\pm$2.33
			&86.61$\pm$2.56
			&92.95$\pm$4.88
			&86.18$\pm$2.35\\
			& GAT &100.90$\pm$6.33
			&86.57$\pm$3.30
			&86.01$\pm$1.87
			&86.18$\pm$2.79
			&89.07$\pm$3.40
			&86.20$\pm$3.22\\
			\midrule
			\multirow{2}{*}{Pubmed} 
			& GCN &39.31$\pm$1.47
			&36.75$\pm$0.68
			&36.79$\pm$0.69
			&36.53$\pm$0.67
			&35.97$\pm$1.16
			&36.39$\pm$0.62\\
			& GAT &46.87$\pm$0.98
			&40.06$\pm$0.76
			&40.06$\pm$0.74
			&40.07$\pm$0.75
			&39.78$\pm$0.77
			&40.05$\pm$0.75\\
			\midrule
			\multirow{2}{*}{Computers} 
			& GCN &42.96$\pm$1.21
			&42.93$\pm$1.17
			&42.87$\pm$1.15
			&41.08$\pm$1.31
			&43.31$\pm$3.47
			&42.49$\pm$1.25\\
			& GAT &37.26$\pm$1.53
			&37.18$\pm$1.48
			&37.07$\pm$1.38
			&36.54$\pm$1.60
			&40.38$\pm$4.06
			&37.11$\pm$1.54\\
			\midrule
			\multirow{2}{*}{Photo} 			
			& GCN &28.92$\pm$1.20
			&29.02$\pm$1.18
			&29.25$\pm$1.31
			&27.19$\pm$1.24
			&37.59$\pm$7.72
			&28.81$\pm$1.23
			\\
			& GAT &26.83$\pm$1.78
			&26.82$\pm$1.61
			&26.79$\pm$1.61
			&26.40$\pm$1.76
			&32.75$\pm$5.37
			&26.93$\pm$1.77\\
			\midrule
			\multirow{2}{*}{CS} 
			& GCN &21.85$\pm$0.74
			&21.38$\pm$0.48
			&21.65$\pm$0.45
			&21.36$\pm$0.46
			&27.38$\pm$5.57
			&21.28$\pm$0.49\\
			& GAT &24.76$\pm$1.46
			&24.57$\pm$0.87
			&24.59$\pm$0.72
			&24.49$\pm$0.85
			&29.79$\pm$3.86
			&24.49$\pm$0.83\\
			\midrule
			\multirow{2}{*}{Physics} 
			& GCN &11.95$\pm$0.41
			&11.88$\pm$0.34
			&11.90$\pm$0.33
			&11.89$\pm$0.34
			&13.00$\pm$1.27
			&11.87$\pm$0.32\\
			& GAT &12.88$\pm$0.41
			&12.88$\pm$0.39
			&12.84$\pm$0.38
			&12.88$\pm$0.39
			&13.52$\pm$0.67
			&12.87$\pm$0.39\\
			\midrule
			\multirow{2}{*}{CoraFull}
			& GCN &143.07$\pm$2.02
			&142.71$\pm$1.80
			&142.85$\pm$1.98
			&141.74$\pm$1.61
			&146.55$\pm$12.81
			&140.10$\pm$1.92\\
			& GAT &139.77$\pm$2.16
			&139.57$\pm$1.89
			&139.72$\pm$1.91
			&138.97$\pm$1.86
			&150.70$\pm$17.93
			&139.06$\pm$1.84\\
			\bottomrule
		\end{tabular}
	\end{table}

	\begin{table}[ht]
		\caption{GNN calibration results in terms of Brier score $(\times 10^{-2})$.}
		\label{tab:calib_comparison_brier}
		\vspace{7pt} 
		\centering
		\small
		\setlength{\tabcolsep}{4pt} 
		\begin{tabular}{lccccccc}
			\toprule
			Dataset 
			&Model 
			&Uncal
			&TS
			&VS
			&ETS
			&CaGCN
			&GATS\\
			\midrule
			\multirow{2}{*}{Cora} 
			&GCN &28.68$\pm$2.54
			&25.62$\pm$0.98 
			&25.65$\pm$1.05 
			&25.62$\pm$0.97
			&26.19$\pm$1.05
			&25.59$\pm$1.07\\
			&GAT &34.47$\pm$1.21
			&26.67$\pm$1.05
			&25.31$\pm$1.03
			&26.67$\pm$1.03
			&26.71$\pm$1.17
			&26.71$\pm$1.00\\
			\midrule
			\multirow{2}{*}{Citeseer} 
			& GCN &42.56$\pm$2.82
			&40.76$\pm$0.76
			&40.94$\pm$0.93
			&40.71$\pm$0.78
			&41.57$\pm$1.06
			&40.63$\pm$0.74\\
			& GAT &47.42$\pm$3.26
			&40.62$\pm$0.90
			&40.44$\pm$0.62
			&40.55$\pm$0.90
			&40.99$\pm$0.97
			&40.61$\pm$0.95\\
			\midrule
			\multirow{2}{*}{Pubmed} 
			& GCN &21.31$\pm$0.71
			&20.20$\pm$0.36
			&20.24$\pm$0.38
			&20.20$\pm$0.36
			&20.05$\pm$0.41
			&20.17$\pm$0.36\\
			& GAT &25.33$\pm$0.56
			&22.68$\pm$0.41
			&22.64$\pm$0.41
			&22.68$\pm$0.41
			&22.58$\pm$0.43
			&22.67$\pm$0.41\\
			\midrule
			\multirow{2}{*}{Computers} 
			& GCN &18.57$\pm$0.80
			&18.50$\pm$0.68
			&18.42$\pm$0.64
			&18.51$\pm$0.68
			&18.13$\pm$0.70
			&18.42$\pm$0.65\\
			& GAT &16.79$\pm$0.80
			&16.76$\pm$0.75
			&16.61$\pm$0.64
			&16.76$\pm$0.75
			&16.84$\pm$0.73
			&16.75$\pm$0.73\\
			\midrule
			\multirow{2}{*}{Photo} 			
			& GCN &11.72$\pm$0.66
			&11.60$\pm$0.59
			&11.62$\pm$0.65
			&11.60$\pm$0.60
			&11.67$\pm$0.51
			&11.56$\pm$0.56\\
			& GAT &11.52$\pm$0.88
			&11.45$\pm$0.77
			&11.35$\pm$0.69
			&11.45$\pm$0.77
			&11.59$\pm$0.70
			&11.46$\pm$0.76\\
			\midrule
			\multirow{2}{*}{CS} 
			& GCN &10.28$\pm$0.27
			&10.16$\pm$0.20
			&10.20$\pm$0.19
			&10.16$\pm$0.20
			&10.60$\pm$0.42
			&10.14$\pm$0.21\\
			& GAT &11.36$\pm$0.58
			&11.27$\pm$0.35
			&11.25$\pm$0.30
			&11.27$\pm$0.35
			&11.65$\pm$0.39
			&11.27$\pm$0.34\\
			\midrule
			\multirow{2}{*}{Physics} 
			& GCN &6.13$\pm$0.20
			&6.13$\pm$0.19
			&6.13$\pm$0.19
			&6.13$\pm$0.19
			&6.25$\pm$0.22
			&6.12$\pm$0.19\\
			& GAT &6.54$\pm$0.18
			&6.54$\pm$0.18
			&6.53$\pm$0.17
			&6.54$\pm$0.18
			&6.64$\pm$0.18
			&6.53$\pm$0.18\\
			\midrule
			\multirow{2}{*}{CoraFull}
			& GCN &52.32$\pm$0.68
			&52.09$\pm$0.51
			&52.01$\pm$0.49
			&52.08$\pm$0.51
			&52.20$\pm$1.20
			&51.61$\pm$0.54\\
			& GAT &51.73$\pm$0.77
			&51.60$\pm$0.60
			&51.55$\pm$0.59
			&51.59$\pm$0.60
			&52.53$\pm$1.71
			&51.54$\pm$0.59\\
			\bottomrule
		\end{tabular}
	\end{table}

	\subsection{Results for additional baselines} \label{sec:exp_calib_more_baselines}
	
	While in the main paper we focus on ``temperature scaling style'' methods which directly rescale the output logits, here we compare with the following additional calibration methods which have different principles. These methods are all designed for multi-class classification and do not consider the structural information of the graph.
	\begin{itemize}
	    \item \textbf{Multi-class isotonic regression (IRM)} \citep{zhang2020mixnmatch} is a multi-class generalization of the non-parametric isotonic regression method;
	    \item \textbf{Calibration using spline (Spline)} \citep{gupta2021spline} fits the calibration function with splines;
	    \item \textbf{Dirichlet calibration (DIR)} \citep{kull2019dirichlet} uses the Dirichlet distribution to model the distribution of probabilistic outputs. It also employs an off-diagonal and intercept regularization (ODIR);
	    \item \textbf{Order invariant calibration (OI)} \citep{rahimi2020intraorder} is the order-invariant intra order-preserving model. It uses sorted output logits as calibration input and builds up a neural network with special structures to preserve the accuracy and the intra order of the predicted logits.
	\end{itemize}
	
	The authors of spline calibration specify how to calibrate a specific class or a chosen top-$r$ class, and in their implementation\footnote{\url{https://github.com/kartikgupta-at-anu/spline-calibration}} they focus on calibrating the top-$1$ class. However, it is not clear how to adjust the rest of the predictions to ensure valid probabilistic predictions after calibration. We adopt a heuristic which proportionally rescales the non top-$1$ output probabilities so that the calibrated probabilistic output sums up to one. Also, the authors wrongly claimed that calibrating the top-$1$ score ``does not alter the classification accuracy'' \citep{gupta2021spline}. 
	In practice, the score after calibration might no longer remain top-$1$ and the predictions could be altered.
	
	For Dirichlet calibration, we find out that the scaling factors $\lambda,\mu$ of ODIR affect the performance and need to be tuned depending on the dataset. Thus we do a hyperparameter search for each dataset with search space $(0.01, 0.1, 1, 10, 100, 1000, 10000, 100000)$. 
	
	Table~\ref{gupta2021spline} summarizes the calibration results in terms of ECE. Note that we do not include the results for Dirichlet calibration on CoraFull because it fails to calibrate the GNN backbones and significantly deteriorates the predictive accuracies ($< 10 \%$ v.s.\ $>60\%$ before calibration).
	
	Overall we observe that GATS still achieves the state-of-the-art performance compared to the additional baselines for GNN calibration.
	
		\begin{table}[ht]
		\caption{GNN calibration results (ECE, in percentage) with additional baselines.}
		\label{gupta2021spline}
		\vspace{7pt} 
		\centering
		\small
		\setlength{\tabcolsep}{5pt} 
		\begin{tabular}{lccccccc}
			\toprule
			Dataset 
			&Model 
			&Uncal
			&IRM
			&Spline
			&DIR
			&OI
			&GATS\\
			\midrule
			\multirow{2}{*}{Cora} 
			& GCN &13.04$\pm$5.22 
			&3.69$\pm$1.17
			&4.89$\pm$1.27
			&3.93$\pm$1.26
			&4.83$\pm$1.50			
			&\textbf{3.64$\pm$1.34}\\
			& GAT &23.31$\pm$1.81 
			&3.45$\pm$0.91
			&4.71$\pm$1.76
			&3.42$\pm$0.72
			&4.24$\pm$1.39
			&\textbf{3.18$\pm$0.90}\\
			\midrule
			\multirow{2}{*}{Citeseer} & GCN &10.66$\pm$5.92 
			&5.08$\pm$1.34
			&6.70$\pm$1.42
			&5.40$\pm$1.52
			&6.36$\pm$1.48
			&\textbf{4.43$\pm$1.30}\\
			& GAT 
			&22.88$\pm$3.53 
			&4.15$\pm$1.50
			&6.07$\pm$1.77
			&4.87$\pm$1.36
			&6.08$\pm$1.30
			&\textbf{3.86$\pm$1.56}\\
			\midrule
			\multirow{2}{*}{Pubmed} 
			& GCN 
			&7.18$\pm$1.51 
			&1.64$\pm$0.58
			&1.72$\pm$0.46
			&1.42$\pm$0.33
			&1.23$\pm$0.44			
			&\textbf{0.98$\pm$0.30}\\
			& GAT 
			&12.32$\pm$0.80 
			&1.63$\pm$0.60
			&1.69$\pm$0.60
			&\textbf{0.93$\pm$0.26}
			&1.36$\pm$0.47		
			&1.03$\pm$0.32\\
			\midrule
			\multirow{2}{*}{Computers} 
			& GCN
			&3.00$\pm$0.80 
			&1.98$\pm$0.48
			&\textbf{1.56$\pm$0.44}
			&3.31$\pm$0.63 
			&1.86$\pm$0.55	
			&2.23$\pm$0.49 \\
			& GAT 
			&1.88$\pm$0.82 
			&\textbf{1.32$\pm$0.35}
			&1.56$\pm$0.53
			&2.23$\pm$0.73
			&2.17$\pm$0.72
			&1.39$\pm$0.39\\
			\midrule
			\multirow{2}{*}{Photo} 
			& GCN 
			&2.24$\pm$1.03 
			&1.53$\pm$0.47
			&1.68$\pm$0.57
			&1.61$\pm$0.60
			&1.75$\pm$0.49
			&\textbf{1.51$\pm$0.52}\\
			& GAT 
			&2.02$\pm$1.11 
			&1.53$\pm$0.51
			&1.59$\pm$0.66
			&1.39$\pm$0.62
			&1.85$\pm$0.66
			&\textbf{1.48$\pm$0.61}\\
			\midrule
			\multirow{2}{*}{CS} 
			& GCN &1.65$\pm$0.92 
			&1.29$\pm$0.32
			&1.08$\pm$0.38
			&0.90$\pm$0.19
			&1.55$\pm$0.50
			&\textbf{0.88$\pm$0.30}\\
			& GAT 
			&1.40$\pm$1.25 
			&1.09$\pm$0.35
			&1.16$\pm$0.39
			&0.96$\pm$0.39
			&1.80$\pm$0.80
			&\textbf{0.81$\pm$0.30}\\
			\midrule
			\multirow{2}{*}{Physics} 
			& GCN 
			&0.52$\pm$0.29 
			&0.59$\pm$0.17
			&0.54$\pm$0.23
			&\textbf{0.44$\pm$0.15}
			&0.64$\pm$0.29
			&0.46$\pm$0.16\\
			& GAT 
			&0.45$\pm$0.21 
			&0.56$\pm$0.16
			&0.45$\pm$0.18
			&\textbf{0.42$\pm$0.14}
			&0.60$\pm$0.32
			&\textbf{0.42$\pm$0.14}\\
			\midrule
			\multirow{2}{*}{CoraFull}
			&GCN 
			&6.50$\pm$1.26 
			&4.33$\pm$0.77
			&\textbf{2.92$\pm$0.79}
			&N/A
			&10.61$\pm$1.40
			&3.76$\pm$0.74\\
			&GAT 
			&4.73$\pm$1.39 
			&3.18$\pm$0.56
			&\textbf{2.68$\pm$0.89}
			&N/A
			&8.33$\pm$2.18
			&3.54$\pm$0.63\\
			\bottomrule
		\end{tabular}
	\end{table}
	
	\subsection{Accuracies of calibration methods}
	Since many baseline calibration methods are not accuracy-preserving, in Table~\ref{tab:calib_comparison_acc} we additionally report their test accuracies. Accuracy preserving methods (GATS, TS, ETS, CaGCN, OI) have the same accuracies as the uncalibrated case, which is also reported for reference.
	
	\begin{table}[ht]
		\caption{Test accuracies of uncalibrated results (identical to those from accuracy-preserving methods) and calibrated predictions from non accuracy-preserving methods.}
		\label{tab:calib_comparison_acc}
		\vspace{7pt} 
		\centering
		\small
		\begin{tabular}{lcccccc}
			\toprule
			Dataset 
			&Model 
			&Uncal
			&VS
			&IRM
			&Spline
			&DIR\\
			\midrule
			\multirow{2}{*}{Cora} 
			& GCN &82.78$\pm$0.79 
			&82.90$\pm$0.89
			&82.56$\pm$0.87
			&82.78$\pm$0.80
			&83.16$\pm$0.87\\
			& GAT &81.98$\pm$0.92
			&82.98$\pm$0.77
			&81.74$\pm$1.04
			&81.98$\pm$0.92
			&82.77$\pm$0.85\\
			\midrule
			\multirow{2}{*}{Citeseer} 
			& GCN &72.19$\pm$0.82 
			&72.06$\pm$0.90
			&72.04$\pm$0.79
			&72.16$\pm$0.82
			&72.31$\pm$0.99\\
			& GAT &72.37$\pm$0.68
			&72.25$\pm$0.64
			&72.18$\pm$0.73
			&72.34$\pm$0.73
			&72.53$\pm$0.59\\
			\midrule
			\multirow{2}{*}{Pubmed} 
			& GCN &86.40$\pm$0.27 
			&86.39$\pm$0.29
			&86.31$\pm$0.29
			&86.39$\pm$0.26
			&86.43$\pm$0.25\\
			& GAT &84.46$\pm$0.34
			&84.55$\pm$0.38
			&84.25$\pm$0.38
			&84.44$\pm$0.34
			&84.62$\pm$0.35\\
			\midrule
			\multirow{2}{*}{Computers} 
			& GCN &88.13$\pm$0.56
			&88.19$\pm$0.56
			&88.20$\pm$0.54
			&88.12$\pm$0.55
			&87.78$\pm$0.65\\
			& GAT &89.05$\pm$0.60 
			&89.16$\pm$0.52
			&89.03$\pm$0.60
			&89.04$\pm$0.60
			&89.00$\pm$0.63\\
			\midrule
			\multirow{2}{*}{Photo} 			
			& GCN &92.65$\pm$0.38 
			&92.69$\pm$0.43
			&92.61$\pm$0.42
			&92.65$\pm$0.38
			&92.69$\pm$0.49\\
			& GAT &92.65$\pm$0.54 
			&92.76$\pm$0.45
			&92.58$\pm$0.57
			&92.64$\pm$0.54
			&92.92$\pm$0.44\\
			\midrule
			\multirow{2}{*}{CS} 
			& GCN &93.33$\pm$0.15 
			&93.29$\pm$0.15
			&93.29$\pm$0.16
			&93.32$\pm$0.15
			&93.33$\pm$0.15\\
			& GAT  &92.57$\pm$0.25 
			&92.57$\pm$0.22
			&92.54$\pm$0.24
			&92.56$\pm$0.24
			&92.60$\pm$0.22\\
			\midrule
			\multirow{2}{*}{Physics} 
			& GCN &95.99$\pm$0.14 
			&95.98$\pm$0.14
			&95.98$\pm$0.15
			&95.98$\pm$0.14
			&96.00$\pm$0.14\\
			& GAT &95.70$\pm$0.13 
			&95.71$\pm$0.12
			&95.67$\pm$0.14
			&95.69$\pm$0.14
			&95.72$\pm$0.11\\
			\midrule
			\multirow{2}{*}{CoraFull}
			& GCN &63.07$\pm$0.50 
			&63.24$\pm$0.45
			&62.94$\pm$0.48
			&63.07$\pm$0.50
			&N/A\\
			& GAT &63.00$\pm$0.59 
			&63.10$\pm$0.52
			&62.85$\pm$0.59
			&63.00$\pm$0.58
			&N/A\\
			\bottomrule
		\end{tabular}
	\end{table}
	
    \section{Data efficiency and expressiveness of GATS: GAT results} \label{sec:data_efficiency_gat}
	
\begin{figure}[ht!]
		\centering
		\includegraphics[width=0.35\linewidth]{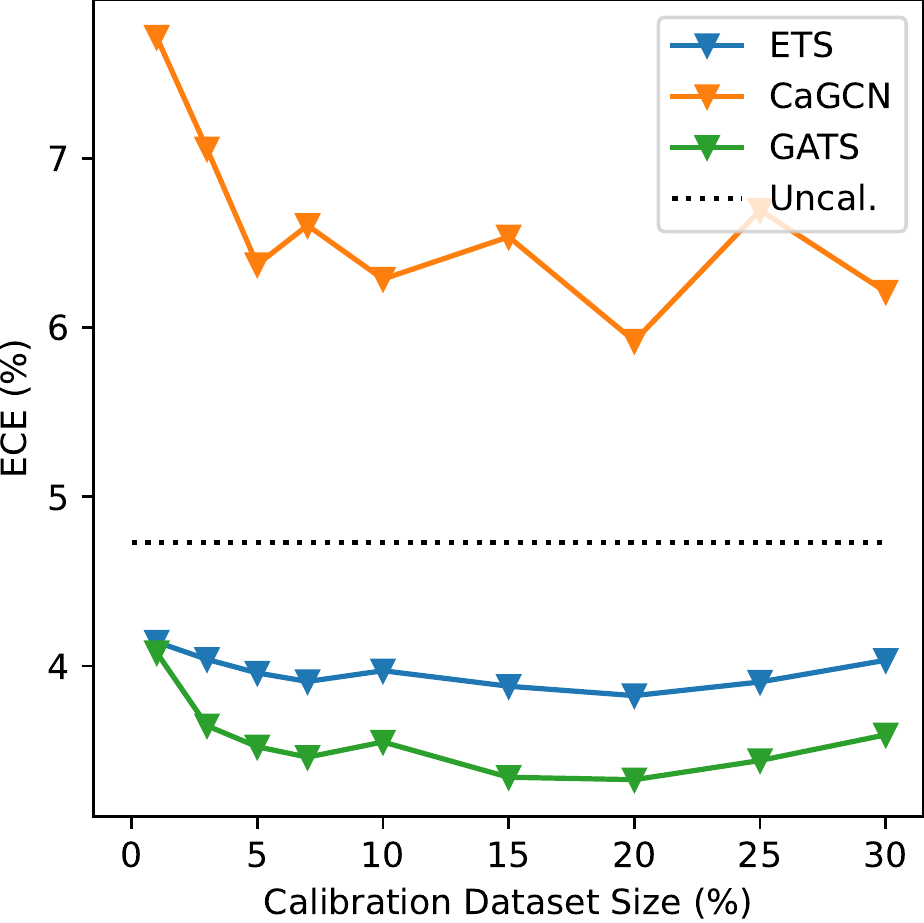}
		\caption{ECEs (in percentage) on CoraFull with GAT backbone for ETS, CaGCN, and GATS using various amounts of calibration data. Again, We observe that GATS is data-efficient and expressive. Here CaGCN fails to calibrate the GAT model on CoraFull.} \label{fig:data_efficiency_gat}
	\end{figure}

	Figure~\ref{fig:data_efficiency_gat} shows the results for the GAT case. We see that GATS is also data efficient and expressive when calibrating GAT models.
	
    \section{CaGCN results discussion}
    
    While the ECEs of CaGCN in its original paper are promising \citep{wang2021cagcn}, we observe that the ECEs of CaGCN are often unstable and sometimes even worse than that of the uncalibrated model in our experiments.
    One possible reason is that we use a different splitting from the CaGCN paper, where they follow a fixed splitting from \citet{kipf2017gcn}.
	A significant difference is that the splitting from \citet{kipf2017gcn} has more validation nodes than the training nodes.
	This differs from typical real-world applications, where the larger fold would often be used to train a good classifier \citep{NEURIPS2018_c1fea270}, and only the smaller fold is available for fitting the calibrator.
	
\begin{figure}[ht!]
		\centering
		\begin{subfigure}[c]{0.3\textwidth}
			\includegraphics[width=\textwidth]{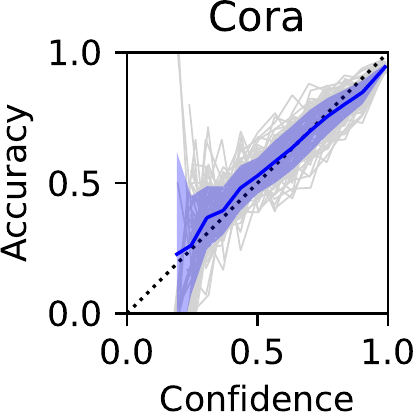}
		\end{subfigure}
		\begin{subfigure}[c]{0.3\textwidth}
			\includegraphics[width=\textwidth]{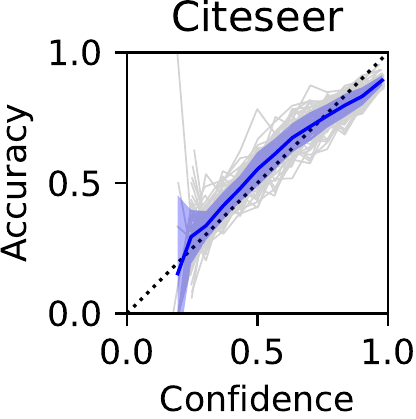}
		\end{subfigure}
		\begin{subfigure}[c]{0.3\textwidth}
			\includegraphics[width=\textwidth]{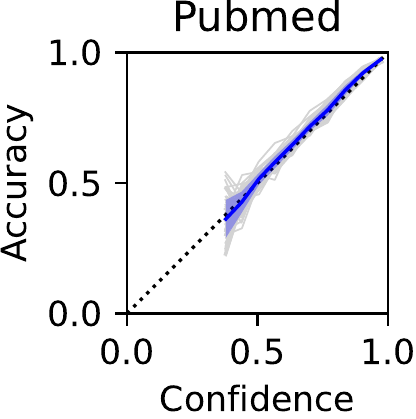}
		\end{subfigure}

		\begin{subfigure}[c]{0.3\textwidth}
			\includegraphics[width=\textwidth]{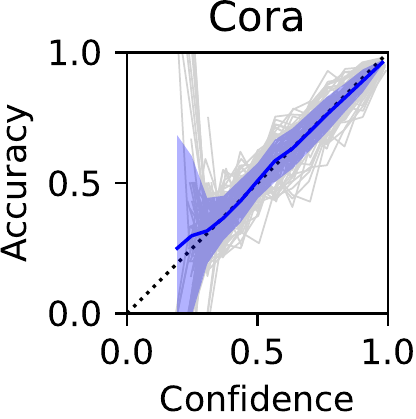}
		\end{subfigure}
		\begin{subfigure}[c]{0.3\textwidth}
			\includegraphics[width=\textwidth]{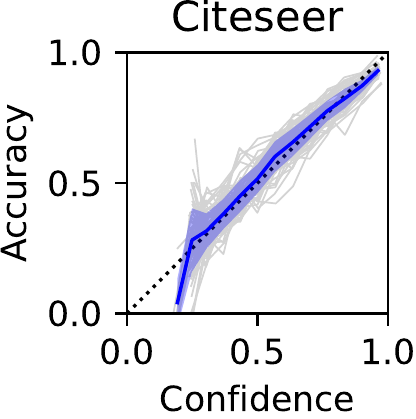}
		\end{subfigure}
		\begin{subfigure}[c]{0.3\textwidth}
			\includegraphics[width=\textwidth]{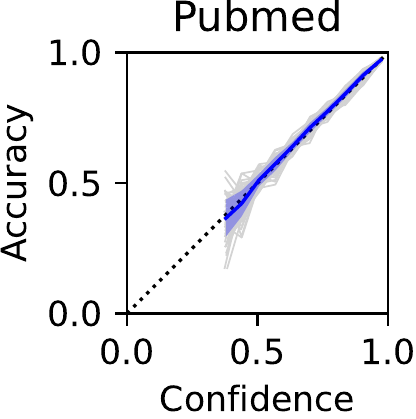}
		\end{subfigure}
		\caption{Top row: Reliability diagram of CaGCN for GCN trained on Cora, Citeseer, and Pubmed. Botom row: Reliability diagram of GATS for GCN trained on Cora, Citeseer, and Pubmed as a reference. We observe that CaGCN is noticeably over-confident in the high confidence region when the size of the validation set is relatively small (e.g.,\ Cora and Citeseer).} \label{fig:reliability cagcn}
	\end{figure}

	In our splitting, the validation sets of Cora and Citeseer are substantially smaller than those in \citet{kipf2017gcn}. We observe that CaGCN yields suboptimal calibration results (see Section~\ref{subsec:expc_comparison}) and predictions with higher confidence tend to be over-confident in the reliability diagram in Figure ~\ref{fig:reliability cagcn}.
	By contrast, the validation set in Pubmed is relatively large since it has more nodes.
	We notice that CaGCN achieves competitive results in Pubmed and the confidence-accuracy curve almost lies on the diagonal. 
	We observe that CaGCN also produces suboptimal calibration results in CoraFull, even though the validation set is large.
	We suspect that this is caused by the class imbalance of the CoraFull data. Class imbalance is known to be a challenge for many calibration methods \citep{teixeira2019gnncalib}.
	
    \section{GATS weight visualization}
    
\begin{figure}[ht!]
		\centering
			\includegraphics[width=\textwidth]{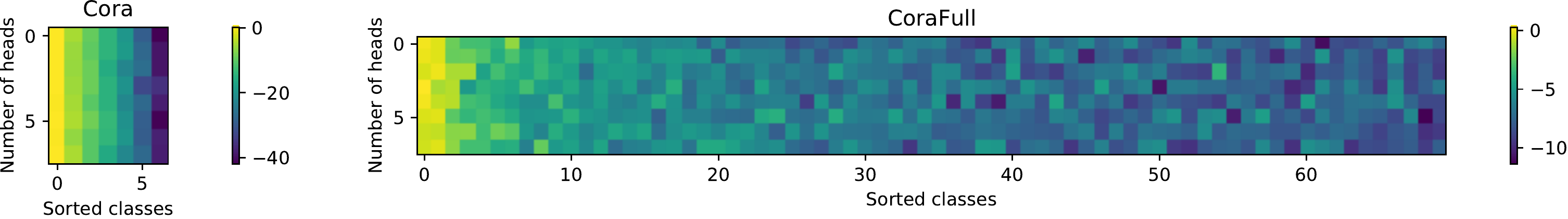}
			\caption{Visualization of the GATS weight $\theta$ on Cora (7 classes) and CoraFull (70 classes).} \label{fig:weight_viz}
	\end{figure}

    GATS learns $\tau^h_i$ from sorted logits. We discover that the absolute value of the learned weights in the linear layer generally follows the ranking of the logits across the class. 
    That is to say, logits with higher value have stronger influence to $\tau^h_i$. In Figure~\ref{fig:weight_viz} we visualize the weights $\theta = (\theta^1, \dots, \theta^H)$ of the linear layers $\phi^h$ in GATS.
    Here, it is interesting to see that the weights $\theta^h$ from different heads $h$ have slight variations. 
    Combining multiple heads in the attention with sorting could be considered as a form of ensemble without the model being overly parameterized. 
    
	\section{Analysis of correlations between the factors}
	
	In this section we visualize the correlation between the local-view factors: the distance to training nodes, the relative confidence level, and the neighborhood similarity.
	Each plot shows how the factor on the y-axis varies when the factor on the x-axis is fixed to a given value. Two factors are independent when we observe a horizontal line in the plot.
	As the relative confidence level is a model dependent factor, GCN and GAT will have different correlation plots when it is involved. 
	
	\FloatBarrier
	\subsection{Distance to training nodes -- relative confidence level}
\begin{figure}[ht!]
		\centering
		\begin{subfigure}[c]{0.135\textwidth}
			\includegraphics[width=\textwidth]{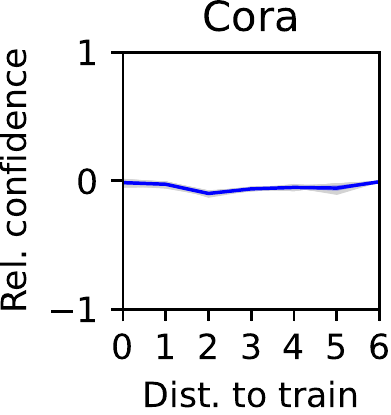}
		\end{subfigure}
		\begin{subfigure}[c]{0.135\textwidth}
			\includegraphics[width=\textwidth]{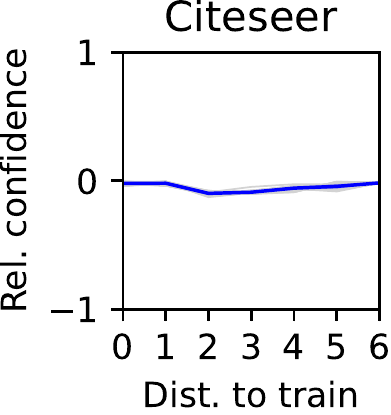}
		\end{subfigure}
		\begin{subfigure}[c]{0.135\textwidth}
			\includegraphics[width=\textwidth]{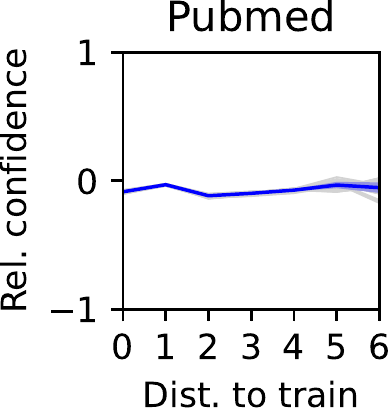}
		\end{subfigure}
		\begin{subfigure}[c]{0.135\textwidth}
			\includegraphics[width=\textwidth]{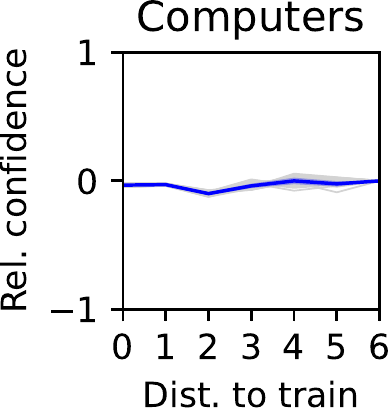}
		\end{subfigure}
		\begin{subfigure}[c]{0.135\textwidth}
			\includegraphics[width=\textwidth]{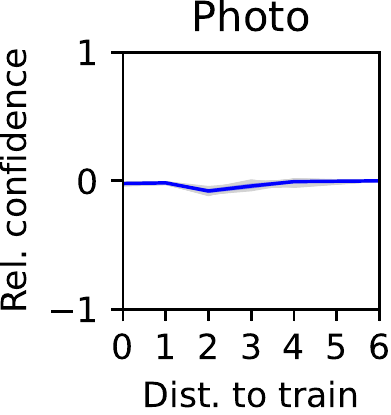}
		\end{subfigure}
		\begin{subfigure}[c]{0.135\textwidth}
			\includegraphics[width=\textwidth]{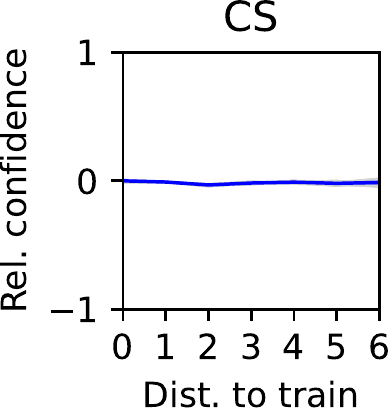}
		\end{subfigure}
		\begin{subfigure}[c]{0.135\textwidth}
			\includegraphics[width=\textwidth]{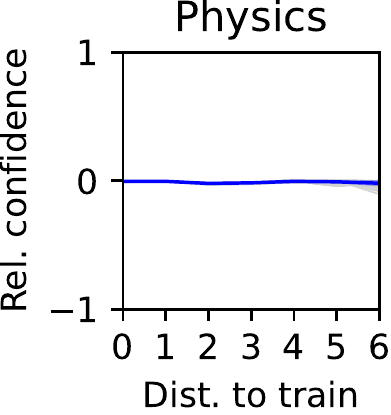}
		\end{subfigure}
		\caption{Relative confidence level of GCN results depending on the minimum distance to training nodes.}
		\label{fig:dist2dconf_gcn}
	\end{figure}

\begin{figure}[ht!]
		\centering
		\begin{subfigure}[c]{0.135\textwidth}
			\includegraphics[width=\textwidth]{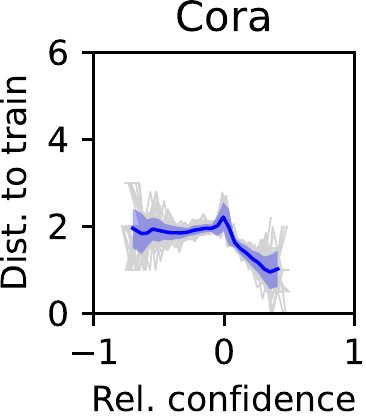}
		\end{subfigure}
		\begin{subfigure}[c]{0.135\textwidth}
			\includegraphics[width=\textwidth]{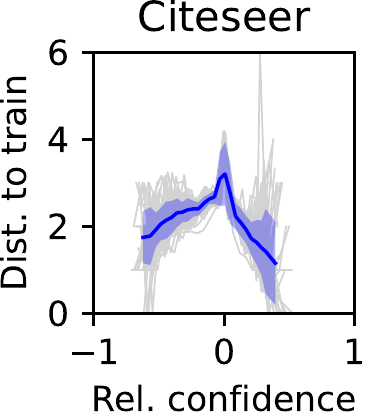}
		\end{subfigure}
		\begin{subfigure}[c]{0.135\textwidth}
			\includegraphics[width=\textwidth]{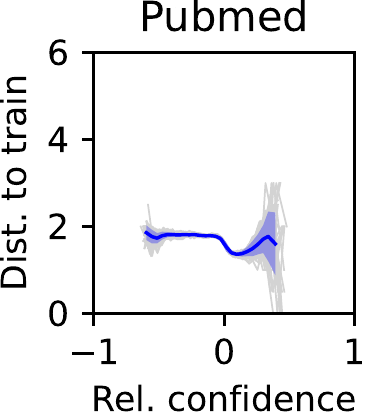}
		\end{subfigure}
		\begin{subfigure}[c]{0.135\textwidth}
			\includegraphics[width=\textwidth]{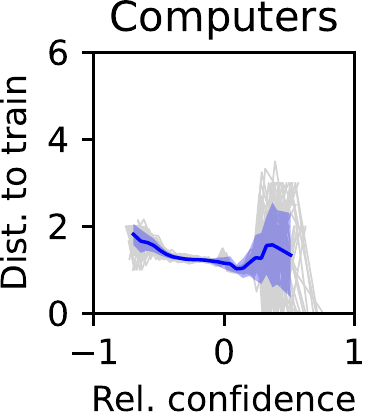}
		\end{subfigure}
		\begin{subfigure}[c]{0.135\textwidth}
			\includegraphics[width=\textwidth]{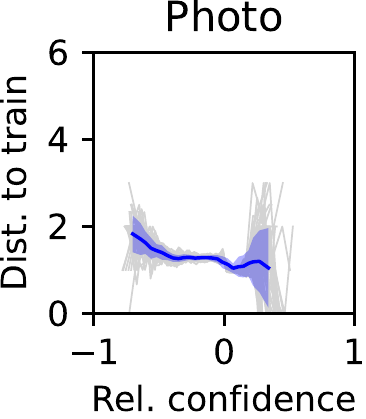}
		\end{subfigure}
		\begin{subfigure}[c]{0.135\textwidth}
			\includegraphics[width=\textwidth]{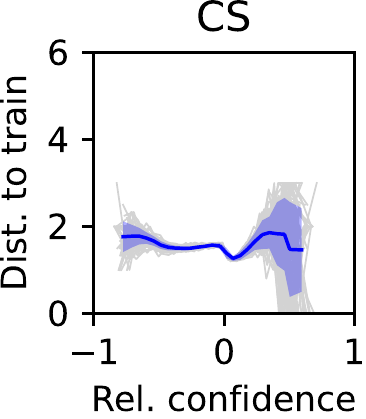}
		\end{subfigure}
		\begin{subfigure}[c]{0.135\textwidth}
			\includegraphics[width=\textwidth]{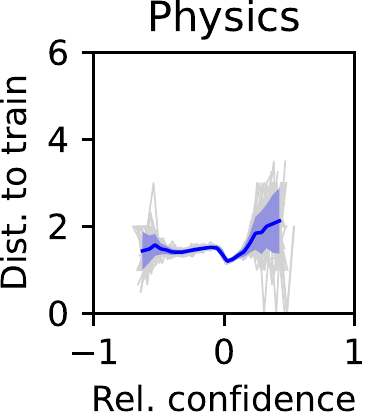}
		\end{subfigure}
		\caption{Minimum distance to training
nodes depending on the relative confidence level of GCN predictions.}
		\label{fig:dconf2dist_gcn}
	\end{figure}

\begin{figure}[ht!]
		\centering
		\begin{subfigure}[c]{0.135\textwidth}
			\includegraphics[width=\textwidth]{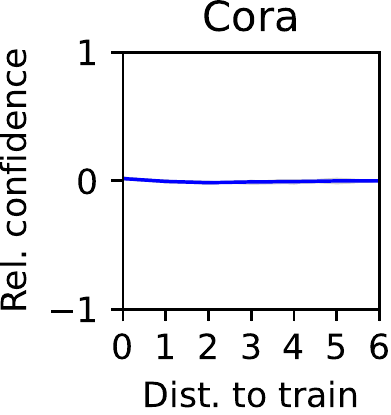}
		\end{subfigure}
		\begin{subfigure}[c]{0.135\textwidth}
			\includegraphics[width=\textwidth]{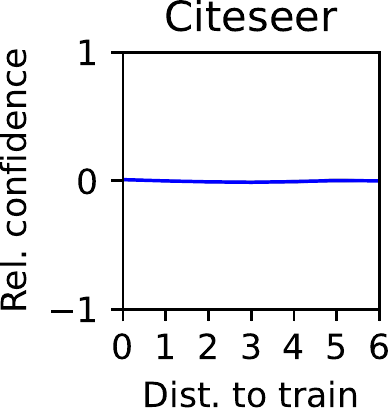}
		\end{subfigure}
		\begin{subfigure}[c]{0.135\textwidth}
			\includegraphics[width=\textwidth]{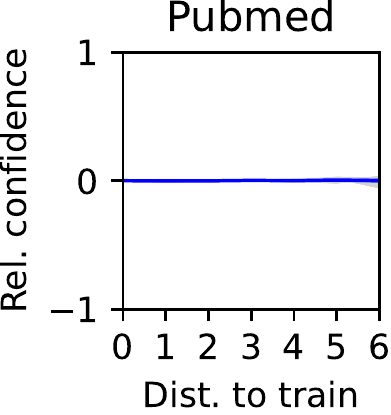}
		\end{subfigure}
		\begin{subfigure}[c]{0.135\textwidth}
			\includegraphics[width=\textwidth]{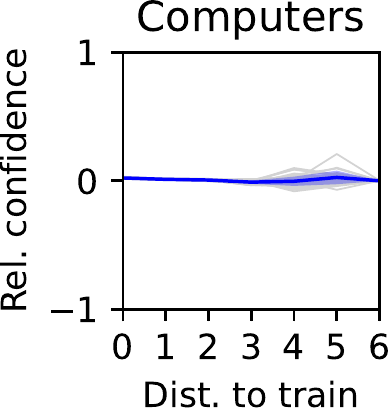}
		\end{subfigure}
		\begin{subfigure}[c]{0.135\textwidth}
			\includegraphics[width=\textwidth]{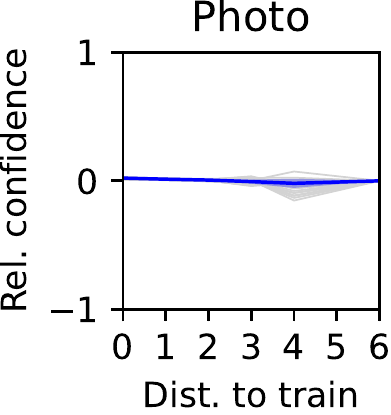}
		\end{subfigure}
		\begin{subfigure}[c]{0.135\textwidth}
			\includegraphics[width=\textwidth]{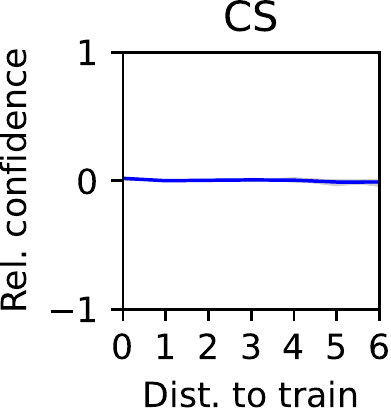}
		\end{subfigure}
		\begin{subfigure}[c]{0.135\textwidth}
			\includegraphics[width=\textwidth]{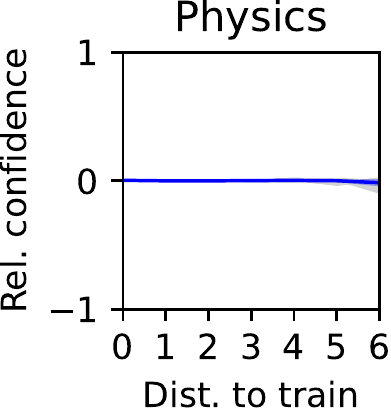}
		\end{subfigure}
		\caption{Relative confidence level of GAT results depending on the minimum distance to training nodes.}
		\label{fig:dist2dconf_gat}
	\end{figure}

\begin{figure}[ht!]
		\centering
		\begin{subfigure}[c]{0.135\textwidth}
			\includegraphics[width=\textwidth]{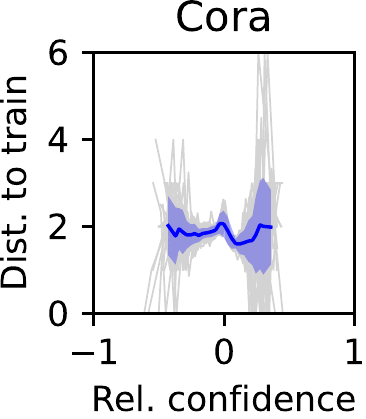}
		\end{subfigure}
		\begin{subfigure}[c]{0.135\textwidth}
			\includegraphics[width=\textwidth]{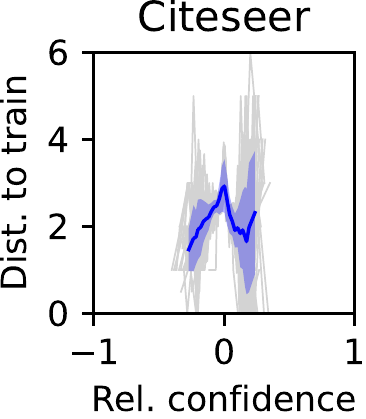}
		\end{subfigure}
		\begin{subfigure}[c]{0.135\textwidth}
			\includegraphics[width=\textwidth]{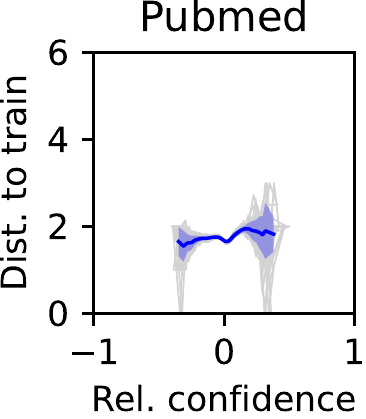}
		\end{subfigure}
		\begin{subfigure}[c]{0.135\textwidth}
			\includegraphics[width=\textwidth]{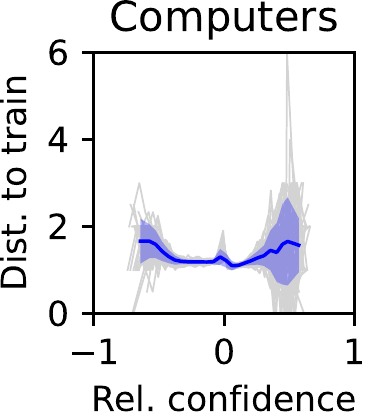}
		\end{subfigure}
		\begin{subfigure}[c]{0.135\textwidth}
			\includegraphics[width=\textwidth]{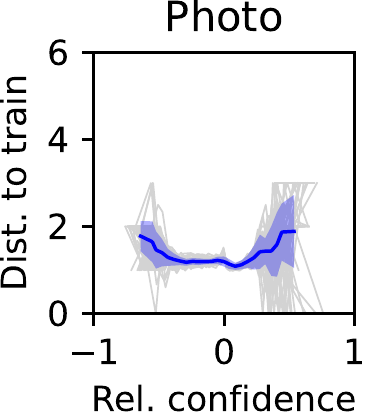}
		\end{subfigure}
		\begin{subfigure}[c]{0.135\textwidth}
			\includegraphics[width=\textwidth]{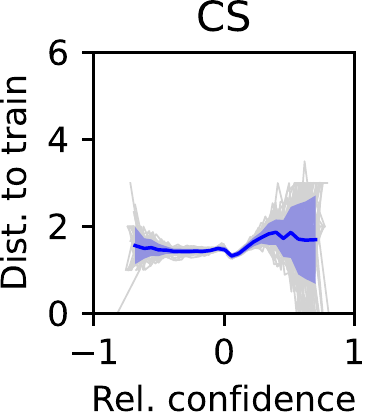}
		\end{subfigure}
		\begin{subfigure}[c]{0.135\textwidth}
			\includegraphics[width=\textwidth]{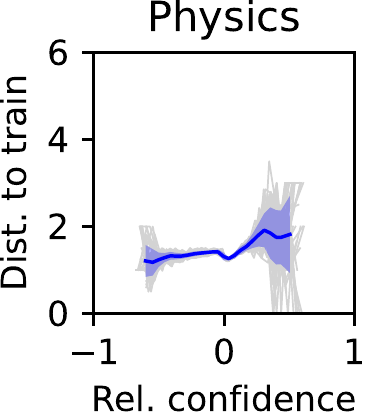}
		\end{subfigure}
		\caption{Minimum distance to training
nodes depending on the relative confidence level of GAT predictions.}
		\label{fig:dconf2dist_gat}
	\end{figure}

	Figures \ref{fig:dist2dconf_gcn}, \ref{fig:dconf2dist_gcn}, \ref{fig:dist2dconf_gat}, and \ref{fig:dconf2dist_gat} show the correlation plots between the distance to training nodes and the relative confidence level. In Figure \ref{fig:dist2dconf_gcn} and \ref{fig:dist2dconf_gat} we see that regardless of the distance to the training nodes, the averaged relative confidence level stays around zero. 
	
	\FloatBarrier
	\subsection{Relative confidence level -- neighborhood similarity}
\begin{figure}[ht!]
		\centering
		\begin{subfigure}[c]{0.135\textwidth}
			\includegraphics[width=\textwidth]{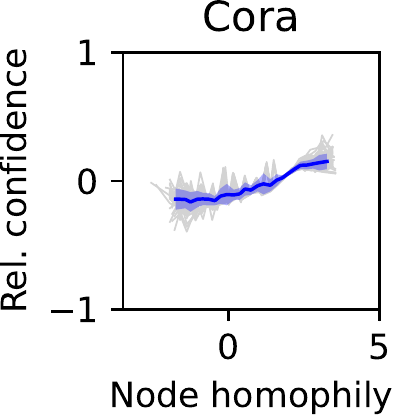}
		\end{subfigure}
		\begin{subfigure}[c]{0.135\textwidth}
			\includegraphics[width=\textwidth]{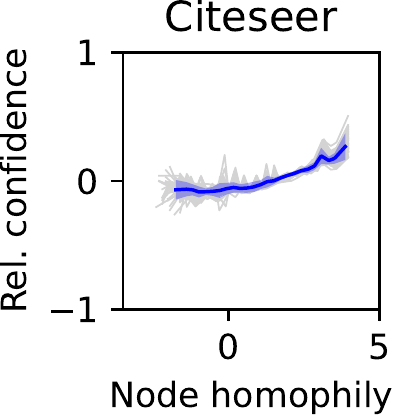}
		\end{subfigure}
		\begin{subfigure}[c]{0.135\textwidth}
			\includegraphics[width=\textwidth]{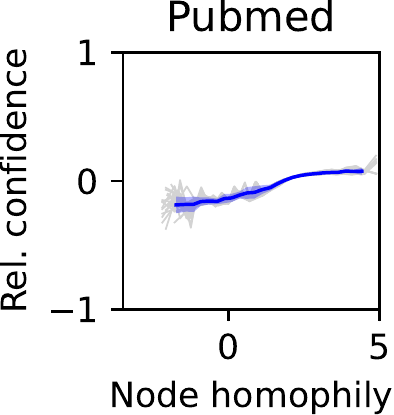}
		\end{subfigure}
		\begin{subfigure}[c]{0.135\textwidth}
			\includegraphics[width=\textwidth]{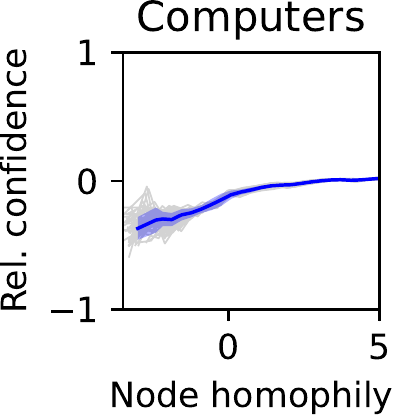}
		\end{subfigure}
		\begin{subfigure}[c]{0.135\textwidth}
			\includegraphics[width=\textwidth]{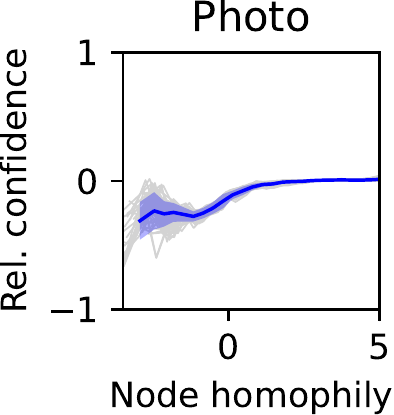}
		\end{subfigure}
		\begin{subfigure}[c]{0.135\textwidth}
			\includegraphics[width=\textwidth]{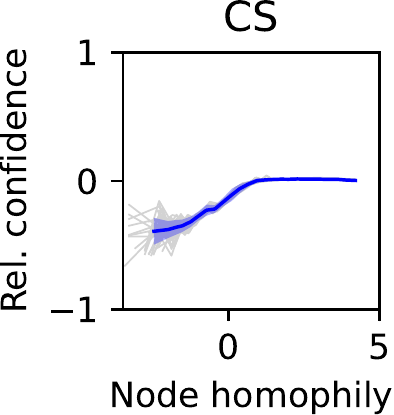}
		\end{subfigure}
		\begin{subfigure}[c]{0.135\textwidth}
			\includegraphics[width=\textwidth]{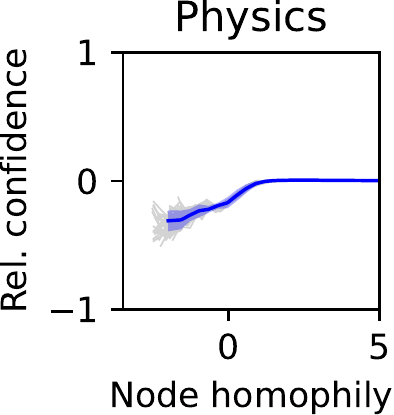}
		\end{subfigure}
		\caption{Relative confidence level of GCN predictions depending on the node homophily.}
		\label{fig:hom2dconf_gcn}
	\end{figure}

\begin{figure}[ht!]
		\centering
		\begin{subfigure}[c]{0.135\textwidth}
			\includegraphics[width=\textwidth]{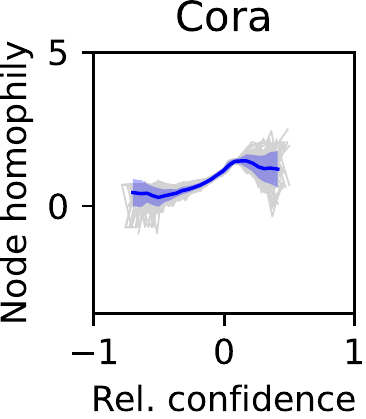}
		\end{subfigure}
		\begin{subfigure}[c]{0.135\textwidth}
			\includegraphics[width=\textwidth]{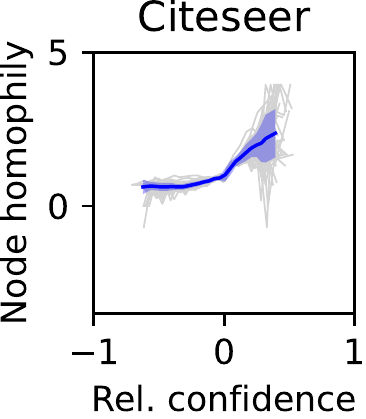}
		\end{subfigure}
		\begin{subfigure}[c]{0.135\textwidth}
			\includegraphics[width=\textwidth]{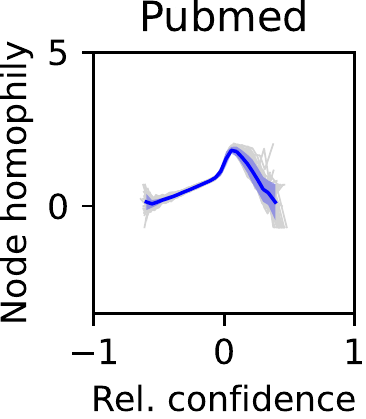}
		\end{subfigure}
		\begin{subfigure}[c]{0.135\textwidth}
			\includegraphics[width=\textwidth]{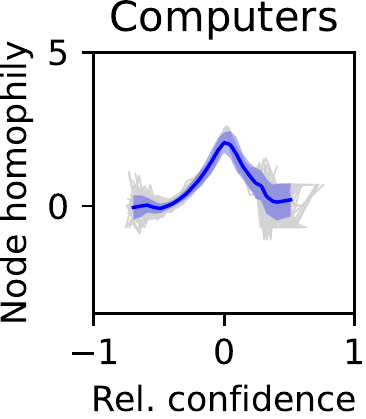}
		\end{subfigure}
		\begin{subfigure}[c]{0.135\textwidth}
			\includegraphics[width=\textwidth]{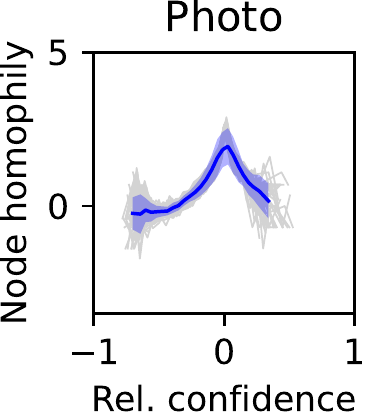}
		\end{subfigure}
		\begin{subfigure}[c]{0.135\textwidth}
			\includegraphics[width=\textwidth]{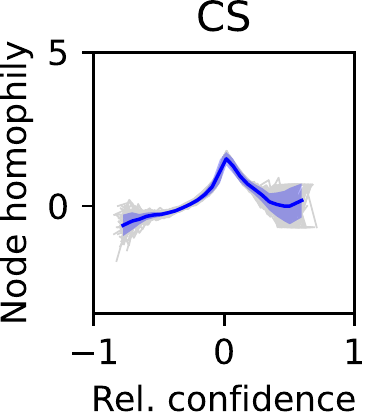}
		\end{subfigure}
		\begin{subfigure}[c]{0.135\textwidth}
			\includegraphics[width=\textwidth]{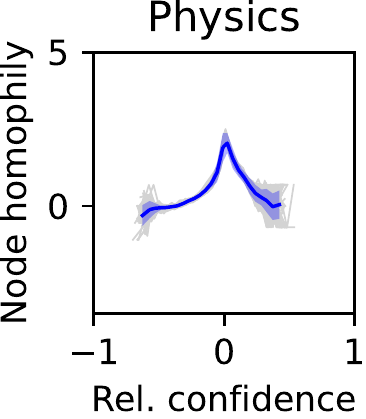}
		\end{subfigure}
		\caption{Node homophily depending on the relative confidence level of GCN predictions.}
		\label{fig:dconf2hom_gcn}
	\end{figure}

\begin{figure}[ht!]
		\centering
		\begin{subfigure}[c]{0.135\textwidth}
			\includegraphics[width=\textwidth]{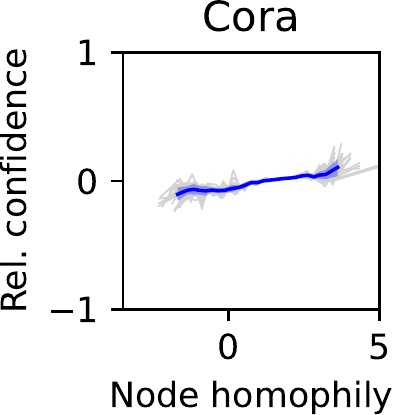}
		\end{subfigure}
		\begin{subfigure}[c]{0.135\textwidth}
			\includegraphics[width=\textwidth]{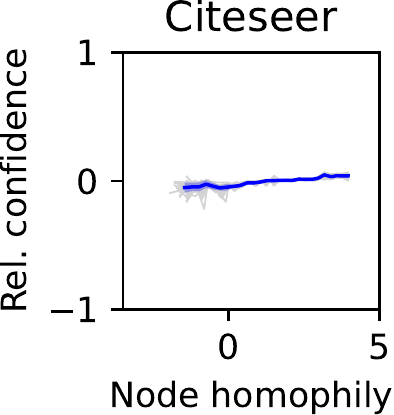}
		\end{subfigure}
		\begin{subfigure}[c]{0.135\textwidth}
			\includegraphics[width=\textwidth]{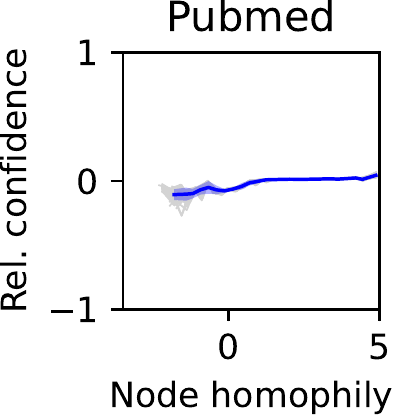}
		\end{subfigure}
		\begin{subfigure}[c]{0.135\textwidth}
			\includegraphics[width=\textwidth]{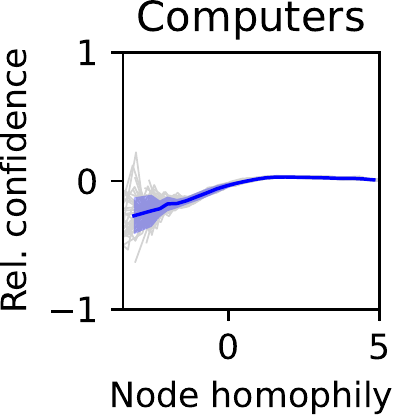}
		\end{subfigure}
		\begin{subfigure}[c]{0.135\textwidth}
			\includegraphics[width=\textwidth]{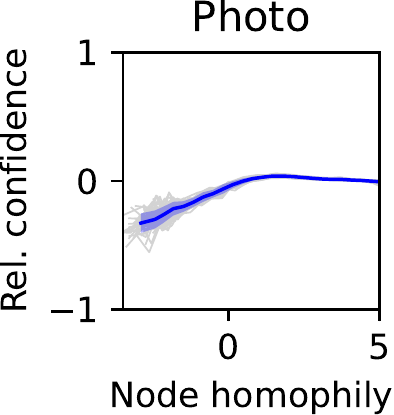}
		\end{subfigure}
		\begin{subfigure}[c]{0.135\textwidth}
			\includegraphics[width=\textwidth]{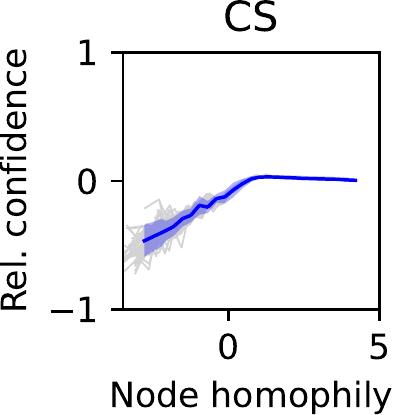}
		\end{subfigure}
		\begin{subfigure}[c]{0.135\textwidth}
			\includegraphics[width=\textwidth]{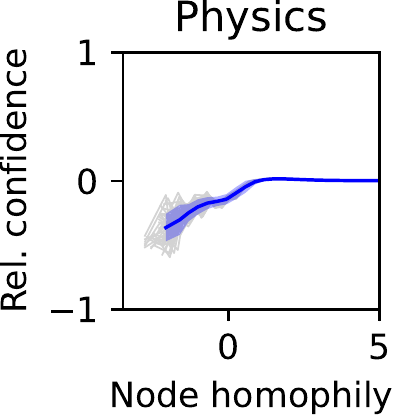}
		\end{subfigure}
		\caption{Relative confidence level of GAT predictions depending on the node homophily.}
		\label{fig:hom2dconf_gat}
	\end{figure}

\begin{figure}[ht!]
		\centering
		\begin{subfigure}[c]{0.135\textwidth}
			\includegraphics[width=\textwidth]{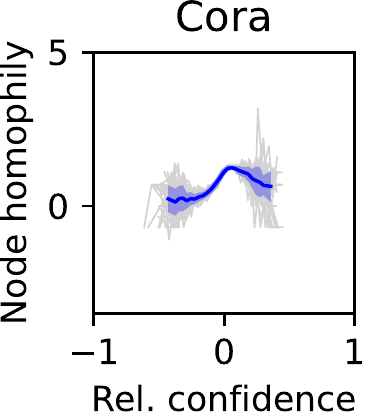}
		\end{subfigure}
		\begin{subfigure}[c]{0.135\textwidth}
			\includegraphics[width=\textwidth]{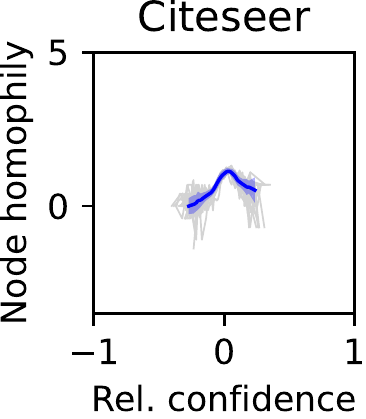}
		\end{subfigure}
		\begin{subfigure}[c]{0.135\textwidth}
			\includegraphics[width=\textwidth]{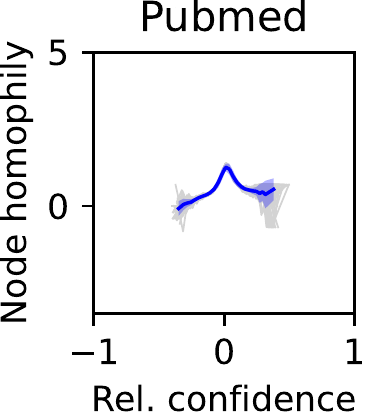}
		\end{subfigure}
		\begin{subfigure}[c]{0.135\textwidth}
			\includegraphics[width=\textwidth]{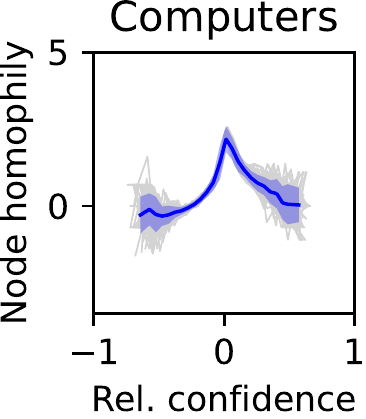}
		\end{subfigure}
		\begin{subfigure}[c]{0.135\textwidth}
			\includegraphics[width=\textwidth]{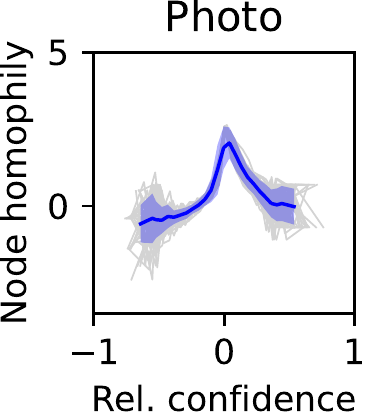}
		\end{subfigure}
		\begin{subfigure}[c]{0.135\textwidth}
			\includegraphics[width=\textwidth]{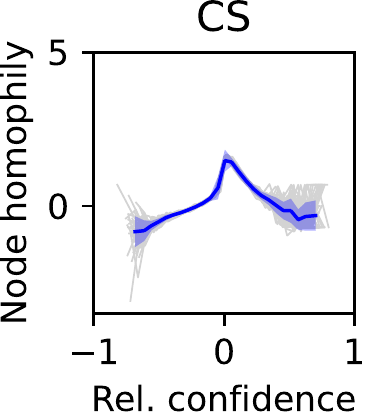}
		\end{subfigure}
		\begin{subfigure}[c]{0.135\textwidth}
			\includegraphics[width=\textwidth]{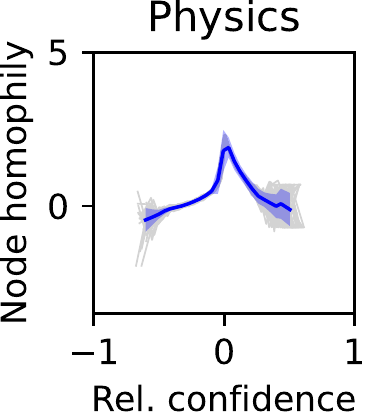}
		\end{subfigure}
		\caption{Node homophily depending on the relative confidence level of GAT predictions.}
		\label{fig:dconf2hom_gat}
	\end{figure}

	Figures \ref{fig:hom2dconf_gcn}, \ref{fig:dconf2hom_gcn}, \ref{fig:hom2dconf_gat}, and \ref{fig:dconf2hom_gat} show the correlations between the relative confidence level and the neighborhood similarity. We observe some partial correlation between these two factors, especially in the negative region of the node homophily. 
	

	\FloatBarrier
	\subsection{Distance to training nodes -- neighborhood similarity}
\begin{figure}[h!]
		\centering
		\begin{subfigure}[c]{0.135\textwidth}
			\includegraphics[width=\textwidth]{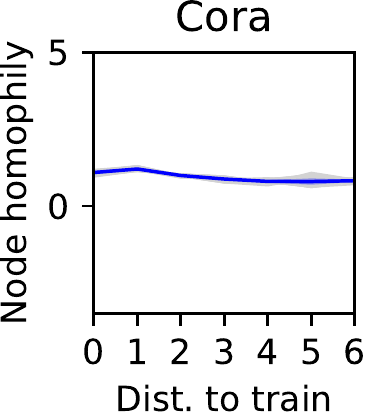}
		\end{subfigure}
		\begin{subfigure}[c]{0.135\textwidth}
			\includegraphics[width=\textwidth]{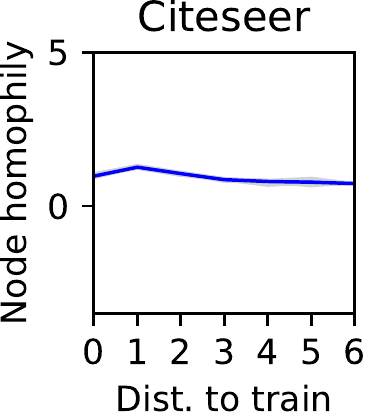}
		\end{subfigure}
		\begin{subfigure}[c]{0.135\textwidth}
			\includegraphics[width=\textwidth]{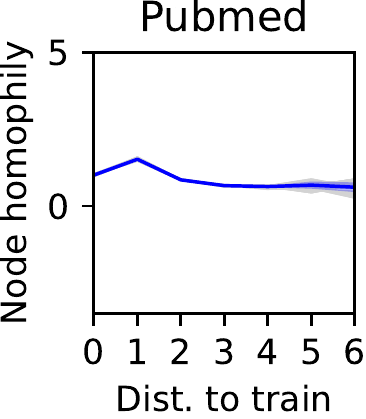}
		\end{subfigure}
		\begin{subfigure}[c]{0.135\textwidth}
			\includegraphics[width=\textwidth]{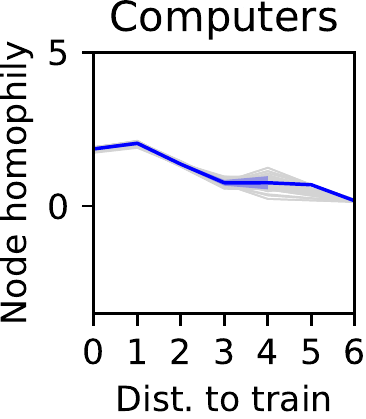}
		\end{subfigure}
		\begin{subfigure}[c]{0.135\textwidth}
			\includegraphics[width=\textwidth]{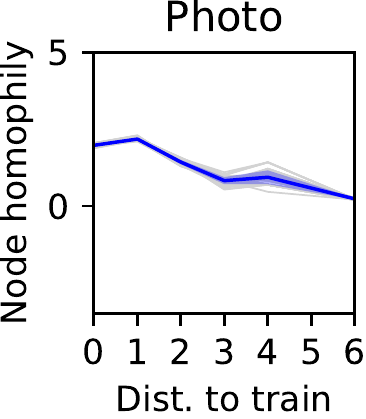}
		\end{subfigure}
		\begin{subfigure}[c]{0.135\textwidth}
			\includegraphics[width=\textwidth]{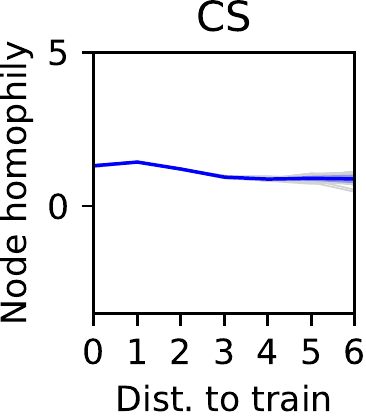}
		\end{subfigure}
		\begin{subfigure}[c]{0.135\textwidth}
			\includegraphics[width=\textwidth]{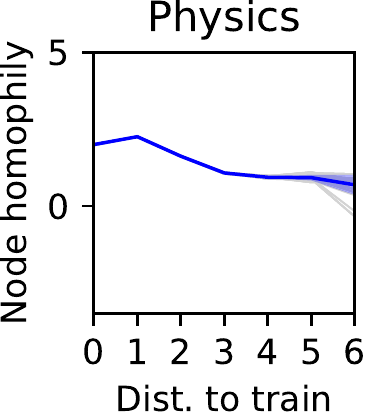}
		\end{subfigure}
		\hspace{0.2\textwidth}
		\caption{Node homophily depending on the minimum distance to training nodes.}
		\label{fig:dist2hom}
	\end{figure}

\begin{figure}[ht!]
		\centering
		\begin{subfigure}[c]{0.135\textwidth}
			\includegraphics[width=\textwidth]{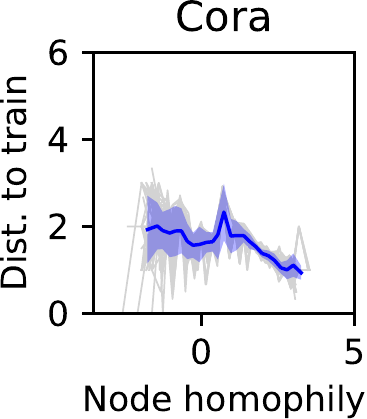}
		\end{subfigure}
		\begin{subfigure}[c]{0.135\textwidth}
			\includegraphics[width=\textwidth]{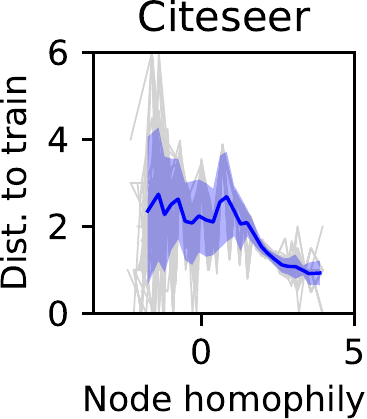}
		\end{subfigure}
		\begin{subfigure}[c]{0.135\textwidth}
			\includegraphics[width=\textwidth]{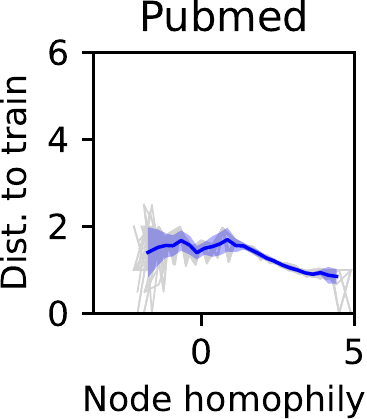}
		\end{subfigure}
		\begin{subfigure}[c]{0.135\textwidth}
			\includegraphics[width=\textwidth]{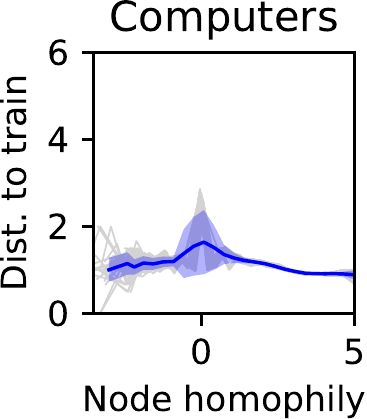}
		\end{subfigure}
		\begin{subfigure}[c]{0.135\textwidth}
			\includegraphics[width=\textwidth]{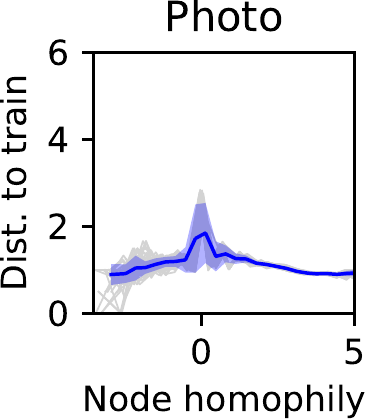}
		\end{subfigure}
		\begin{subfigure}[c]{0.135\textwidth}
			\includegraphics[width=\textwidth]{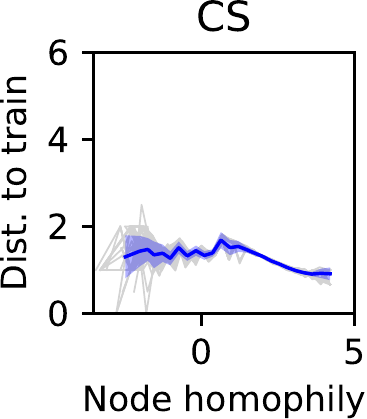}
		\end{subfigure}
		\begin{subfigure}[c]{0.135\textwidth}
			\includegraphics[width=\textwidth]{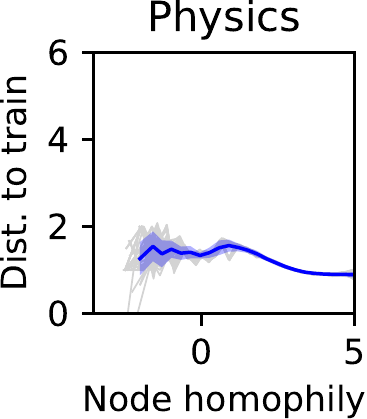}
		\end{subfigure}
		\hspace{0.2\textwidth}
		\caption{Minimum distance to training nodes depending on the node homophily.}
		\label{fig:hom2dist}
	\end{figure}

	Figures \ref{fig:dist2hom} and \ref{fig:hom2dist} show the correlation between the distance to training nodes and neighborhood similarity.
	Note that these two factors are not model-dependent and thus GCN and GAT share the same results.
	We observe that these two factors have a less significant correlation.
	
	\FloatBarrier
	\section{Factor node count analysis}
	In this section we plot the number of test nodes depending on the three local view factors: distance to training nodes, relative confidence level, and neighborhood similarity.

	\subsection{Node count for distance to training nodes}
\begin{figure}[ht!]
		\centering
		\begin{subfigure}[c]{0.135\textwidth}
			\includegraphics[width=\textwidth]{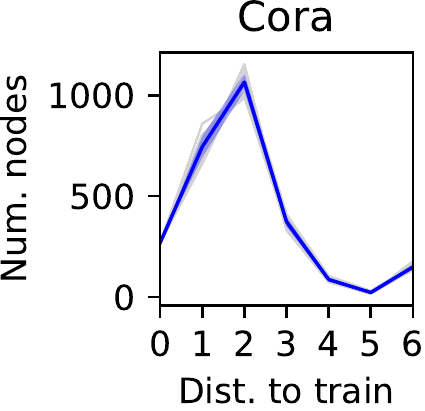}
		\end{subfigure}
		\begin{subfigure}[c]{0.135\textwidth}
			\includegraphics[width=\textwidth]{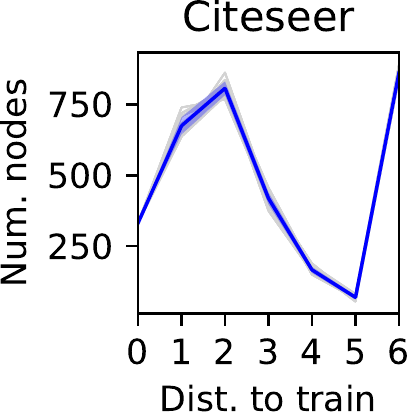}
		\end{subfigure}
		\begin{subfigure}[c]{0.135\textwidth}
			\includegraphics[width=\textwidth]{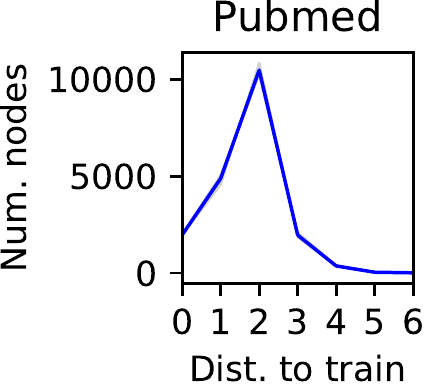}
		\end{subfigure}
		\begin{subfigure}[c]{0.135\textwidth}
			\includegraphics[width=\textwidth]{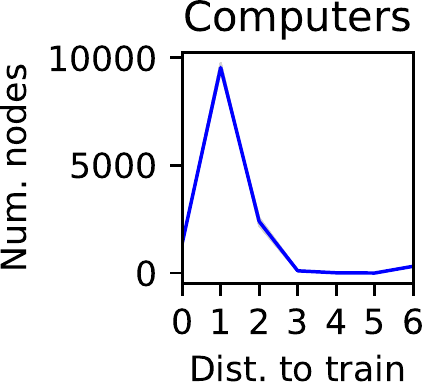}
		\end{subfigure}
		\begin{subfigure}[c]{0.135\textwidth}
			\includegraphics[width=\textwidth]{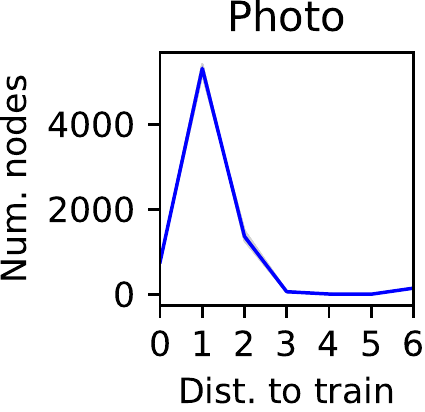}
		\end{subfigure}
		\begin{subfigure}[c]{0.135\textwidth}
			\includegraphics[width=\textwidth]{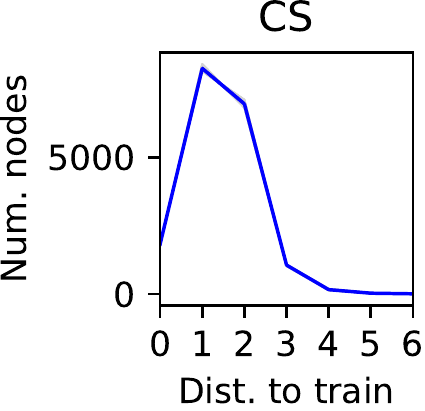}
		\end{subfigure}
		\begin{subfigure}[c]{0.135\textwidth}
			\includegraphics[width=\textwidth]{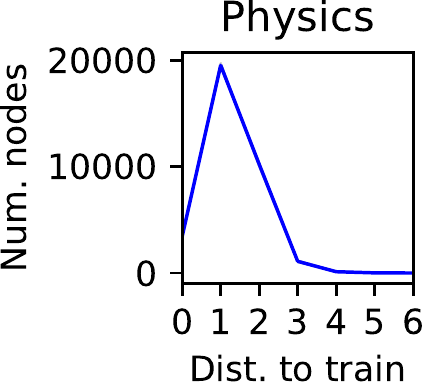}
		\end{subfigure}
		\caption{Number of nodes for the minimum distance to training nodes.} 
		\label{fig:count_dist}
	\end{figure}

	Figure \ref{fig:count_dist} summarizes the node count results of the distance to training nodes.
	We see the majority of nodes can be connected to the training nodes by one or two hops.
    
    \FloatBarrier
	\subsection{Node count for relative confidence level}
\begin{figure}[ht!]
		\centering
		\begin{subfigure}[c]{0.135\textwidth}
			\includegraphics[width=\textwidth]{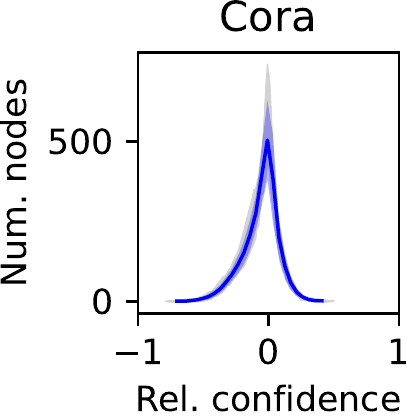}
		\end{subfigure}
		\begin{subfigure}[c]{0.135\textwidth}
			\includegraphics[width=\textwidth]{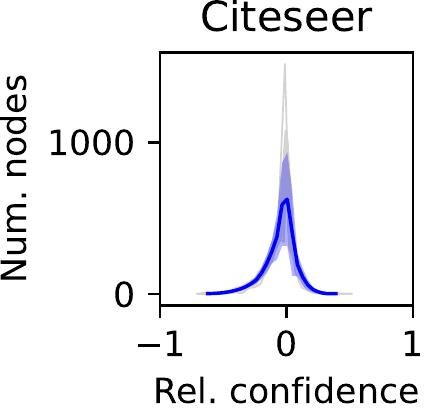}
		\end{subfigure}
		\begin{subfigure}[c]{0.135\textwidth}
			\includegraphics[width=\textwidth]{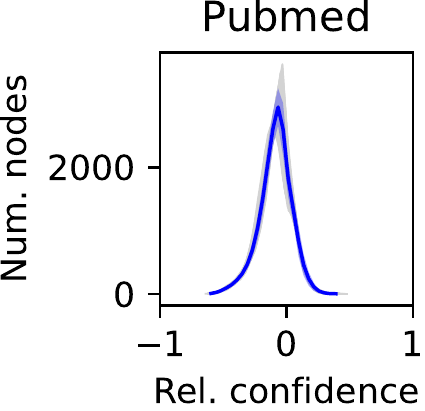}
		\end{subfigure}
		\begin{subfigure}[c]{0.135\textwidth}
			\includegraphics[width=\textwidth]{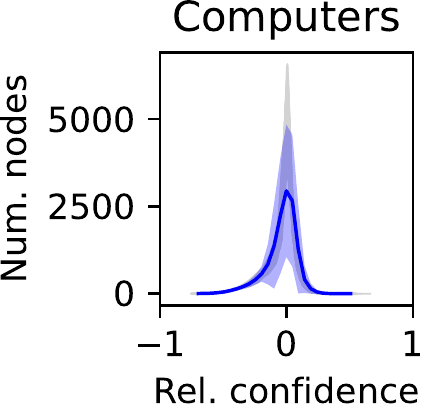}
		\end{subfigure}
		\begin{subfigure}[c]{0.135\textwidth}
			\includegraphics[width=\textwidth]{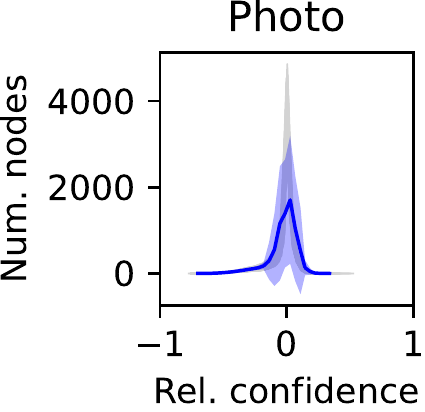}
		\end{subfigure}
		\begin{subfigure}[c]{0.135\textwidth}
			\includegraphics[width=\textwidth]{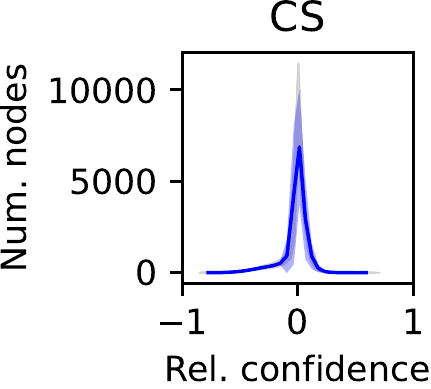}
		\end{subfigure}
		\begin{subfigure}[c]{0.135\textwidth}
			\includegraphics[width=\textwidth]{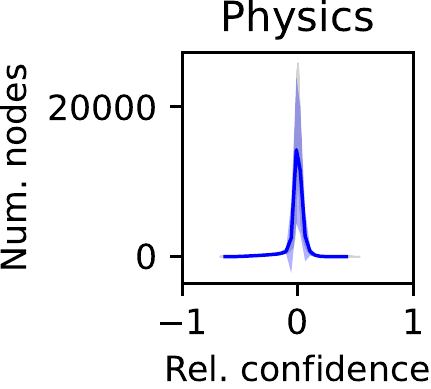}
		\end{subfigure}
		\caption{Number of nodes for the relative confidence level of GCN predictions.} \label{fig:count_diff_conf_gcn}
	\end{figure}

\begin{figure}[ht!]
		\centering
		\begin{subfigure}[c]{0.135\textwidth}
			\includegraphics[width=\textwidth]{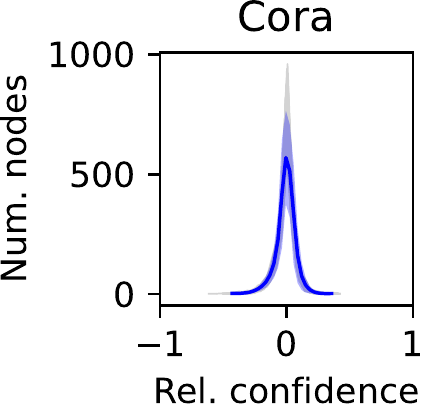}
		\end{subfigure}
		\begin{subfigure}[c]{0.135\textwidth}
			\includegraphics[width=\textwidth]{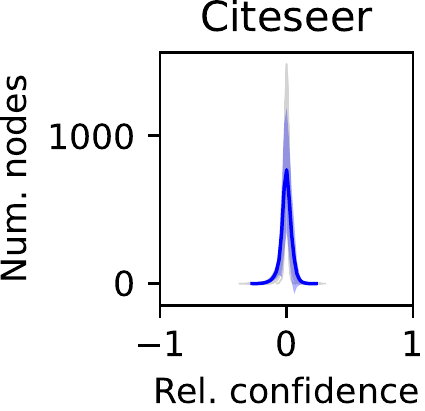}
		\end{subfigure}
		\begin{subfigure}[c]{0.135\textwidth}
			\includegraphics[width=\textwidth]{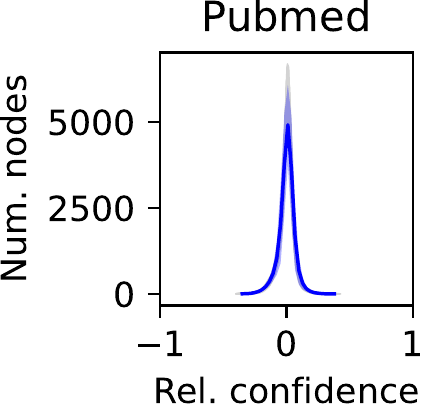}
		\end{subfigure}
		\begin{subfigure}[c]{0.135\textwidth}
			\includegraphics[width=\textwidth]{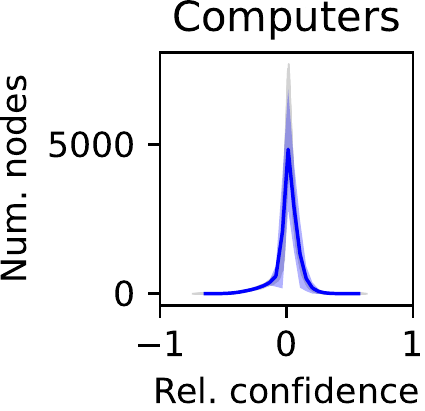}
		\end{subfigure}
		\begin{subfigure}[c]{0.135\textwidth}
			\includegraphics[width=\textwidth]{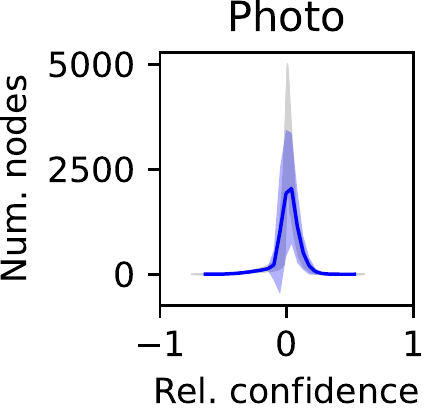}
		\end{subfigure}
		\begin{subfigure}[c]{0.135\textwidth}
			\includegraphics[width=\textwidth]{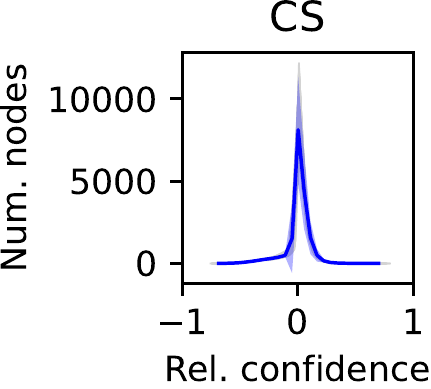}
		\end{subfigure}
		\begin{subfigure}[c]{0.135\textwidth}
			\includegraphics[width=\textwidth]{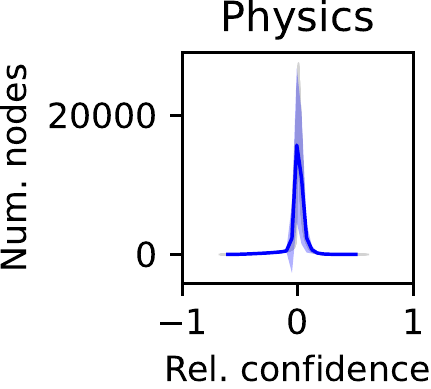}
		\end{subfigure}
		\caption{Number of nodes for the relative confidence level of GAT predictions.} \label{fig:count_diff_conf_gat}
	\end{figure}

	Figure \ref{fig:count_diff_conf_gcn} and \ref{fig:count_diff_conf_gat} are the node count results of the relative confidence level for GCN and GAT respectively.
	We observe that most of the nodes are concentrated around the zero relative confidence level.

	\FloatBarrier
	\subsection{Node count for neighborhood similarity}
\begin{figure}[ht!]
		\centering
		\begin{subfigure}[c]{0.135\textwidth}
			\includegraphics[width=\textwidth]{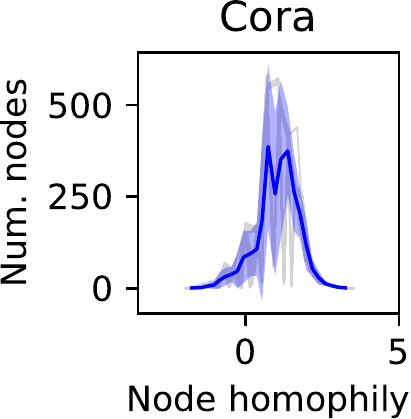}
		\end{subfigure}
		\begin{subfigure}[c]{0.135\textwidth}
			\includegraphics[width=\textwidth]{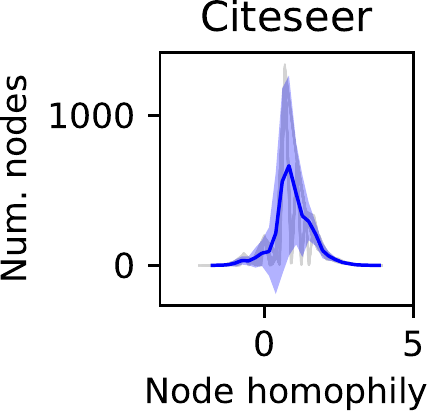}
		\end{subfigure}
		\begin{subfigure}[c]{0.135\textwidth}
			\includegraphics[width=\textwidth]{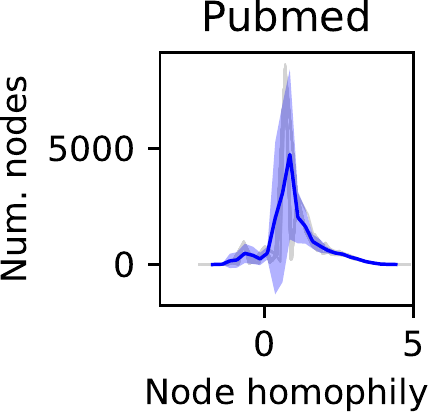}
		\end{subfigure}
		\begin{subfigure}[c]{0.135\textwidth}
			\includegraphics[width=\textwidth]{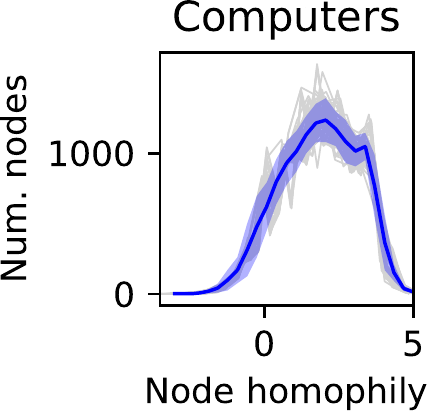}
		\end{subfigure}
		\begin{subfigure}[c]{0.135\textwidth}
			\includegraphics[width=\textwidth]{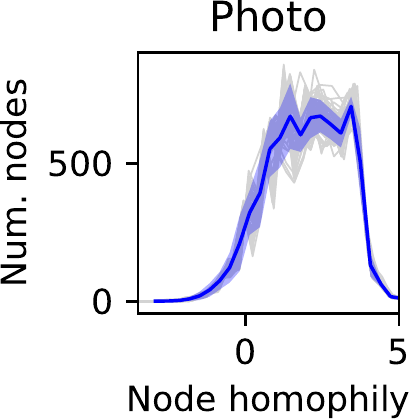}
		\end{subfigure}
		\begin{subfigure}[c]{0.135\textwidth}
			\includegraphics[width=\textwidth]{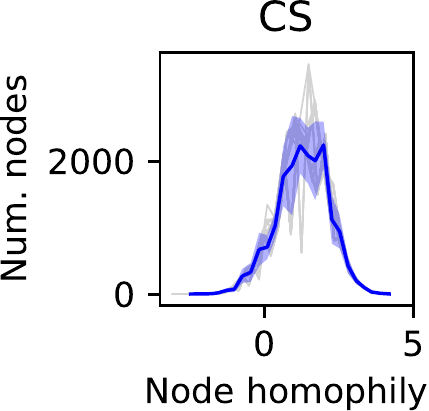}
		\end{subfigure}
		\begin{subfigure}[c]{0.135\textwidth}
			\includegraphics[width=\textwidth]{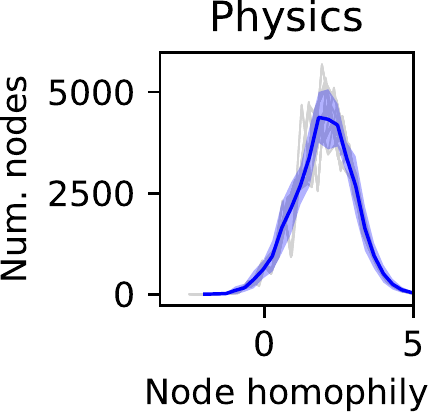}
		\end{subfigure}
		\caption{Number of nodes for the node homophily.} \label{fig:count_node_hom}
	\end{figure}

	Figure \ref{fig:count_node_hom} shows that the majority of nodes lie in the positive homophily region.
	
\end{document}